\documentclass[times,twocolumn,authoryear,final]{elsarticle}

\usepackage{ycviu}
\usepackage{framed,multirow}

\usepackage{amssymb}
\usepackage{latexsym}

\usepackage{url}
\definecolor{newcolor}{rgb}{.8,.349,.1}

\usepackage{multirow}
\usepackage{caption} 
\usepackage[table,dvipsnames]{xcolor} % added table for \rowcolor
\usepackage{tabularx}
\usepackage{makecell}
\usepackage{soul}
\usepackage{enumitem}
\usepackage{amsmath}
\usepackage{stmaryrd}
\usepackage[mathcal]{euscript} % nice looking mathcal notations

\usepackage{pifont}% for checkmark
\usepackage{booktabs,siunitx} % for \toprule and \bottomrule
\usepackage{cite}
\usepackage{flushend}

\usepackage{hyperref}
\hypersetup{
     colorlinks = true,
}

\usepackage{placeins}
\usepackage{subcaption}
\usepackage{tikz}
\usepackage{pgfplots}
    \usetikzlibrary{
        pgfplots.colorbrewer,
        positioning,
        calc,
        spy,
        fadings,
        patterns,
        arrows.meta
    }
    \pgfplotsset{
        cycle list/.define={my marks}{
            every mark/.append style={solid,fill=\pgfkeysvalueof{/pgfplots/mark list fill}},mark=*\\
            every mark/.append style={solid,fill=\pgfkeysvalueof{/pgfplots/mark list fill}},mark=square*\\
            every mark/.append style={solid,fill=\pgfkeysvalueof{/pgfplots/mark list fill}},mark=triangle*\\
            every mark/.append style={solid,fill=\pgfkeysvalueof{/pgfplots/mark list fill}},mark=diamond*\\
    },
}
\tikzstyle{closeup} = [
  opacity=1.0,          
  size=2.95cm,         
  connect spies,  % connects windows by a black line
]

\usepgfplotslibrary{groupplots}
\usepackage{filecontents}

\usepackage{adjustbox}

\usepackage{ctable}

\usepackage{highlight}

\usepackage{ulem}
\def\equationautorefname{Equation}

\def\equationautorefname~#1\null{Eq.~(#1)\null}

\newcommand{\imagealone}{Monodepth2}
\newcommand{\imagesparselidar}{Monodepth2-L}

\newcommand{\PnP}{_{\text{P$n$P}}}
\newcommand{\pnp}{P$n$P}

\newcommand{\ssACMNet}{LiDARTouch-ACMNet} % adapted ACMNet
\newcommand{\ssNLSPN}{LiDARTouch-NLSPN} % adapted NLSPN
\newcommand{\ssSD}{LiDARTouch-S2D} % Self-sup Sparse2Dense
\newcommand{\ssSAN}{LiDARTouch-SAN} % Self-sup Sparse-Auxiliary-Network
\newcommand{\ssmonodepthtwol}{LiDARTouch-Monodepth2-L} % Self-sup MD2-L

\newcommand{\src}{\text{s}}
\newcommand{\tgt}{\text{t}}

\makeatletter
\DeclareRobustCommand\onedot{\futurelet\@let@token\@onedot}
\def\@onedot{\ifx\@let@token.\else.\null\fi\xspace}

\def\eg{\emph{e.g}\onedot} 
\def\ie{\emph{i.e}\onedot}

\makeatother

\renewcommand{\ie}{i.e., }
\renewcommand{\eg}{e.g., }

\renewcommand{\arraystretch}{1.1} % more space between rows
\newcommand{\fzz}[1]{\fontsize{9}{0}{\selectfont #1}}

\newcommand{\parag}[1]{\smallskip\noindent\textbf{#1}~~}

\DeclareRobustCommand*\circled[1]{\tikz[baseline=(char.base)]{
    \node[shape=circle, draw, inner sep=1pt, minimum height=3pt] (char) {#1};}
}

\journal{Computer Vision and Image Understanding}

\begin{document}

\begin{frontmatter}

\title{LiDARTouch: Monocular metric depth estimation with a few-beam LiDAR}

\author[1,2]{Florent \snm{Bartoccioni}} 
\author[1]{\'Eloi \snm{Zablocki}}
\author[1]{Patrick \snm{Pérez}}
\author[1]{Matthieu \snm{Cord}}
\author[2]{Karteek \snm{Alahari}}

\address[1]{Valeo.ai, 100 rue de Courcelle, 75017 Paris, France}
\address[2]{Univ.\ Grenoble Alpes, Inria, CNRS, Grenoble INP, LJK, 38000 Grenoble, France.}

%%%%%%%%%%%%%%%%%%%%%%%%%%%%%%%%%%%%%%
%%%%%%%%%%%%%% ABSTRACT %%%%%%%%%%%%%% 
%%%%%%%%%%%%%%%%%%%%%%%%%%%%%%%%%%%%%%

\begin{abstract}
Vision-based depth estimation is a key feature in autonomous systems, which often relies on a single camera or several independent ones. 
In such a monocular setup, dense depth is obtained with either additional input from one or several expensive LiDARs, e.g., with 64 beams, or camera-only methods, which suffer from scale-ambiguity and infinite-depth problems. 
In this paper, we propose a new alternative of densely estimating \emph{metric} depth by combining a monocular camera with 
a light-weight LiDAR, 
e.g., with 4 beams, typical of today's automotive-grade mass-produced laser scanners.
Inspired by recent self-supervised methods, we introduce a novel framework, called \textit{LiDARTouch}, to estimate dense depth maps from monocular images with the help of ``touches'' of LiDAR, \ie without the need for dense ground-truth depth.
In our setup, the minimal LiDAR input contributes on three different levels: as an additional model's input, in a self-supervised LiDAR reconstruction objective function, and to estimate changes of pose (a key component of self-supervised depth estimation architectures).
Our LiDARTouch framework achieves new state of the art in self-supervised depth estimation on the KITTI dataset, thus supporting our choices of integrating the very sparse LiDAR signal with other visual features.
Moreover, we show that the use of a few-beam LiDAR alleviates scale ambiguity and infinite-depth issues that camera-only methods suffer from. We also demonstrate that methods from the fully-supervised depth-completion literature can be adapted to a self-supervised regime with a minimal LiDAR signal.
\end{abstract}


\end{frontmatter}

%\linenumbers
%%%%%%%%%%%%%%%%%%%%%%%%%%%%%%%%%%%%%%%%%%
%%%%%%%%%%%%%%%%%%%%%%%%%%%%%%%%%%%%%%%%%%
%%%%%%%%%%%%%% INTRODUCTION %%%%%%%%%%%%%% 
%%%%%%%%%%%%%%%%%%%%%%%%%%%%%%%%%%%%%%%%%%
%%%%%%%%%%%%%%%%%%%%%%%%%%%%%%%%%%%%%%%%%%
\section{Introduction}\label{sec:intro}

\begin{table*}[t]
\vspace{-0.1cm}
\centering
\bgroup
\def\arraystretch{1.0}
\caption{\label{tab:highlevel_overview}\textbf{High-level positioning of LiDARTouch vs depth estimation and depth completion methods.}
Our LiDARTouch framework addresses critical weaknesses of self-supervision depth estimation approaches, while being cheaper and far more scalable than fully-supervised depth completion methods.}
\resizebox{\textwidth}{!}{
\begin{tabular}{@{}l@{\hspace{0.3cm}}llcl@{}}
\toprule
&  & \multicolumn{2}{c}{Supervision} &  \\ \cmidrule{3-4}
Approach & Input & Depth regression & Photo. reconst. & Strengths (S) and Weaknesses (W) \\ \midrule 
Depth estimation & Image & No & Yes & S: scales well (self-supervised, very cheap sensor)\\
&&&& W: relative depth, catastrophic estimations (moving objects)\vspace{0.1cm}\\
Depth completion & Image and  & w.r.t. dense GT depth & No & S: metric depth, very good performance \\  
& dense LiDAR &&& W: scales poorly (expensive sensors and GT annotations)\vspace{0.1cm}\\
LiDARTouch (ours) & Image and & w.r.t. few-beam LiDAR & Yes & S: scales well (self-supervised, cheap sensors)\\
& few-beam LiDAR &&& S: metric depth, good performance\\
\bottomrule
\end{tabular}
}
\egroup
\vspace{-0.4cm}
\end{table*}

%%%%%%%%%%%
% CONTEXT
%%%%%%%%%%%
Accurately estimating depth in scenes is a prerequisite for a wide range of computer vision tasks, from computing semantic occupancy grid \citep{bevseg, RGBD_nav_impaired} to object detection without labels~\citep{3DdetectWithoutLabels, RGBD_unsup_detect} and multi-modal unsupervised domain adaptation \citep{xmuda}.
In particular, autonomous systems require an acute spatial understanding of their surroundings to plan and act safely, and the capacity to estimate depth is central to achieving this~\citep{neural_motion_paper,infer, lift_splat_shoot}.
For such applications, two lines of approach exist to infer depth in a scene, depending on the available data: LiDAR-based completion and camera-only estimation methods.
LiDAR-based depth \emph{completion} methods produce depth maps from one or multiple \emph{dense} LiDARs (\eg 32 or 64 beams)~\citep{Depth_Normal_Constraints,guided_depth_completion,jaritz_sparse_and_dense,non_local_spatial_propagation} and essentially interpolate the scene structure from the input signal.
However, these approaches are so far unfit for automotive-grade settings, as they rely on expensive sensors --- often costing more than a car alone --- and require a rich supervisory signal for training, composed of 64-beam LiDAR point clouds densely accumulated over time at a very high acquisition cost.
An alternative is explored by camera-only methods that predict dense depth maps with either stereo~\citep{psmnet,kendall_stereo} or monocular~\citep{monodepth17,monodepth2,packnet, struct2depth,Vid2Depth, DDVO,sfm_learner,Kuznietsov,GeoNet,packnet-semguided} setups.
These models address the task of depth \emph{estimation} and, contrary to the depth completion setup, do not leverage LiDAR point clouds.
While such methods are appealing, as they rely on much cheaper and versatile sensors,
monocular approaches suffer from ambiguity in the map scale they produce: most of them can only generate \emph{relative} depth maps, \ie up to an unknown global scaling factor, which makes them unusable in a real-world setting.\\
Moreover, their predictions can be catastrophic for objects with no relative motion with respect to the ego-camera, \eg vehicles in front, which are likely estimated at infinite depth~\citep{sfm_learner,monodepth2,Vid2Depth,DDVO,GeoNet,packnet,struct2depth}.
Lastly, they are critically impeded by low-light conditions (at night or indoors) and adverse weather (in heavy rain, dense fog or snow storm)~\citep{depth_evaluation_realistic}.

%%%%%%%%%%%
% THE SETTING:USE 4-BEAM LIDAR + SELF-SUPERVISION
% Present why 4-beam lidar
%%%%%%%%%%%
In this paper, we propose the LiDARTouch framework, where dense \emph{metric} depth is estimated by combining a monocular camera with a \emph{minimal} sparse LiDAR input (\eg 4 beams).
Our motivations to use a sparse LiDAR input are diverse.
First, from a practical perspective, 4-beam laser scanners are currently embedded in consumer-grade vehicles and they are a hundred times less expensive than their dense (64-beam) counterparts.
Second, we expect that such a LiDAR signal, although being extremely sparse, can provide valuable cues for monocular depth estimation, thus alleviating scale-ambiguity and infinite-depth problems.
Third, we hypothesize that a light LiDAR touch will result in the overall model correctly estimating the depth of moving objects, notably cars, alleviating the infinite-depth issue.
Finally, from a security perspective, such an approach makes it difficult to attack the camera signal alone~\citep{adversarial_patch}, due to a form of data redundancy between the camera and LiDAR.

%%%%%%%%%%
% SELF-SUPERVISED
%%%%%%%%%%
Leveraging recent advances in monocular depth estimation \citep{sfm_learner,monodepth2,packnet,depth_hints}, our approach is \emph{self-supervised}.
This setting is significantly less data-hungry than the fully-supervised alternative, which requires densified and stereo-filtered depth maps as ground truth \citep{DORN,Depth_Normal_Constraints,guided_depth_completion,jaritz_sparse_and_dense,non_local_spatial_propagation}.
We emphasize that this self-supervised learning setting, combined with the fact that it only involves widely available and low-priced sensors, makes the overall approach particularly scalable.
Indeed, it becomes possible to estimate dense and metric depth maps on datasets and domains lacking depth ground truth \citep{Argoverse,waymo_opendataset,nuscenes2019}.
Moreover, from an industrial perspective, the LiDARTouch framework naturally scales with the data acquired by a vehicle fleet without the need for any annotation.
Under this new regime, we propose the adaptation of recent methods from the two aforementioned streams of approaches for inferring depth.
On the one hand, we adapt fully-supervised depth completion methods, namely ACMNet \citep{ACMNet} and NLSPN \citep{non_local_spatial_propagation}, to a much sparser LiDAR using our self-supervised setup.
On the other hand, we strengthen the very embodiment of self-supervised monocular camera-only methods, namely Monodepth2 \citep{monodepth2}, to integrate the new complementary LiDAR information.
We then perform an extensive study on the contribution brought by the sparse LiDAR signal at different levels as: (1) an additional input, (2) a new information source to estimate better poses, and (3) a form of self-supervision.
A high-level positioning of LiDARTouch with respect to depth estimation and completion approaches is summarized in \autoref{tab:highlevel_overview}.

%%%%%%%%%%%
% EVALUATION AND KEY RESULTS
%%%%%%%%%%%
To evaluate the adapted models and validate our hypotheses, we propose a novel training and evaluation protocol on the KITTI dataset~\citep{kitti} which includes the degradation of the raw 64-beam LiDAR data to obtain 4 beams.
We also propose a new metric to quantitatively measure the infinite-depth problem. This allows us to verify one of our core hypotheses that the use of very limited LiDAR information corrects infinite-depth degeneracies of camera-only methods.
In comparison to depth completion methods, our LiDARTouch framework overcomes the need for depth ground truth and leads to highly improved results with respect to approaches that are naïvely adapted to the self-supervised setting.
In addition, we show that it is possible to successfully adapt architectures from the depth completion literature, as well as camera-based depth estimation methods, into a unified framework which alleviates problems from which these two lines of approaches suffer.

\begin{figure*}
 \centering
 \begin{subfigure}[b]{0.49\linewidth}
     \includegraphics[width=\linewidth]{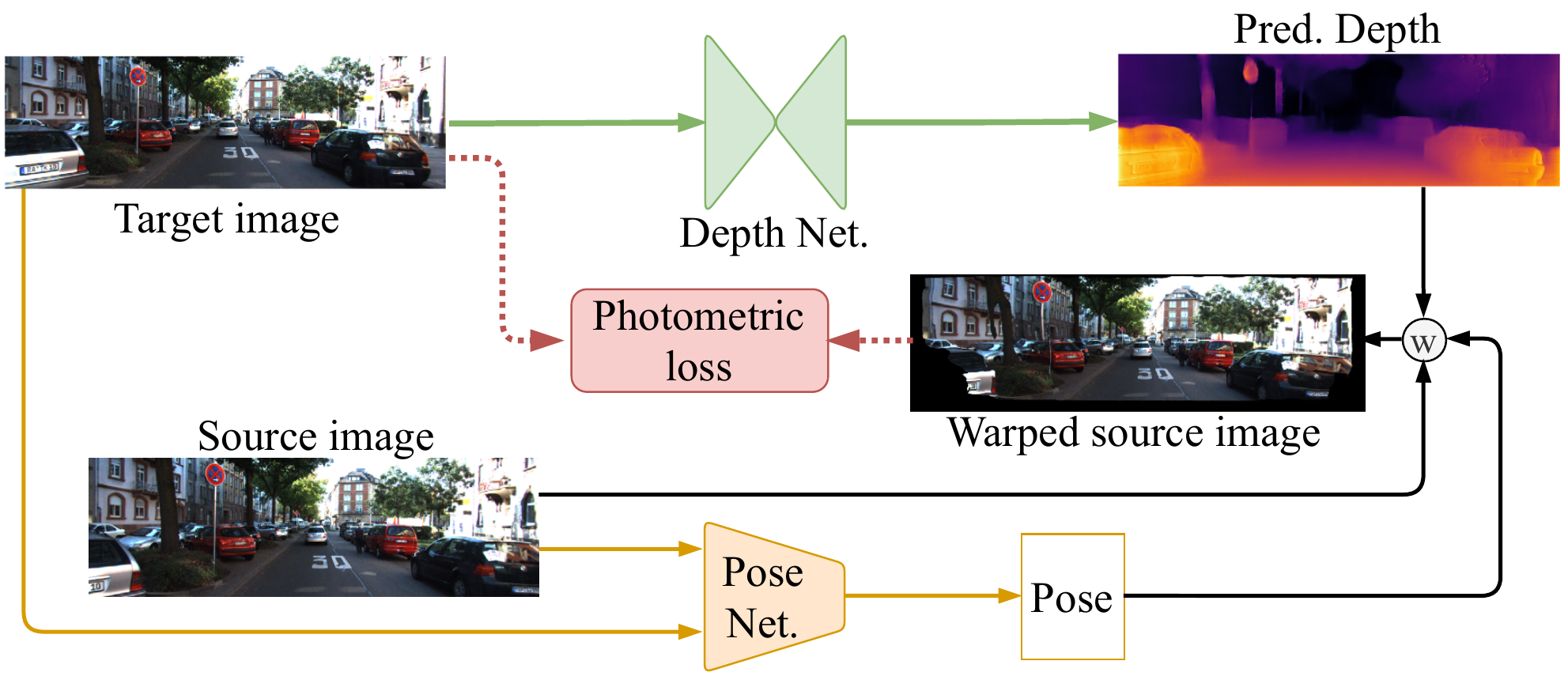}
     \caption{\centering Self-supervised image-only depth estimation. %\protect{\circled{w}}: warping given pose and depth.
     }
     \label{fig:supervision:estimation}
 \end{subfigure}%
 \hfill
 \begin{subfigure}[b]{0.49\linewidth}
     \includegraphics[width=\linewidth]{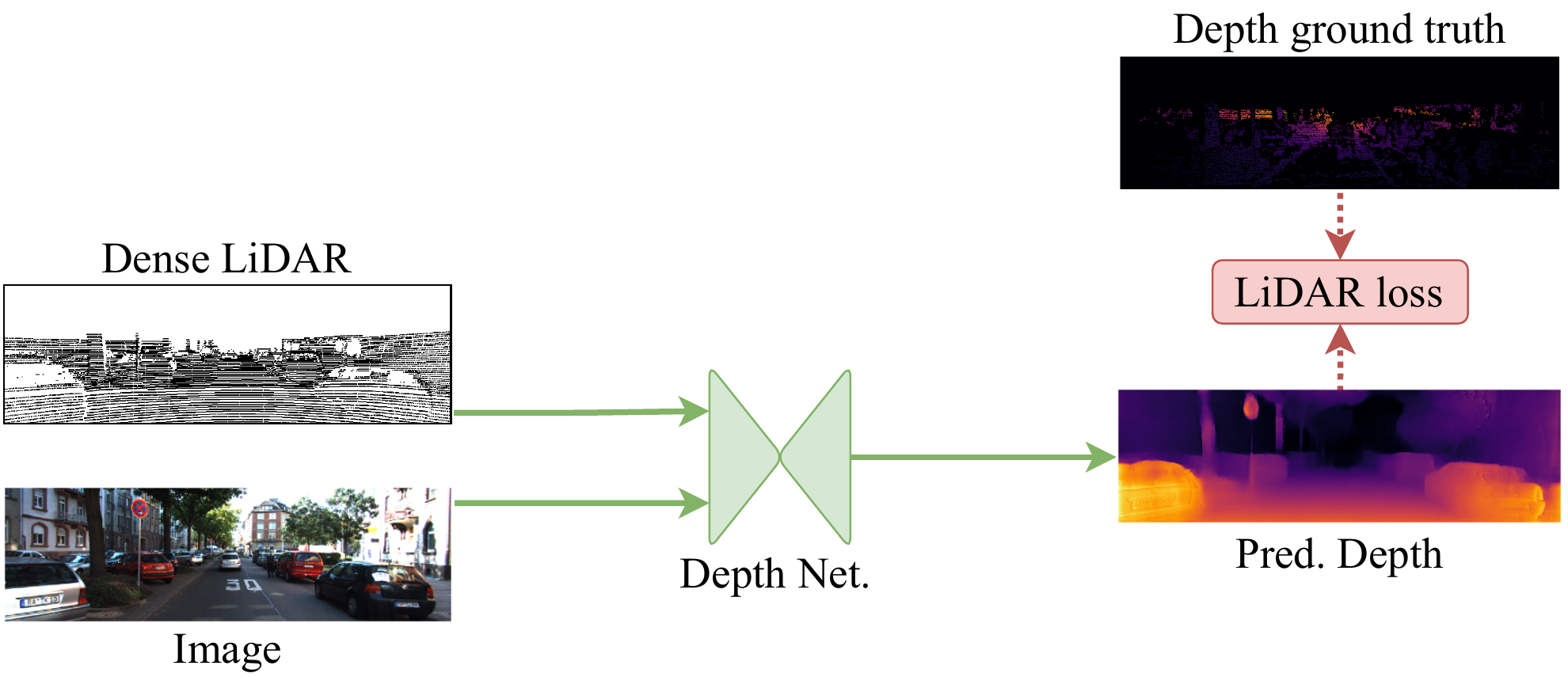}
     \caption{\centering Fully-supervised depth completion.}
     \label{fig:supervision:completion}
 \end{subfigure}
\caption{\textbf{Illustration of the two paradigms for depth estimation.} (a) The left figure shows the classical learning system from self-supervised image-only depth estimation literature, e.g., SfMLearner~\citep{sfm_learner} or Monodepth2~\citep{monodepth2}.
The model is trained to resynthesize the target image given (i) neighboring source images with different viewpoints, (ii) the estimated depth of the target image, and (iii) the relative change of pose between the target and source views. \circled{w} denotes image warping given pose change and target depth map.
(b) This figure summarizes the depth completion pipeline, e.g., models ACMNet~\citep{ACMNet} or NLSPN~\citep{non_local_spatial_propagation}, which employs a multi-modal depth prediction network that is learned by regressing a provided ground-truth depth.
}
\label{fig:related_work}
\end{figure*}
\begin{figure}[hb!] 
\centering
\begin{tikzpicture}[
    image/.style = {
        text width=\linewidth,
        inner sep=0pt, 
        outer sep=0pt,
        draw,
        line width=0.3mm
        },
    label/.style = {
        inner sep=2pt,
        font=\small,
        align=center,
    },
    node distance = 1mm and 1mm
]

%%%%%%%%%%%%%%%%%%%%%%%%%%  4-beam  %%%%%%%%%%%%%%%%%%%%%%%%
    \node [image, line width=0.2mm] (sparse) %{\includegraphics[width=\linewidth]{figures/LIDAR/val_287_sparselidar.png}};
    {\includegraphics[width=\linewidth]{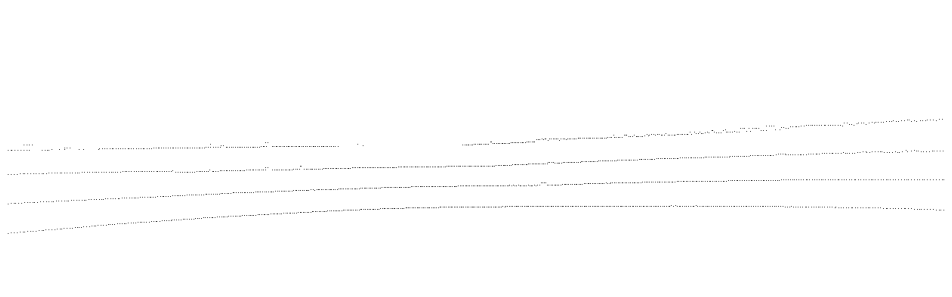}};
    % LEGEND
    \node [label, anchor=north] at ($(sparse.north)$) {\emph{Minimal} / \emph{Sparse} 4-beam};
    
%%%%%%%%%%%%%%%%%%%%%%%% 64-beam  %%%%%%%%%%%%%%%%%%%%%%%%
    \node [image, below=of sparse] (dense) {\includegraphics[width=\linewidth]{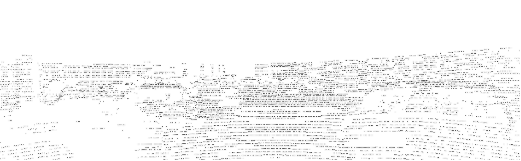}};

    % LEGEND
    \node [label, anchor=north] at ($(dense.north)$) {\emph{Dense} 64-beam};
    
%%%%%%%%%%%%%%%%%%%%%%%%% GT  %%%%%%%%%%%%%%%%%%%%%%%%
    \node [image, below=of dense] (GT) {\includegraphics[width=\linewidth]{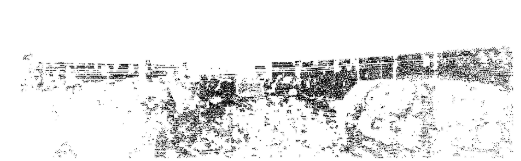}};
    
    % LEGEND
    \node [label, anchor=north] at ($(GT.north)$) {\emph{Accumulated} LiDAR};
     
\end{tikzpicture}
\caption{\textbf{Different LiDAR densities.} \emph{Dense} 64-beam point clouds are typically used as the input of depth completion approaches, which are supervised with \emph{accumulated} LiDAR seen as ground truth (GT). These point clouds are far denser than the \emph{minimal} LiDAR we use. Note that LiDAR data is often not available in the upper part of the scenes.}
\label{fig:lidar_density}
\end{figure}
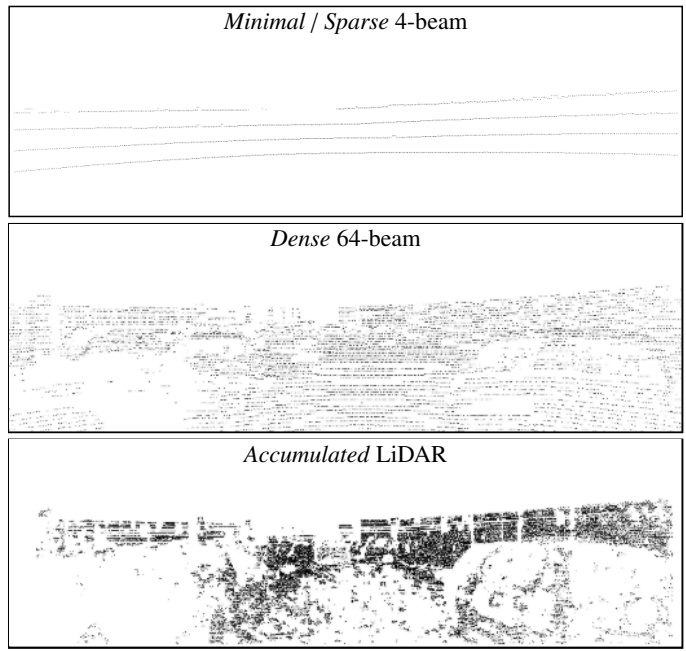

We make the following contributions:
\begin{enumerate}[topsep=0cm,partopsep=0cm,leftmargin=0.5cm,noitemsep]
\item We propose LiDARTouch, a new \emph{self-supervised} depth estimation framework, where a \emph{minimal} LiDAR and a monocular camera are available without access to any ground-truth depth annotations.
This configuration is close to \textit{in situ} conditions of today's vehicles, which is seldom addressed in other works. 

\item We demonstrate that models trained within our LiDARTouch framework close the performance gap between self-supervised monocular depth estimation and fully-supervised depth completion learning schemes, proving that the need for ground-truth acquisition and costly sensors can be alleviated.

\item We show that models trained within our LiDARTouch framework do not suffer from critical scale-ambiguity and infinite-depth issues, in contrast to camera-only models. We evaluate this a novel metric to quantitatively measure the infinite-depth issue for the first time in the literature.

\item We demonstrate that LiDARTouch is a versatile learning framework by successfully applying it to a range of network architectures: Networks from the depth-completion literature are revamped to work with very sparse LiDAR instead of dense ones and camera-only models are adapted to integrate LiDAR data.

\item We study the influence of LiDAR inputs at each stage of our framework extensively. Our experiments show that integrating sparse LiDAR in a self-supervised scheme is not trivial. We provide key insights for the community on how the fusion scheme, the pose method and the supervisions interact.

\end{enumerate}

%%%%%%%%%%%%%%%%%%%%%%%%%%%%%%%%%%%%%%%%%%%%%%%%%%% 
%%%%%%%%%%%%%%%%%%%%%%%%%%%%%%%%%%%%%%%%%%%%%%%%%%% 
%%%%%%%%%%% Background and Related Work %%%%%%%%%%% 
%%%%%%%%%%%%%%%%%%%%%%%%%%%%%%%%%%%%%%%%%%%%%%%%%%% 
%%%%%%%%%%%%%%%%%%%%%%%%%%%%%%%%%%%%%%%%%%%%%%%%%%% 
\section{Background and related work}
\label{sec:rw}
In the remainder of this paper we refer to a LiDAR as \emph{dense} if it has more than 32 beams, and call it \emph{sparse} or \emph{minimal} otherwise. 
Depth ground truth, required by fully-supervised methods, is obtained from a dense LiDAR signal, accumulated over several sweeps. A camera stereo setup is then used to remove trail artifacts from moving objects. We will refer to such densified point-cloud data as \emph{accumulated} LiDAR.
These three density levels are illustrated in \autoref{fig:lidar_density}.
We now detail the two lines of approaches related to our work: camera-only monocular self-supervised methods and LiDAR-based fully-supervised depth completion systems.\newline

\parag{Monocular self-supervised methods.} 
In a fully- or semi-supervised setting, several models estimate depth in a camera-only monocular setup \citep{DORN,Kuznietsov,semidepth}, but acquiring depth ground truth for outdoor environments at scale is challenging and expensive.
To overcome this issue, a few camera-based works~\citep{monodepth17,sfm_learner, struct2depth} propose a \emph{self-supervised} alternative to the use of ground-truth depth. 
Leveraging a set of consecutive frames, this paradigm predicts the depth for one of them and the relative changes in pose across nearby views.
The model is trained by minimizing a photometric reconstruction error defined over these views (\autoref{fig:supervision:estimation}).
Two important issues with such approaches hinder their widespread usage: the scale ambiguity of the produced depth maps and the infinite-depth problem.

The \emph{scale-ambiguity} problem stems from the view synthesis formulation being ill-posed. The formulation is scale ambiguous, as the target view can be correctly reconstructed regardless of the scale of the prediction. As a consequence, estimated depth maps are \emph{relative} --- up to an unknown global scaling factor --- and models thus need additional supervision to accurately estimate a \emph{metric} depth.
Several self-supervised approaches rely on ground-truth LiDAR signal to scale their depth estimation at test time~\citep{sfm_learner, monodepth2, struct2depth, GeoNet, Vid2Depth, DDVO}.
Alternatively, the recent PackNet model~\citep{packnet} proposes to automatically scale estimations with additional constraints imposed by the instantaneous velocity of the ego-vehicle.
Some works have also moved to a stereo setup to disambiguate the scale factor, using additional information, at train time only \citep{monodepth17, depth_adversialreconstruct} or also at run time \citep{psmnet,kendall_stereo, lidarstereonet}, thus abandoning the monocular setup.

The second issue of infinite depth arises when objects move at the same speed as the camera.
In this common situation, a trivial solution for the model is to predict that these objects are infinitely far and big, as they do not change in appearance through time~\citep{sfm_learner,monodepth2,packnet}. 
Recent proposals to address this problem exploit semantic segmentation of classes known to be often dynamic (\eg cars, trucks)~\citep{struct2depth, struct2depth_motion}, or automatically prune the dataset by removing these objects~\citep{packnet-semguided}.
The robustness of both these approaches to novel test scenarios, however, remains unclear.

% ours
In our work, we build on camera-only methods to additionally integrate LiDAR information and show that: (i)~very few direct depth measures suffice to have a metrically-scaled dense depth estimation, and (ii)~the infinite-depth issue can be partially or completely solved with the use of LiDAR input, depending on its resolution and position, without any additional assumptions.

\parag{Depth completion methods}
typically estimate a dense depth map from raw LiDAR measurements.
Current deep-learning based methods for depth completion~\citep{Depth_Normal_Constraints,guided_depth_completion,jaritz_sparse_and_dense,non_local_spatial_propagation, sparse_to_dense, monocular_fisheye_camera,ACMNet} usually learn to regress ground-truth depth maps in a fully-supervised setup (\autoref{fig:supervision:completion}).
Such approaches generally operate over RGB and LiDAR inputs.

A popular approach is to use one encoder per modality and fuse them at each resolution scale  \citep{guided_depth_completion, packnet-san} or at the feature bottleneck only~\citep{jaritz_sparse_and_dense}. An other option is early fusion, where both modalities are concatenated at the very begining of the architecture \citep{Depth_Normal_Constraints, non_local_spatial_propagation, selfsup_sparse_to_dense}
Some fusion module, as the one of GuideNet~\citet{guided_depth_completion}, only considers the image as a guiding signal for the LiDAR features. This assumes that the LiDAR input is sufficient, \ie high-resolution, for estimating depth, and thus unsuitable for our case.
This limits the approach~\citet{guided_depth_completion} to estimate depth from high-resolution 64-beam LiDAR both at train and run time, making it incomparable to ours as we do not have access to such data. 
On the contrary, the SAN architecture~\citep{packnet-san}, can handle various levels of LiDAR sparsity with sparse convolutions.
Alternatively, networks like ACMNet~\citep{ACMNet} and NLSPN~\citep{non_local_spatial_propagation} propagate sparse LiDAR features into image features where depth measurements are not available.
ACMNet~\citep{ACMNet} uses a multi-scale co-attention-guided graph propagation strategy for depth completion. It propagates the sparse and irregularly distributed LiDAR measurements through a nearest-neighbor encoding. In addition, it uses a symmetric gated fusion strategy to fuse multi-modal contextual information throughout the decoder.
The NLPSN architecture~\citep{non_local_spatial_propagation} jointly estimates an initial depth map, a pixel-wise confidence and non-local affinity kernels. This initial depth map is iteratively refined with the input LiDAR features using the predicted confidence map and affinity kernels.

All the aforementioned depth completion methods employ a 64-beam input LiDAR and are trained with accumulated LiDAR as supervision. Here, most of the scene structure is available and the task amounts to color-guided depth interpolation.
This design prevents these works from being easily adapted to new domains.
Indeed, acquisition of ground-truth data is expensive and not scalable, as it is obtained from high-resolution LiDARs and stereo cameras. 
In contrast, our work specifically focuses on minimal 4-beam LiDAR directly, with no densely accumulated LiDAR data as supervision.
We emphasize that in this very sparse 4-beam regime, almost no structural information can be directly extracted for the input signal.
The task we propose is then more akin to depth estimation than depth completion.

A closely related work to ours is the model of \citet{selfsup_sparse_to_dense}, which also uses LiDAR as a \emph{supervisory} signal in a monocular self-supervised setting.
LiDAR and camera signals are merged through an early fusion and the change of pose is estimated by solving a Perspective-$n$-Point problem.
However, their setup is different to ours. Their study focus on the dense depth completion regime, \ie with a 64-beam LiDAR, while we work on depth estimation with a minimal 4-beam LiDAR. 
Moreover, they do not compare against other existing architectures in the self-supervised setting.
In contrast, we perform thorough evaluations with existing work by adapting camera-only and depth completion methods to our extremely scarce LiDAR regime.
Additionally, we propose a different supervision scheme and the use of multiple views in photometric reconstruction. These choices lead to a substantial improvement on the KITTI dataset.
Finally, we provide in-depth analyses on the  impact brought by the LiDAR signal at different levels.

\begin{figure*}[]
    \centering
    \includegraphics[width=\linewidth]{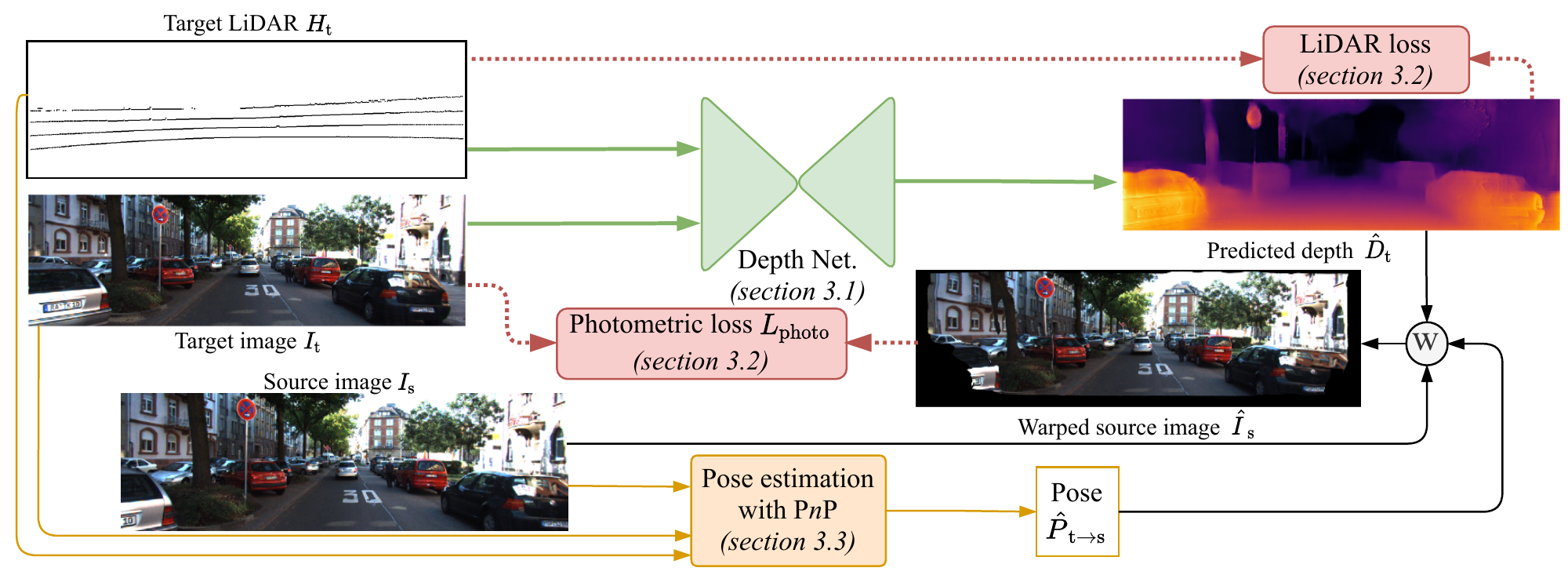}
    \caption{\textbf{Overview of our LiDARTouch learning framework.} 
    The proposed framework leverages ideas from both the camera-only depth estimation approach (illustrated in \autoref{fig:supervision:estimation}) and fully-supervised depth completion methods (illustrated in \autoref{fig:supervision:completion}).
    In LiDARTouch, the light touch of LiDAR is integrated at three different stages: as an input of the depth network, as a self-supervision signal, and to estimate a scaled pose.
    }
    \label{fig:learning_systems}
\vspace{-0.2cm}
\end{figure*}

%%%%%%%%%%%%%%%%%%%%%%%%%%%%%%%%%%%%%%%
%%%%%%%%%%%%%%%%%%%%%%%%%%%%%%%%%%%%%%%
%%%%%%%%%%% System Overview %%%%%%%%%%% 
%%%%%%%%%%%%%%%%%%%%%%%%%%%%%%%%%%%%%%%
%%%%%%%%%%%%%%%%%%%%%%%%%%%%%%%%%%%%%%%
\section{LiDARTouch framework}
\label{sec:learning_system}

This section is organized as three parts, each corresponding to a different and complementary use of the light LiDAR signal.
In \autoref{sec:learning_system:depth_network}, we present the architecture of the depth network, shown in green in \autoref{fig:learning_systems}, which estimates depth by fusing the monocular image with the sparse LiDAR pointcloud. 
In \autoref{sec:learning_system:objectives}, we detail the self-supervision objectives involving a photometric reconstruction along with a LiDAR self-supervision, as illustrated in red in \autoref{fig:learning_systems}.
Lastly, \autoref{sec:learning_system:pose_setups} introduces methods to estimate the relative change of pose between the source and target views, depicted by the orange part of \autoref{fig:learning_systems}.

\begin{figure*}[t]
 \centering
 \begin{subfigure}[h]{0.49\linewidth}
     \centering
     \includegraphics[trim = 0 0 1.5cm 0, clip,width=\linewidth]{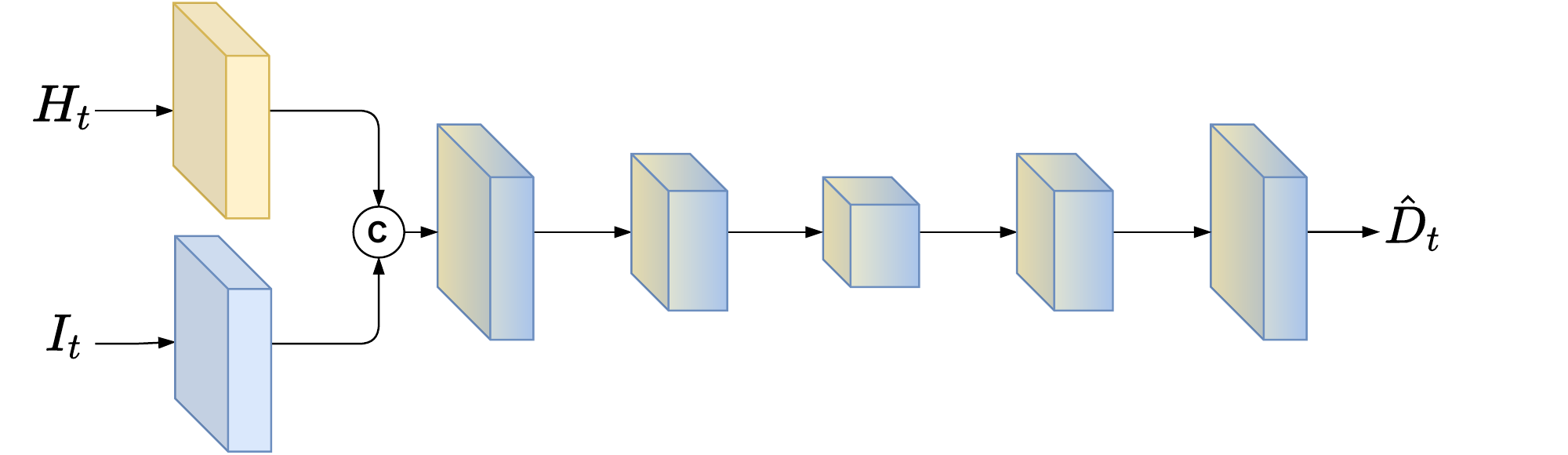}
     \caption{S2D~\citep{selfsup_sparse_to_dense}}
     \label{fig:architectures:S2D}
 \end{subfigure}
 \hfill
 \begin{subfigure}[h]{0.49\linewidth}
     \centering
     \includegraphics[trim = 0 0 1.5cm 0, clip,width=\linewidth]{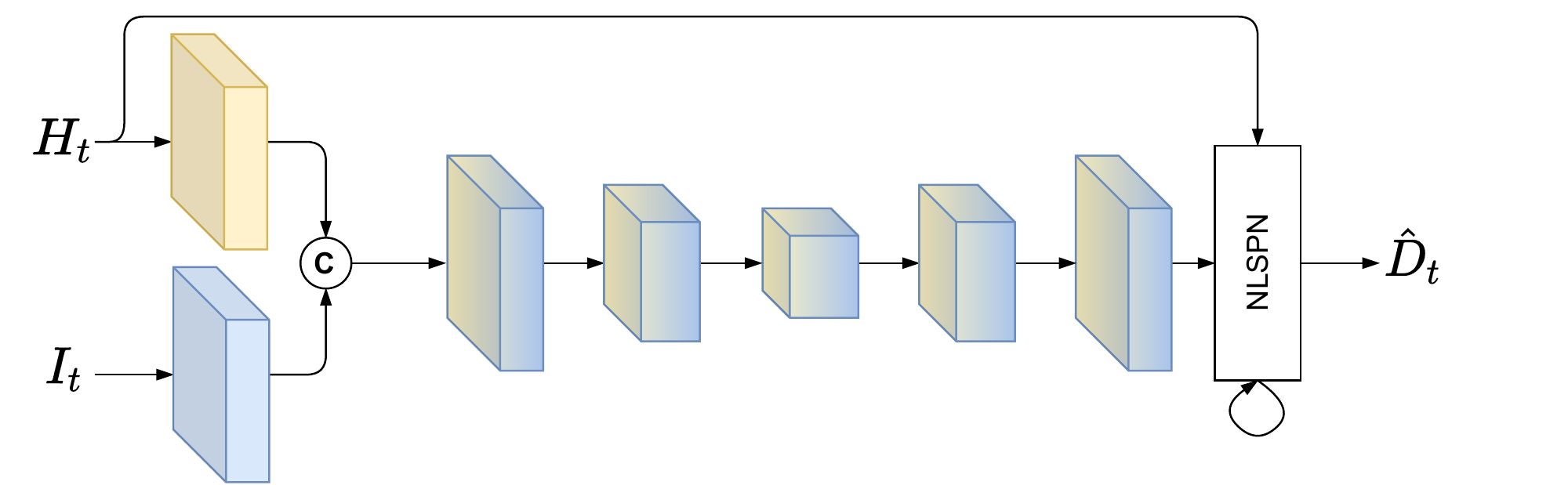}
     \caption{NLSPN~\citep{non_local_spatial_propagation}}
     \label{fig:architectures:NLSPN}
 \end{subfigure}
 \\[2ex] % more vertical space between the figures
 \begin{subfigure}[h]{0.49\linewidth}
     \includegraphics[trim = 0 0 1.5cm 0, clip,width=\linewidth]{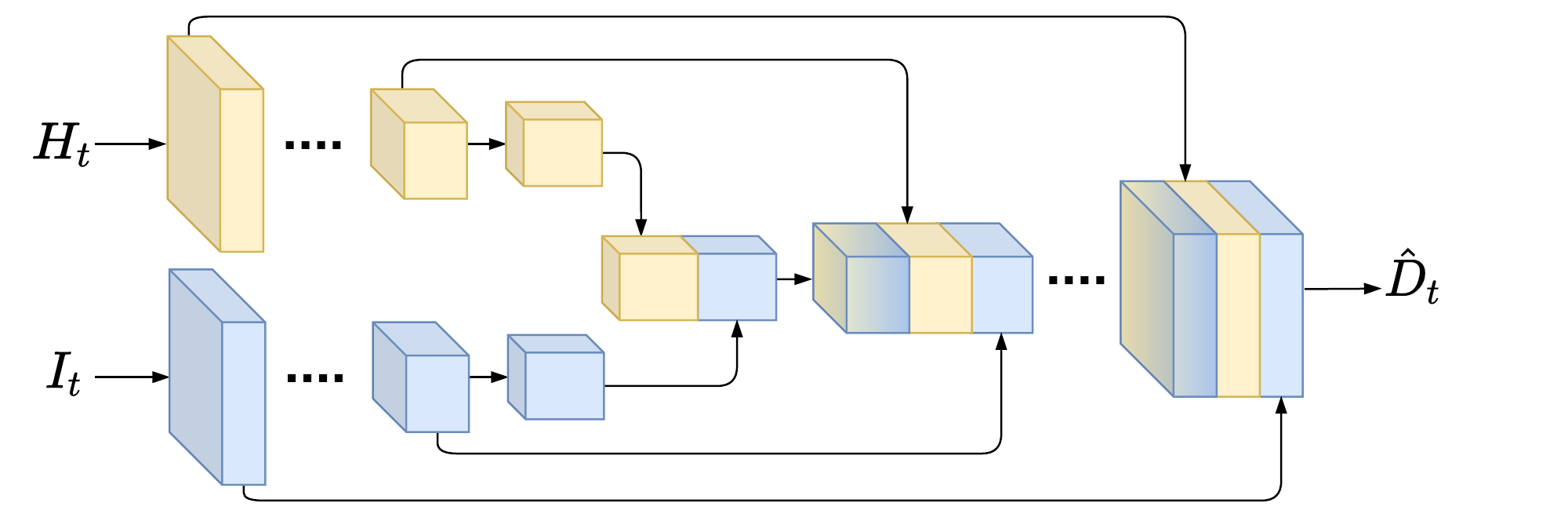}
     \caption{\imagesparselidar, extension of ~\citep{monodepth2}}
     \label{fig:architectures:IL}
 \end{subfigure}
 \hfill
 \begin{subfigure}[h]{0.49\linewidth}
     \includegraphics[trim = 0 0 1.5cm 0, clip,width=\linewidth]{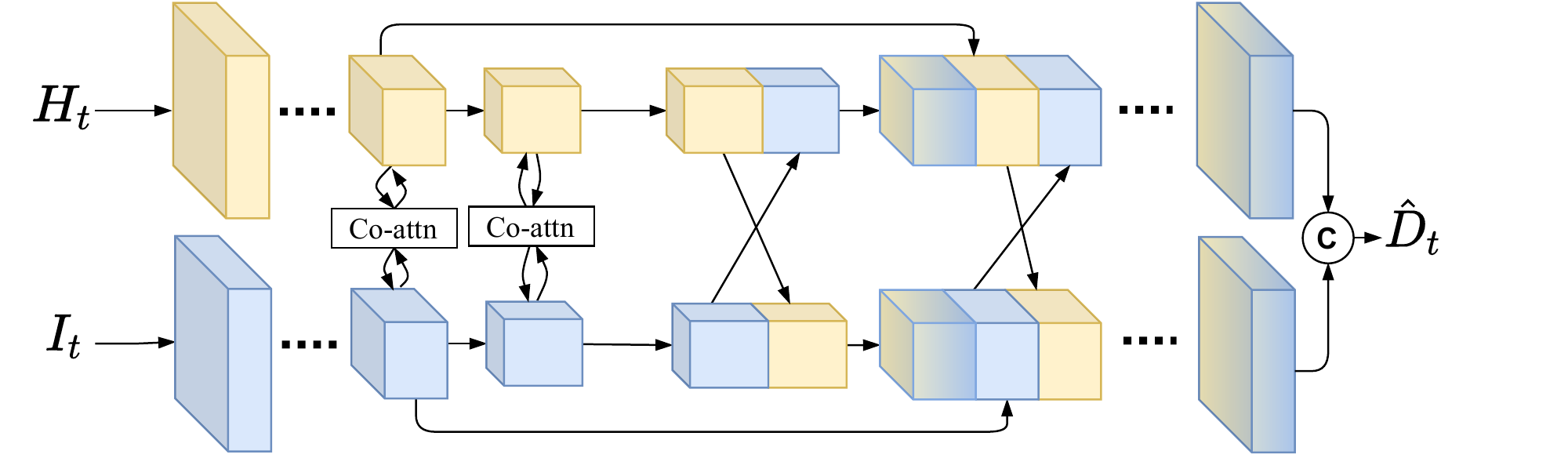}
     \caption{ACMNet~\citep{ACMNet}}
     \label{fig:architectures:ACMNet}
 \end{subfigure}
\caption{\textbf{Depth networks with different image-LiDAR fusion strategies.}
We depict early (a), hybrid (early and late for b) as well as  multiscale (c and d) fusion-based architectures. Volumes in yellow indicate LiDAR feature tensors and blue ones are image feature tensors. We indicate the mixing of modalities with a color grading of the two colors on the volumes. The architecture (c) is our extension of \citep{monodepth2} to make it operate over  \emph{minimal} LiDAR input. We denote the concatenation operator by \circled{c}.
% \PPc{Elaborate a bit. And say here that the extension in (c) is to operate over the additional LiDAR input. Also how the two colors indicate when modalities are pure or mixed.}
}
\label{fig:architectures_no_input}
\end{figure*}
\subsection{Depth network}
\label{sec:learning_system:depth_network}

The core of our depth estimation system is a neural network taking the target image $I_{\tgt}$ coupled with $H_{\tgt}$, the LiDAR data projected in the image plane, as input, and predicting a depth map $\hat{D}_{\tgt}$.
Given the multi-modal nature of the input, our depth network employs a fusion strategy, that can be either early or multi-scale.
In this paper, we consider four different architectures that are illustrated in \autoref{fig:architectures_no_input}.
Three of them are from the recent depth-completion literature, namely NLSPN~\citep{non_local_spatial_propagation}, S2D~\citep{selfsup_sparse_to_dense} and ACMNet~\citep{ACMNet}.
The fourth one, we refer to as \imagesparselidar{}, is an  extension of the camera-only model Monodepth2~\citep{monodepth2} to operate over the additional LiDAR input (we provide details of this extension in~\autoref{app:implem_details}).

The two architectures NLSPN~\citep{non_local_spatial_propagation} and S2D~\citep{selfsup_sparse_to_dense}, illustrated in Figures \ref{fig:architectures:NLSPN} and \ref{fig:architectures:S2D} respectively, employ an early-fusion strategy, combining image and LiDAR features from the start, through concatenation.
Early fusion directly mixes features from both modalities, thus potentially enabling richer interactions across them. 
The NLSPN architecture additionally re-injects the LiDAR signal at the end of the processing, as a late refinement strategy to mitigate signal degradation due to normalization layers.

In contrast, \imagesparselidar{} and ACMNet architectures,  represented in Figure \ref{fig:architectures:IL} and \ref{fig:architectures:ACMNet} respectively, use a multi-scale fusion.
They both encode LiDAR and visual data separately so that these modalities are processed differently and their learned features are progressively integrated together.
This design merges modalities more carefully than the early-fusion strategy, which is desirable as visual and LiDAR inputs carry complementary semantics.
The two encoders, based on ResNet-18~\citep{resnet}, are independent and modality-specific features are fused with a series of concatenations.
ACMNet, on the other hand, employs a more sophisticated co-attention strategy to mutually guide the features in the encoders and mix the features in the decoders to finally fuse them into one prediction.

%%%%%%%%%%%%%%
% SELF-SUPERVISION OBJECTIVES
%%%%%%%%%%%%%
\subsection{Self-supervision objectives}
\label{sec:learning_system:objectives}

Our challenging setting, where depth ground truth is unavailable for training the model, prevents the depth network architecture to be supervised directly.
We address this by training the network under the supervision of two combined objectives.
The first one, photometric reconstruction $L_\text{photo}$, is inspired by recent  advances in self-supervised camera-only monocular depth estimation \citep{sfm_learner,monodepth17,monodepth2}.
However, as discussed in \autoref{sec:rw}, training with this objective alone leads to scale and infinite-depth issues.
Consequently, we leverage a LiDAR self-reconstruction objective, which uses sparse yet complementary LiDAR information to mitigate these issues. 

\parag{Self-supervised photometric reconstruction $L_\text{photo}$.}
We recall that the photometric reconstruction problem is a surrogate task aimed at resynthesizing a target image, given neighboring source images with different viewpoints~\citep{sfm_learner,monodepth2,selfsup_sparse_to_dense}. 
Solutions to this task build on optimization approaches for disparity, motion and depth estimation without learning, based on photo-consistency.
The central idea is to combine pose and depth predictions to project a neighboring source image into the target view. The underlying intuition is that to accurately resynthesize the target view from the source one, both the depth and pose estimation must be accurate.

Formally, the \emph{target} image $I_{\tgt}$ is considered with a set $S$ of \emph{source} images $I_{\src}$ in its temporal vicinity.
First, the depth network predicts the dense depth map $\hat{D}_{\tgt}$ for the target image $I_{\tgt}$.
Second, the relative changes of pose $\hat{P}_{\tgt \shortrightarrow \src}$ between the target and source views are estimated --- we detail this in~\autoref{sec:learning_system:depth_network}.
One pose transformation ${\hat{P}_{\tgt \shortrightarrow \src} = \big( \begin{smallmatrix} \hat{R} & \hat{r} \\ 0 & 1 \end{smallmatrix} \big) \in \text{SE}(3)}$
~is estimated for each source image $I_{\src} \in S$, where $\hat{R}$ is a rotation matrix and $\hat{r}$ the translation component.
Given the estimates of depth and pose, and the camera intrinsics $K$, a source image $I_{\src}$ can be warped via a differentiable geometric transformation into synthetic image $\hat{I}_{\src}$ in the target view. More precisely, for homogeneous coordinates $p_{\tgt}$ of a pixel in the target image, the projected coordinates $p_{\src}$ in the source image are computed with:
\begin{equation}
\label{eq:projection}
    p_{\src} \simeq K \hat{P}_{\tgt \shortrightarrow \src}  \hat{D}_{\tgt}(p_{\tgt}) K^{-1} p_{\tgt}\,.
\end{equation}
For a pair $(I_{\src}, I_{\tgt})$ of source-target images, the reconstructed image $\hat{I}_{\src}$ is enforced to match the target image $I_{\tgt}$ by a pixel-wise image reconstruction error based on both an $L_1$ intensity loss and a structural similarity (SSIM) loss~\citep{ssim}. Note that this formulation assumes Lambertian surfaces.\\

More formally, at a given pixel location $p$, this loss reads:
\begin{equation}
\label{eq:photo_loss}
    L_\text{photo}(p) = \min_{I_{\src} \in S} \Bigl\{ \frac{\alpha}{2}\big(1 - \mathrm{SSIM}(I_{\tgt}, \hat{I}_{\src})(p)\big) + (1-\alpha) \big| I_{\tgt}(p) - \hat{I}_{\src}(p) \big| \Bigr\},
\end{equation}
where $\alpha$ is a hyper-parameter balancing the contributions of the two terms.
Moreover, taking the minimum value over all source images $I_{\src} \in S$ limits the impact of errors resulting from occlusions and disocclusions in the scene due to motion of the ego-car and/or of the other scene elements~\citep{monodepth2}.
To take into account objects with no motion with respect to the ego-car, this loss is only applied to pixels whose appearance between frames varies~\citep{monodepth2}.

%%%%%%%%%%%%%%%%%%%%%%%%%%%
%  LiDAR self-supervision
%%%%%%%%%%%%%%%%%%%%%%%%%%%
\parag{LiDAR self-supervision.}
As detailed in \autoref{sec:rw}, a model solely trained with the photometric reconstruction loss $L_\text{photo}$ suffers from a scale-ambiguity issue and may be affected by the infinite-depth problem. 
In the following, we describe the new role of the low-density input LiDAR as a supervisory signal to mitigate this problem.
We assume that this complementary information source can provide minimal-yet-crucial cues to disambiguate the estimated depth, at a global scale level and especially for moving objects. Furthermore, a sparse depth signal can refine the photometric supervision for small objects, thus improving overall performances~\citep{depth_hints}.
Inspired by the depth completion and the stereo depth estimation literature, we consider three different ways of using LiDAR as a supervisory signal: a straightforward $L_1$ regression along with two refinements that either control the interference with the photometric reconstruction or take into account the inherent noise of the LiDAR signal.

First, we consider a naïve self-supervision scheme, an  $L_1$ loss for all pixels having a LiDAR measurement, in addition to the photometric loss $L_\text{photo}$:
\begin{equation}
    L_\text{naïve}(p)=\begin{cases}
       \lvert\hat{D}_{\tgt}(p) - H_t(p)\rvert + L_\text{photo}(p) & \text{ if } H_\tgt(p) > 0,\\
       L_\text{photo}(p) & \text{ otherwise},\\
    \end{cases}
    \label{eq:naive}
\end{equation}  
where $p$ is an index over the pixels, $\hat{D}_{\tgt}$ the estimated depth and $H_\tgt$ the input LiDAR projected in the target image plane. The latter being sparse, not all pixels have LiDAR data available; we use the encoding $H_{\tgt}(p) = 0$ for such pixels.

Second, we consider the \emph{masked} self-supervised objective proposed in~\citet{selfsup_sparse_to_dense}. It makes the LiDAR regression and the photometric loss exclusive by masking-out the photometric loss $L_\text{photo}$ on pixels with a LiDAR measurement. Denoting $L_\text{masked}$ as this loss, it is given by:
\begin{equation}
    L_\text{masked}(p)=\begin{cases}
       \lvert\hat{D}_{\tgt}(p) - H_\tgt(p)\rvert & \text{ if } H_\tgt(p) > 0,\\
       L_\text{photo}(p) & \text{ otherwise.}\\
    \end{cases}
    \label{eq:masked}
\end{equation}
This loss is similar to $L_\text{naïve}$ but avoids
potential conflicts between
the photometric and LiDAR reconstructions.

Lastly, inspired by \citet{depth_hints}, we also introduce the \emph{hinted} self supervision, $L_\text{hinted}$, that takes into account  the inherent noise of the LiDAR signal.
Despite being a direct depth measurement, raw LiDAR signal is noisy for a number of reasons, including potentially imprecise calibration, approximated projection, and the fact that the camera and LiDAR are not exactly positioned at the same place, which results in objects observable by one but hidden to the other. 
Therefore, the loss $L_\text{hinted}$ integrates the LiDAR self-supervision only where image reconstruction is more precise by using the LiDAR signal instead of the estimated depth.
More precisely, two versions of the photometric contribution of the pixel are computed: the regular pixel-wise photometric loss $L_\text{photo}$, using the estimated depth map $\hat{D}_{\tgt}$ in \autoref{eq:projection}, and ${L}^H_\text{photo}$ using the input projected LiDAR $H_\tgt$ instead of $\hat{D}_{\tgt}$ in \autoref{eq:projection}. Then we only supervise with the LiDAR reconstruction when ${L}^H_\text{photo} < L_\text{photo}$.
The objective is thus:
\begin{equation}
    L_\text{hinted}(p)=\begin{cases}
       \lvert\hat{D}_{\tgt}(p) - H_t(p)\rvert + L_\text{photo}(p) & \text{ if } L^H_\text{photo}(p) < L_\text{photo}(p)\\
       L_\text{photo}(p) & \text{ otherwise.}\\
    \end{cases}
    \label{eq:hinted}
\end{equation}

%%%%%%%%%%%%%%
% POSE ESTIMATION
%%%%%%%%%%%%%
\subsection{Pose estimation}
\label{sec:learning_system:pose_setups}

The formulation of the photometric reconstruction involves the change of pose $\hat{P}_{\tgt \shortrightarrow \src}$ between the target image $I_{\tgt}$ and source view $I_{\src}$ for the source image warping.
A first possibility, which is widely used in monocular self-supervised depth estimation~\citep{sfm_learner,monodepth2, packnet,struct2depth}, uses a so-called \emph{pose network} jointly trained with the depth network.
However, due to the monocular ambiguity, this approach can only estimate a relative pose and thus relative depth maps, which then must be rescaled by an unknown factor.
Instead, we explore another way to estimates a metric pose, by leveraging the LiDAR information and solving a Perspective-$n$-Point problem~\citep{EPnP,P3P}. As such, depth estimation should also align to a real-world scaling.

\parag{Perspective-$n$-Point (\pnp{}).}
The \pnp{} problem originally seeks the absolute pose of a camera given a set of 3D points and their corresponding 2D image projections.
In our case, we use the \pnp{} formulation to estimate the change of pose between the target and source views, \ie given the target image $I_\tgt$ and LiDAR measurements, as well as the source image $I_\src$.

First, pairs of pixels $(p_\tgt, p_\src)$ matching in both views $I_\tgt$ and $I_\src$ are found using the SIFT descriptor~\citep{sift_Lowe} based on a DoG keypoint detector.
Then, the sole pairs for which $p_\tgt$ has a LiDAR measurement are considered.
This gives us the pairs of 3D-2D points, where points $p_\tgt$ 
are complemented with depth measurements and match the 2D points $p_\src$ of the source image $I_\src$.
Given these pairs, we can precisely estimate the \emph{metric}-scaled 6D rigid transformation between the target and source poses by minimizing the cumulative projection error.

In challenging real-life situations, and especially when dealing with a 4-beam LiDAR, finding matching pixels that have LiDAR measurements can be arduous, making this method prone to errors.
Hence, we follow \citet{selfsup_sparse_to_dense} to remove outliers in the set of point correspondences by using RANSAC in conjunction with the \pnp{} solving algorithm. 
When this filtering step is insufficient for the algorithm solving the \pnp{} problem to converge, we discard the training sample.
\section{Experimental protocol}
\label{sec:protocol}

The first component of our protocol is the dataset used for the experiments, namely KITTI~\citep{kitti}, and our preprocessing to reduce the raw 64-beam LiDAR to a 4-beam one (\autoref{sec:protocol:dataset}). We then introduce baselines in \autoref{sec:protocol:external_baselines}.
Additional details are given in the appendix. 

\subsection{Dataset and evaluation metrics}
\label{sec:protocol:dataset}

To train models in our LiDARTouch framework, we need a dataset that provides a camera stream with aligned sparse LiDAR data for training. We also require this dataset to have ground-truth depth data with an associated benchmark to assess and compare our test performances.
We are aware of only one dataset matching both of these requirements, namely KITTI. 
It contains 1.5 hours of recorded driving sessions in urban environment from a video stream synchronized with LiDAR data. Depth ground truth is available: it is derived from dense LiDAR signals accumulated over five sweeps and stereo filtered.
Overall, we use this dataset to train and evaluate the quality of the predictions of our framework, and to compare against baselines and variants.
On the KITTI dataset~\citep{kitti}, we use the so-called Eigen split~\citep{eigen} for train, val and test with a minor modification for the val and test. The ground-truth LiDAR of~\citet{sparsity_invariant_cnns} is not available for some of the frames of the Eigen splits (fewer than 10). Following common practice~\citep{monodepth2,packnet}, we removed them from the val and test splits. Thus, the total number of examples are 22537, 873 and 652 respectively for the train, val and test sets.

The LiDAR data provided in KITTI is obtained with high-end 64-beam sensors, appropriate for \emph{evaluating} our self-supervised models, but much denser than what is expected to \emph{train} our LiDARTouch framework. 
Consequently, we perform a filtering step to extract 4 beams out of the raw 64-beam LiDAR data. To conform with prior works~\citep{selfsup_sparse_to_dense,jaritz_sparse_and_dense, packnet-semisup} and better compare with them, we sample LiDAR beams uniformly: 1 beam is kept every 16.
Note that with such a sampling, while 4 beams are extracted, only three beams effectively project onto the image plane as one beam falls out of the considered visual region.

\parag{Evaluation metrics.}
Evaluation is conducted against accumulated ground-truth LiDAR obtained following~\citet{sparsity_invariant_cnns}, with the metrics defined in~\citet{eigen}. 
This includes the absolute (Abs Rel) and square (Sq Rel) relative errors, the root mean square error (RMSE), and its log version (RMSE$_{\log}$), as well as precision-under-threshold metrics measuring the percentage of depth predictions $\hat{D}$ close enough to the ground-truth depth $D$, in the sense of the value ${\delta := \text{max}(\frac{\hat{D}}{D}, \frac{D}{\hat{D}})}$ being under a user-defined threshold.
Following~\citet{eigen}, we consider three thresholds: $\delta < 1.25$, $\delta < 1.25^2$ and $\delta < 1.25^3$.

\subsection{Notations, ablations and external baselines}
\label{sec:protocol:external_baselines}

\parag{Notations.}
To refer to the network architecture, independently of the rest of the learning framework, we use Monodepth2, Monodepth2-L, NLSPN, ACMNet and S2D.
When we refer to whole models, i.e., architectures trained under the LiDARTouch framework, we append the `LiDARTouch' prefix. For example, we note `\ssACMNet' when we adapt the ACMNet architecture into the LiDARTouch framework.

For clarity, the inputs and the supervision schemes that are employed by the models are recalled in the tables of the experiments section.
The input of each depth prediction model includes an image (noted `$\mathcal{I}$') and, optionally, a sparse 4-beam LiDAR point cloud (`$\mathcal{L}^4$').
We considered the following supervisions strategies: self-supervised photometric reconstruction (`P') associated to loss \autoref{eq:photo_loss}, supervised LiDAR ground-truth regression with $L_1$ loss (`L$_\text{gt}$'), or LiDAR self-supervision (`L$_{4}$') with one of the three options in Eqs.\,(\ref{eq:naive}), (\ref{eq:masked}), or (\ref{eq:hinted}).

\parag{Ablation: Pose estimation with a pose network.}
\label{sec:protocol:pose_baselines}
In \autoref{sec:learning_system:pose_setups}, we presented the \pnp{} algorithm, which estimates metric pose changes from source to target views.
To highlight the gains enabled by the use of the extra LiDAR information for computing the pose, we experiment by training a \emph{pose network} instead, a widely used component of monocular depth estimation models~\citep{sfm_learner,monodepth2, packnet,struct2depth}.
For each target-source image pair, the pose network outputs the 6D rigid transformation between views.
It is differentiable and trained jointly with the depth network. 
When only trained with the photometric error (\autoref{eq:photo_loss}), the 6D transformation is estimated up to a scale factor due to the monocular ambiguity. This results in a relative depth estimation requiring to be rescaled by the LiDAR depth ground-truth median value (not available in our case).

A solution is to use data from the IMU/GNSS to supervise the pose estimation scale. In the context of depth estimation, such an approach has been explored by \citet{packnet}.
Formally, we first obtain the approximate change in pose between the source and target views ($P_{\tgt \shortrightarrow \src}$) from integrated inertial measurements. Then, we extract its translation component $r$ and make the predicted pose translation component $\hat{r}$ regress its magnitude:
\begin{equation}
\label{eq:velocity}
    L_\text{imu} = \Bigl\vert \lVert r\rVert_2  - \lVert\hat{r}\rVert_2 \Bigr\vert.
\end{equation}
As for a given pose there is a unique depth minimizing \autoref{eq:photo_loss}, constraining the pose's magnitude to a metric scale forces the depth estimation to be metric as well.

\parag{Baselines: Monocular methods.}
We compare against state-of-the-art monocular self-supervised approaches such as SfMLearner \citep{sfm_learner}, Vid2Depth \citep{Vid2Depth}, GeoNet \citep{GeoNet}, DDVO \citep{DDVO}, Monodepth2 \citep{monodepth2} and PackNet-SfM \citep{packnet}. Note that these methods can only produce relative depth maps, as they use an unsupervised pose network, so they have to be rescaled using the ground-truth LiDAR. Comparisons with these methods is thus unfair, in their favor.

Additionally, we compare with methods that directly produce metric depth by leveraging additional supervision. This includes (1) DORN~\citep{DORN}, a camera-only method fully-supervised by a dense LiDAR signal, (2) \citet{Kuznietsov}, a semi-supervised method using stereo reconstruction and dense LiDAR supervision, and (3) PackNet-SfM~\citep{packnet} model supervised with IMU prior. 

\parag{Baselines: Depth completion methods.}
We also compare against supervised depth completion methods, namely ACMNet~\citep{ACMNet}, NLSPN~\citep{non_local_spatial_propagation} and S2D~\citep{selfsup_sparse_to_dense}.
However, their original versions are not trained and evaluated on the same splits as monocular methods. We re-train and evaluate them on the Eigen split, in their fully-supervised setting but with only a 4-beam LiDAR input.
Additionally, we also train and evaluate these depth completion methods when the depth ground truth is simply replaced by the 4-beam LiDAR input for supervision signal. We refer to this setting as `Naïve self-sup.'.

%%%%%%%%%%%%%%%%%%%%%%%%%%%%%%%%%%%%%%%%%%%%%%%%%%%%%%%%%%%%%%%%%%%%%%%%%%%%%
%%%%%%%%%%%%%%%%%%%%%%%%%%%%%%%%%%%%%%%%%%%%%%%%%%%%%%%%%%%%%%%%%%%%%%%%%%%%%
%%%%%%%%%%%%%%%% Contributions brought by a touch of LiDAR %%%%%%%%%%%%%%%%%% 
%%%%%%%%%%%%%%%%%%%%%%%%%%%%%%%%%%%%%%%%%%%%%%%%%%%%%%%%%%%%%%%%%%%%%%%%%%%%%
%%%%%%%%%%%%%%%%%%%%%%%%%%%%%%%%%%%%%%%%%%%%%%%%%%%%%%%%%%%%%%%%%%%%%%%%%%%%%

%{\color{RawSienna}
\section{Influence of a touch of LiDAR}
\label{sec:lidar_influence}

In this section, we validate setups where the depth network converges to a metric scale. 
In particular, in \autoref{sec:lidar_influence:ablative_study}, we disentangle the contributions brought by LiDAR with an ablation study on the three levels of integration presented in \autoref{sec:learning_system}: as a self-supervision signal, as a depth network’s input, and as additional information for pose estimation.
We also investigate various combinations of LiDAR self-supervision schemes and depth networks in \autoref{sec:lidar_influence:variants}.

%%%%%%%%%%%%%%%%%%%%%
% ABLATIVE STUDY
%%%%%%%%%%%%%%%%%%%%%
\subsection{Ablation of LiDAR}
\label{sec:lidar_influence:ablative_study}

We begin with an ablation study to assess the contribution brought by sparse LiDAR at three different levels: supervision, input and pose. We define our LiDARTouch framework as using a \pnp{} for pose estimation, LiDAR self-supervision ($\text{L}_4$) with the \emph{masked} loss variant, and a bi-modal depth network (i.e., taking RGB and LiDAR as input). Models that belong to this framework are highlighted as light blue cells in \autoref{tab:scaling}.
For the sake of clarity, in this section we focus on the leftmost three columns for direct comparison with LiDARTouch. Other learning setups are discussed in detail in \autoref{app:ablation}.

%%%%%%%%%%%%%%%% LIDAR IN INPUT
\parag{LiDAR as an input.}
First, we study the contribution brought by LiDAR when it is used as an input to the depth network in addition to the image signal. 
Results in the first column of~\autoref{tab:scaling} show that the \imagealone{} architecture, which does not use LiDAR as input, is consistently outperformed by all the other bi-modal architectures leveraging LiDAR input.
These architectures achieve a relative improvement of 11-13\% compared to \imagealone{}. This validates the positive influence of integrating few-beam LiDAR as an input.

%%%%%%%%%%%%%%%% LIDAR IN SUPERVISION
\parag{Self-supervision with the sparse LiDAR.}
Next, we study the impact of using a 4-beam LiDAR as a self-supervisory signal by removing it from the LiDARTouch framework, which leaves only the photometric loss (`P'). This corresponds to the second column in \autoref{tab:scaling}. 
Overall, the results support our claim that the use of LiDAR self-supervision improves or is similar in performance with respect to the photometric-only supervision schemes. 

Although ACMNet, NLSPN and S2D architectures show slightly better performance when trained with P\textit{n}P and the photometric-only loss, \ie without any LiDAR self-supervision, they are severely affected by the infinite-depth issue (see \autoref{sec:infinite_depth}).

Moreover, when using the photometric loss alone (`P' in the table), \imagealone{} and \imagesparselidar{} are hard to train. Indeed, while \pnp{} pose is metric by construction, the depth network is initialized randomly and has to converge to a metric scale with the photometric reconstruction as the sole learning signal. Without any precaution, we observe large numerical differences in scale at initialization between the pose and depth, which provoke unstable training for the depth network. 
To address this instability, we divide the translation component of the \pnp{} pose by a factor $\alpha$ during training and multiply the depth prediction consequently at inference (details in \autoref{app:imu_prior}). This procedure is inspired by the baseline scaling introduced in Monodepth2 for the stereo setting~\citep{monodepth2}. We indicate models that need to be trained using this strategy with `$\ast$' in \autoref{tab:scaling}.
On the other hand, under the LiDARTouch framework, all depth networks train well without requiring training tricks.

\definecolor{Mycolor2}{HTML}{c1eaf5}

\begin{table}[]
\setlength\tabcolsep{0pt} % make LaTeX figure out width of inter-column spaces
\caption{\textbf{Pose estimation ablation}.
We report precision (\%) under threshold ($\delta<1.25$) on the KITTI test split; higher is better. 
% \rule[raise]{width}{thickness}
As we are interested in \emph{metric} depth estimations, contrary to common practice~\citep{monodepth2, packnet}, estimations are \textbf{not} rescaled with LiDAR GT. Light blue cells indicate configurations corresponding to our LiDARTouch framework. Some models are more difficult to train and indicated in light grey cells. In particular, `$\ast$' implies that a rescaling of the pose was used for a stable training, and `$\dagger$' indicates that the LiDAR signal had to be dilated to avoid overfitting to the LiDAR input (more details in \ref{app:dilated}). When using a pose network with photometric supervision only (dark gray cells), the estimation can only be \emph{relative} and the scores are all below 1\%. More details are provided in \autoref{sec:lidar_influence:ablative_study}.
}
\vspace{0.1cm}
\resizebox{\columnwidth}{!}{%
\begin{tabular}{@{\extracolsep{2pt}}r@{\hskip 4pt}ccc@{\hskip 4pt}cccc@{}}
\toprule
 & & \multicolumn{2}{c}{w/ P$n$P} & \multicolumn{4}{c}{w/ pose network} \\
\cmidrule{3-4} \cmidrule{5-8}
%\textbf{Network}             &         & imu     & lidar   & lidar+imu &       & lidar \\ \midrule
\textbf{Depth network} & & \parbox{1cm}{\centering P+L$_4$} & \parbox{1cm}{\centering P} & \parbox{1cm}{\centering P+L$_4$} & \parbox{1cm}{\centering P} & \parbox{1cm}{\centering P+imu}  & \parbox{1cm}{P+L$_4$\\+imu} 
\\
\midrule
\imagealone         &  &\ok{86.1} & \degenerate{}{86.2}$^{\ast}$ & \ok{83.9} & \relative{ - } & \ok{86.2} & \ok{86.5}  \\
\imagesparselidar   & \multirow{4}*{\rotatebox[origin=c]{90}{\parbox{1.8cm}{\textcolor{SkyBlue}{LiDARTouch}}}} 
                    &\lidartouch{96.9} & \degenerate{}{96.2}$^{\ast}$ & \ok{94.9} & \relative{ - } & \ok{96.4} & \ok{96.5}  \\
%\lidartouch{\multirow{4}*{\rotatebox[origin=c]{90}{\parbox{1.92cm}{\sidetitle{LiDARTouch}}}}} \\
%
ACMNet              & &\lidartouch{97.4} & \ok{97.5} & \overfit{91.2}$^{\dagger}$ & \relative{ - } & \ok{27.0} & \ok{95.3} \\
NLSPN               & &\lidartouch{95.9} & \ok{96.8} & \overfit{94.2}$^{\dagger}$ & \relative{ - } & \ok{38.5} & \overfit{94.1}$^{\dagger}$ \\
S2D                 & &\lidartouch{96.2} & \ok{96.4} & \overfit{93.9}$^{\dagger}$ & \relative{ - } & \ok{28.7} & \overfit{94.0}$^{\dagger}$ \\ \bottomrule
\end{tabular}
}
\label{tab:scaling}
\vspace{0.3cm}
\end{table}

\parag{Pose estimation with a sparse LiDAR.}
We now show that a precise computation of the change of pose is critical to estimate depth maps that are correctly scaled, and that a touch of LiDAR is beneficial for this purpose. 
To demonstrate this, we experiment by replacing \pnp{} in our LiDARTouch setup with a pose network that does not use any LiDAR information, as detailed in \autoref{sec:protocol:pose_baselines}. This ablation of LiDARTouch corresponds to the third the column, `P+L$_4$' under `w/ pose network', in \autoref{tab:scaling}.

The main difference between these two setups is that  \pnp{} methods produce \emph{metric} poses by construction, which leads to the collapse of depth solutions to metric depths. In opposition, the use of a pose network requires a joint alignment and convergence to a metric scale between the depth and pose networks as they are both randomly initialized.
While \imagesparselidar{} achieves this, it can be observed that the use of a pose network instead of \pnp{} degrades performance up to 6\% when compared to LiDARTouch.
Above all, we observe a tendency for ACMNet, NLSPN and S2D to overfit the LiDAR signal (see \autoref{fig:overfit:pred} for an example). 

We find that the multi-scale prediction and supervision during training of \imagealone{} and \imagesparselidar{} are key for the models not to overfit the sparse 4-beam LiDAR data. Indeed, supervision at the lowest scale (1:8) increases the number of pixels getting supervision from LiDAR as pixels with associated LiDAR signal are expanded due to the difference in scale.

Building on this observation, we propose a procedure to simulate this behavior in order to avoid LiDAR overfitting for mono-scale networks without changing their architectures.
To simulate a LiDAR self-supervision at a lower scale, we apply a \emph{dilation} morphological operation on the 4-beam LiDAR at the supervision level. 
This artificially increases the number of pixels receiving LiDAR supervision, albeit in a noisy manner, and enables the mono-scale depth networks ACMNet, NLSPN and S2D to produce globally coherent metric depth estimations.
We report results of models trained with this procedure (indicated by `$\dagger$') in~\autoref{tab:scaling} and provide technical details as well as qualitative examples in~\autoref{app:dilated}.

On the other hand, training under our LiDARTouch framework eliminates the need for such tricks. Indeed, results demonstrate that our LiDARTouch framework, using LiDAR as self-supervision, in input and in pose computation yields competitive performances for all the five architectures, a more stable training compared to any other configuration, and alleviates the infinite-depth problem as we will show in~\autoref{sec:infinite_depth}.

%%%%%%%%%%%%%%%%%%%%%
% LiDAR LOSS VARIANTS
%%%%%%%%%%%%%%%%%%%%%
\begin{table}[]
\addtolength{\tabcolsep}{-2pt}
\begin{center}
%\caption{\textbf{Influence of the LiDAR self-supervision for various depth prediction networks.} RMSE metric (lower is better) on the Eigen test split of KITTI, with photometric self-supervision alone (P), or in conjunction with one of the three considered variants of minimal-LiDAR self-supervision (L$_4$). All models are trained with P$n$P for pose estimation.}
\caption{\textbf{Variants comparison of the LiDAR self-supervision.} RMSE metric (lower is better) on the Eigen test split of KITTI. Models are trained with photometric self-supervision (P) in conjunction with one of the three considered variants of minimal-LiDAR self-supervision (L$_4$). All models are trained with P$n$P for pose estimation.}
%\resizebox{0.9\columnwidth}{!}{%
\vspace{15pt}\begin{tabular}{@{}lrrrrrr@{}}
%\toprule
%\parbox{1.3cm}{\raggedleft \textbf{Self-superv}} & \imagesparselidar & ACMNet & NLSPN & S2D \\
%\rotatebox{90}{\parbox{2mm}{\multirow{3}{*}{rota}}}
\textbf{Self-supervision} & \rotatebox{30}{Monodepth2\hspace{-0.8cm}} & \rotatebox{30}{\imagesparselidar\hspace{-1.1cm}} & \rotatebox{30}{ACMNet\hspace{-0.8cm}} & \rotatebox{30}{NLSPN\hspace{-0.8cm}} & \rotatebox{30}{S2D} \\
\toprule
%\textbf{Self-supervision} & \imagesparselidar & ACMNet & NLSPN & S2D \\
%P & 65.500 & 24.750 & 2.464 & 2.707 & 2.803 \\
%\midrule
P + L$_4$(naïve)  & 4.504       & 2.796                         & 2.490     & 3.084    &  2.839    \\
P + L$_4$(hinted)  & 4.794       & 2.813                         & 2.563     & 3.271    &  2.982    \\
P + L$_4$(masked)  & 4.517       & 2.696                         & 2.504     & 3.014    &  2.776    \\
\bottomrule
\end{tabular}
%}
\label{tab:lidar_losses}
\end{center}
\end{table}

% supervisions
\newcommand{\tselfsup}{M}
\newcommand{\tsup}{D}
\newcommand{\tvel}{M+v}
\newcommand{\tlidar}{M+l}

\begin{table*}[]
\addtolength{\tabcolsep}{-2pt}
\centering
\caption{
\textbf{Comparison against monocular depth estimation methods.}
Results are reported on the KITTI Eigen split~\citep{eigen} with improved ground truth~\citep{sparsity_invariant_cnns}.
% gt rescaling
A few self-supervised methods produce relative-depth maps and their prediction must be rescaled using ground-truth information; this is identified by `\textit{gt rescaled}' in the table. 
% pretraining
Some of the methods also benefit from an extra pre-training, on ImageNet~\citep{ImageNet} or Cityscapes~\citep{cityscape}, denoted with $\circ$ or $\star$ superscripts, respectively.
% reimplem
The model \emph{\imagealone} in italic indicates our re-implementation of \citep{monodepth2} without pre-training and post-processing.
% Notation
Input includes the image only (`$\mathcal{I}$'), or combined with the few-beam LiDAR point cloud (`$\mathcal{L}^4$').
Supervision includes photometric loss (`P'), IMU prior (`imu'), stereo reconstruction (`ste') and LiDAR supervision with either dense ground truth (`L$_{\text{gt}}$') or sparse 4-beam LiDAR (`L$_4$'). 
\label{tab:sota}
}
\resizebox{\textwidth}{!}{%
\begin{tabular}{@{}c @{\hspace{0.1cm}} c @{\hspace{0.1cm}} l l l *{7}{c}@{}}
%\Xhline{1pt}
\toprule
& & \textbf{Method} & Input & Superv. & \fzz{Abs Rel\,$\downarrow$} & \fzz{Sq Rel\,$\downarrow$} & \fzz{RMSE\,$\downarrow$} & \fzz{RMSE$_{\log}$\,$\downarrow$} & \fzz{$\delta<1.25$\,$\uparrow$} & \fzz{$\delta<1.25^2$\,$\uparrow$} & \fzz{$\delta<1.25^3$\,$\uparrow$} \\
%\Xhline{1pt}
\hline %\toprule
\multirow{2}*{\rotatebox[origin=c]{90}{\textbf{Sup.}}}
& & DORN$^\circ$~\citep{DORN} & $\mathcal{I}$ & L$_\text{gt}$ & 0.072 & 0.307 & 2.727 & 0.120 & 0.932 & 0.984 & 0.995\\  % kuznietsov
& & Kuznietsov et al.$^\circ$~\citep{Kuznietsov} & $\mathcal{I}$ & L$_\text{gt}$+ste & 0.089 & 0.478 & 3.610 & 0.138 & 0.906 & 0.980 & 0.995\\  % kuznietsov
\hline
\multirow{15}*{\rotatebox[origin=c]{90}{\textbf{Self-supervised}}} & \multirow{7}*{\rotatebox[origin=c]{90}{\textit{gt rescaled}}}
%\multirow{8}*{\rotatebox[origin=c]{90}{g-t rescaled}}
& SfMLearner$^\star$~\citep{sfm_learner} & $\mathcal{I}$ & P & 0.176 & 1.532 & 6.129 & 0.244 & 0.758 & 0.921 & 0.971\\  % third
& & Vid2Depth$^\star$~\citep{Vid2Depth} & $\mathcal{I}$ & P & 0.134 & 0.983 & 5.501 & 0.203 & 0.827 & 0.944 & 0.981\\  % third
& & GeoNet$^\star$~\citep{GeoNet} & $\mathcal{I}$ & P & 0.132 & 0.994 & 5.240 & 0.193 & 0.883 & 0.953 & 0.985\\  % third
& & DDVO~\citep{DDVO} & $\mathcal{I}$ & P & 0.126 & 0.866 & 4.932 & 0.185 & 0.851 & 0.958 & 0.986\\  % half
& & \emph{\imagealone} (our reimplem.) & $\mathcal{I}$ & P & 0.099 & 0.591 & 4.030 & 0.149 & 0.897 & 0.976 & 0.993 \\ 
& & Monodepth2$^\circ$~\citep{monodepth2} & $\mathcal{I}$ & P & 0.090 & 0.545 & 3.942 & 0.137 & 0.914 & 0.983 & 0.995\\  % half
& & PackNet-SfM$^\star$~\citep{packnet} & $\mathcal{I}$ & P & 0.071 & 0.359 & 3.153 & 0.109 & 0.944 & 0.990 & 0.997\\  %full
\cline{2-12}
& & \emph{\imagealone} w/ IMU supervision & $\mathcal{I}$ & P+imu & 0.110 & 0.729 & 4.565 & 0.172 & 0.862 & 0.965 & 0.989 \\
& & PackNet-SfM$^\star$ \citep{packnet} & $\mathcal{I}$ & P+imu & 0.075 & 0.384 & 3.293 & 0.114 & 0.938 & 0.984 & 0.995 \\
& & \ssSAN &  $\mathcal{I}$+$\mathcal{L}^4$  & P+L$_4$ & 0.063 & 0.396 & 3.318 & 0.118 & 0.946 & 0.982 & 0.993 \\
& & \ssNLSPN &  $\mathcal{I}$+$\mathcal{L}^4$ & P+L$_4$ & 0.053 & 0.336 & 3.013 & 0.106 & 0.959 & 0.987 & 0.994 \\
& & \ssSD &  $\mathcal{I}$+$\mathcal{L}^4$  & P+L$_4$ & 0.059 & 0.285 & 2.776 & 0.102 & 0.962 & 0.988 & 0.995 \\
& & \ssmonodepthtwol & $\mathcal{I}$+$\mathcal{L}^4$ & P+L$_4$ & 0.047 & 0.267 & 2.696 & 0.090 & 0.969 & \underline{0.991} & \underline{0.996} \\ % full
& & \ssACMNet &  $\mathcal{I}$+$\mathcal{L}^4$  & P+L$_4$ & \underline{0.044} & \underline{0.242} & \underline{2.504} & \underline{0.086} & \underline{0.974} & \underline{0.991} & \underline{0.996} \\
\bottomrule
\end{tabular}
}
\end{table*}

%\begin{table*}[]
%    \centering
%    \begin{tabular}{@{}l l l *{7}{c}@{}}
%        \toprule
%        \textbf{Model} & Input & Superv. & \fzz{Abs Rel\,$\downarrow$} & \fzz{Sq Rel\,$\downarrow$} & \fzz{RMSE\,$\downarrow$} & \fzz{RMSE$_{\log}$\,$\downarrow$} & \fzz{$\delta<1.25$\,$\uparrow$} & \fzz{$\delta<1.25^2$\,$\uparrow$} & \fzz{$\delta<1.25^3$\,$\uparrow$} \\
%        \hline
%        S2D~\citep{selfsup_sparse_to_dense} & $\mathcal{I}$+$\mathcal{L}^4$ & \textit{P+L$_4$} & - & - & $>$4.0 & - & - & - & - \\
%        \ssSD & $\mathcal{I}$+$\mathcal{L}^4$  & \textit{P+L$_4$} & 0.059 & 0.285 & 2.776 & 0.102 & 0.962 & 0.988 & 0.995 \\
%        \bottomrule
%    \end{tabular}
%    \caption{\textbf{Comparison with S2D~\citep{selfsup_sparse_to_dense}.}
%    Scores of S2D~\citep{selfsup_sparse_to_dense} are reported from the Figure~6b of their original paper.
%    \label{tab:s2d_comparison}
%    }
%\end{table*}
\subsection{LiDAR self-supervision variants}
\label{sec:lidar_influence:variants}

We compare in \autoref{tab:lidar_losses} the variants for the LiDAR loss defined in \autoref{sec:learning_system:objectives}, namely the \emph{naïve} compound loss \autoref{eq:naive}, the \emph{masked} one \autoref{eq:masked}, which prevents interferences with the photometric error, and the \emph{hinted} loss \autoref{eq:hinted}, which handles the noise of the LiDAR signal.
These experiments are conducted for the four different depth networks considered in \autoref{sec:learning_system:depth_network}.
Overall, averaged over all architectures, the \emph{masked} version of the LiDAR loss achieves the best results, demonstrating the need to reduce interferences between the LiDAR and photometric supervisions.
On the other hand, we observe that the \emph{hinted} loss yields the worst results. We expected the \emph{naïve} loss to have the worst performance as it does not consider the noise in LiDAR, but it appears that the control the \emph{hinted} loss imposes is too strong and discards too many of the already scarce LiDAR measurements.
Hence, it confirms that the \emph{masked} LiDAR self-supervision is the most effective.

%%%%%%%%%%%%%%%%%%%%%%%%%%%%%%%%%%%%%%%%%%%%%%%%%%%%%%
%%%%%%%%%%%%%%%%%%%%%%%%%%%%%%%%%%%%%%%%%%%%%%%%%%%%%%
%%%%%%%%%%%%%%%% EXTERNAL BASELINES %%%%%%%%%%%%%%%%%% 
%%%%%%%%%%%%%%%%%%%%%%%%%%%%%%%%%%%%%%%%%%%%%%%%%%%%%%
%%%%%%%%%%%%%%%%%%%%%%%%%%%%%%%%%%%%%%%%%%%%%%%%%%%%%%

\section{Comparison against related works}
\label{sec:external_basleines}

In \autoref{tab:sota}, we report evaluations of the four architectures presented in \autoref{sec:learning_system:depth_network}, trained within our LiDARTouch framework against camera-only baselines.

\parag{Self-supervised camera-only methods.}
First, we show that training under our framework outperforms self-supervised monocular depth estimation methods~\citep{sfm_learner,Vid2Depth,GeoNet,DDVO,monodepth2,packnet}. We note that contrary to other methods, ours uses few-beam LiDAR as input. Furthermore, self-supervised monocular depth estimation approaches only estimate relative depth and thus are rescaled with ground truth before evaluation. With our approach, this unrealistic and impractical rescaling step is no longer needed. 

\parag{Supervised camera-only methods.}
We also obtain better results than monocular depth estimation models trained with ground truth and optional stereo \citep{DORN,Kuznietsov}, while not requiring either of those. While the latter does not use few-beam LiDAR as input, not requiring ground truth at train time makes our method trainable at scale on any domain.
 
Overall, we showed that by integrating few-beam LiDAR in the pipeline, we substantially increase performances on all metrics over other methods not using few-beam LiDAR.

\newcommand{\sidetitle}[1]{\textbf{\fontsize{7}{0}{\selectfont #1}}}

%%%%%%%%%%%%%%%%%%%%%%% 4 beams

\begin{table}[ht!]
\addtolength{\tabcolsep}{-2pt}
\centering

\caption{\textbf{Comparison against supervised and naively self-supervised depth completion schemes.} Input includes the image and the 4-beam LiDAR ($\mathcal{I}$+$\mathcal{L}^4$)}

\resizebox{\linewidth}{!}{
\begin{tabular}{@{}crccccc@{}}
\toprule
& \textbf{Network} & Superv. & \fzz{Abs Rel\,$\downarrow$} & \fzz{Sq Rel\,$\downarrow$} & \fzz{RMSE\,$\downarrow$} & \fzz{$\delta<1.25$\,$\uparrow$}  \\

\hline

%%%% GT SUP
\multirow{4}*{\rotatebox[origin=c]{90}{\sidetitle{GT sup.}}}
& ACMNet & L$_\text{gt}$ &  0.030 & 0.143 & 2.112 & 0.983 \\
& NLSPN & L$_\text{gt}$ & 0.044 & 0.214 & 2.617 & 0.971 \\
& S2D & L$_\text{gt}$ & 0.035 & 0.152 & 2.271 & 0.979 \\
& SAN &  L$_\text{gt}$ & 0.037 & 0.172 & 2.491 & 0.976 \\
\hline

%%%% NAIVE SELF-SUP
\multirow{4}*{\rotatebox[origin=c]{90}{\parbox{1.3cm}{\centering \sidetitle{Naive\\self-sup.}}}}
& ACMNet  & L$_4$ &  0.714 & 9.751 & 15.88 & 0.057 \\
& NLSPN & L$_4$ & 4.133 & 268.4 & 51.96 & 0.010 \\
& S2D & L$_4$ &  0.849 & 12.84 & 17.53 & 0.077 \\
& SAN &  L$_4$ & 0.426 & 6.226 & 14.148 & 0.243 \\
\hline

%%%% LiDARTouch
\multirow{4}*{\rotatebox[origin=c]{90}{\parbox{1.4cm}{\sidetitle{LiDARTouch}}}}
& ACMNet & P+L$_4$ &  0.044 & 0.242 & 2.504 & 0.974 \\
& NLSPN & P+L$_4$ & 0.053 & 0.336 & 3.013 & 0.959 \\
& S2D & P+L$_4$ & 0.059 & 0.285 & 2.776 & 0.962 \\
& SAN &  P+L$_4$ & 0.063 & 0.396 & 3.318 & 0.946 \\
\bottomrule
\end{tabular}
}
\label{tab:adapted_depth_completion:4beams}

\end{table}

We compare our LiDARTouch framework against two supervision schemes from the depth completion literature: full-supervision with ground truth (L$_\text{gt}$) and self-supervision (L$_4$-naïve). These results are reported for the three architectures in \autoref{tab:adapted_depth_completion:4beams}.

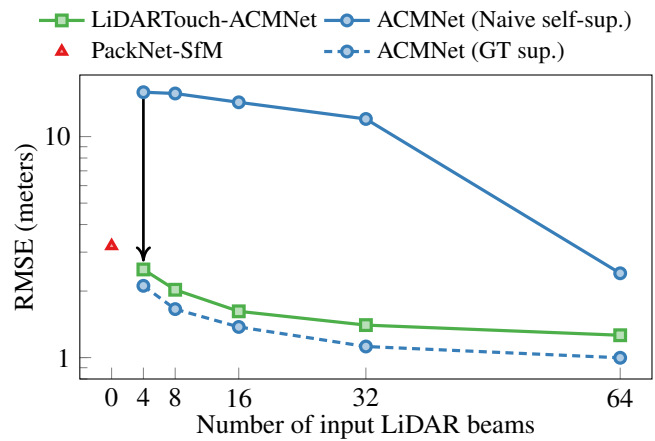
\begin{figure}[ht!]
\begin{tikzpicture} %[trim axis left, trim axis right]
% manual link https://ftp.cc.uoc.gr/mirrors/CTAN/graphics/pgf/contrib/pgfplots/doc/pgfplots.pdf
%\hspace{12pt}
\begin{axis}[
    % cycle list https://tex.stackexchange.com/questions/170221/pgfplots-line-colors
    % link to color palette: http://www.traag.net/wp/wp-content/uploads/2014/06/display_colors1.png
    % load a color `cycle list' from the `colorbrewer' library
    legend cell align={left}, % align legend on the left
    cycle list/Set1-3,
    % define fill color for the marker
    mark list fill={.!40!white},
    % create new `cycle list` from existing `cycle list`
    cycle multiindex* list={
        Set1-3
            \nextlist
        my marks
            \nextlist
        [3 of]linestyles
            \nextlist
        very thick
            \nextlist
    },
    xmin=-4, xmax=68,
    ymin=0.8, ymax=19,
    ylabel= RMSE (meters),
    %ylabel near ticks,
    ylabel style={yshift=-15pt},
    ymode=log,
    log ticks with fixed point,
    ytick pos=left,
    xlabel=Number of input LiDAR beams,
    xtick={0,4,8,16,32,64},
    xlabel style={yshift=5pt},
    xtick pos=bottom,
    legend style={draw=none, at={(0.45,1.25)}, anchor=north,legend columns=2,
    font={\small} 
    }, 
    scale only axis,height=4cm,width=0.85\linewidth,
    ]
        
\pgfplotsset{cycle list shift=+17}
\addplot+[solid] plot coordinates {
    (4, 2.5048)
    (8, 2.0276)
    (16, 1.6195)
    (32, 1.403)
    (64, 1.2622)
};
\addlegendentry{LiDARTouch-ACMNet}

\pgfplotsset{cycle list shift=+3}
\addplot+[solid] plot coordinates {
    (64,    2.4068)
    (32,    12.0147)
    (16,    14.2953)
    (8,     15.6683)
    (4,     15.8803)
};
\addlegendentry{ACMNet (Naive self-sup.)}

\pgfplotsset{cycle list shift=+4}
\addplot+[only marks] plot coordinates {
    (0,     3.2)
};
\addlegendentry{PackNet-SfM}

\pgfplotsset{cycle list shift=+1}    
\addplot+[densely dashed] plot coordinates {
    (64,    0.998)
    (32,    1.1224)
    (16,    1.3761)
    (8,     1.6573)
    (4,     2.1124)
};
\addlegendentry{ACMNet (GT sup.)}

\draw [-{To[length=2mm,width=2mm]}, line width=1pt] (80,2.8) -- (80,1.0);
\end{axis}
\end{tikzpicture}
\caption{
\textbf{Comparison of different supervision schemes for the ACMNet architecture.} In the depth-completion setting, results are highly degraded when ground-truth depth information is no longer available for supervision (blue plots, `GT sup.'\ \textit{vs}.\ `Naive self-sup.').
By combining ideas from self-supervised monocular depth estimation along with a careful integration of the LiDAR signal, we show that our self-supervised LiDARTouch framework can reach performance very close to the one offered by fully-supervised depth completion, as illustrated by the black arrow.
Note that the $y$-axis is log-scaled.
}
\label{fig:frameworks_comparison}
\end{figure}

\parag{Supervised depth completion methods.}
Unsurprisingly, supervising the training of any of the architectures with the privileged ground-truth depth yields better results than our LiDARTouch framework. However, LiDARTouch remain very competitive, \eg 2.504 vs.\ 2.112 in RMSE for ACMNet.
We also investigate the impact of the density of the input LiDAR on these scores in \autoref{fig:frameworks_comparison}. We observe that LiDARTouch is consistently close to the fully-supervised depth completion alternative when the number of layers varies.

\parag{Self-supervised depth completion method.}
The results in \autoref{tab:adapted_depth_completion:4beams} show that the models trained with naïve 4-beam LiDAR self-supervision are unable to converge to decent results. Architectures cannot generalize from such a sparse LiDAR input as the supervisory signal is not sufficient. 
Moreover, in \autoref{fig:frameworks_comparison}, we remark that the naïve self-supervision scheme makes performance plummet when the LiDAR data becomes sparser.
Furthermore, for the sake of completeness, we also experiment with SAN~\citep{packnet-san}, a recent depth completion method with similar fusion scheme to the \imagesparselidar{} we propose in \autoref{sec:learning_system:depth_network}.
Overall the results of SAN in \autoref{tab:sota} and \autoref{tab:adapted_depth_completion:4beams} fall within the expected range, i.e., better than  camera-only methods.

%%%%%%%%%%%%%%%%%%%%%%%%%%%%%%%%%%%%%%%%%%%%%%%%%%%%%%%%%%%%%%%%%%%%%%%% 
%%%%%%%%%%%%%%%%%%%%%%%%%%%%%%%%%%%%%%%%%%%%%%%%%%%%%%%%%%%%%%%%%%%%%%%% 
%%%%%%%%%%%%%%%%%%%%%%%%%%%% INFINITE DEPTH %%%%%%%%%%%%%%%%%%%%%%%%%%%%
%%%%%%%%%%%%%%%%%%%%%%%%%%%%%%%%%%%%%%%%%%%%%%%%%%%%%%%%%%%%%%%%%%%%%%%%
%%%%%%%%%%%%%%%%%%%%%%%%%%%%%%%%%%%%%%%%%%%%%%%%%%%%%%%%%%%%%%%%%%%%%%%%

\section{Alleviating the infinite-depth problem}
\label{sec:infinite_depth}

We now study the infinite-depth problem affecting traditional pipelines and how well does the LiDARTouch framework solve it.
First, we introduce a new metric to assess the degree and the frequency to which a model dramatically overestimates the distance to cars ahead (\autoref{sec:infinite_depth:contrib_metric}).
This metric is employed for a quantitative evaluation of the problem in \autoref{sec:infinite_depth:quantitative}.
Besides, we also provide a qualitative analysis of the problem and the significant improvements offered by LiDARTouch (\autoref{sec:infinite_depth:qualitative}).

%%%%%%%%%%%%%%%%%%%%%%%%
% CONTRIB metric
%%%%%%%%%%%%%%%%%%%%%%%%
\subsection{Catastrophic Distance Rate (CDR) metric}
\label{sec:infinite_depth:contrib_metric}

Monocular image-only depth estimation methods suffer from the infinite-depth problem: vehicles with a motion close to that of the ego vehicle (in other words, with almost no relative motion) can be estimated as being infinitely far away. In the context of autonomous vehicles, such anomalies can lead to potentially dangerous outcomes.
This critical weakness of image-only methods is not well reflected in the commonly-used evaluation metrics, as errors associated with these local flaws are overwhelmed by global scores aggregated at a dataset level.

This problem was qualitatively evaluated in some recent work~\citep{sfm_learner,monodepth2,struct2depth,struct2depth_motion,packnet} but no precise measurement of its severity has yet been proposed. 
To address this issue, we define a novel quantitative metric, called the catastrophic distance rate (CDR), to assess the degree to which a model tends to make such disastrous predictions.

CDR measures the percentage of cars whose estimated distance to the ego-car is catastrophically poor in the test set.
To this end, we use instance segmentation masks for all the vehicles of every image of the test set.
With these vehicle instances, CDR is computed in a two-step process:
\begin{enumerate}
    \item Instance mask filtering to keep the ones potentially concerned by the infinite-depth problem;
    \item Computation of the depth error measured on these instance masks. 
\end{enumerate}

\begin{figure}[]
    \centering
    \includegraphics[clip,width=\linewidth]{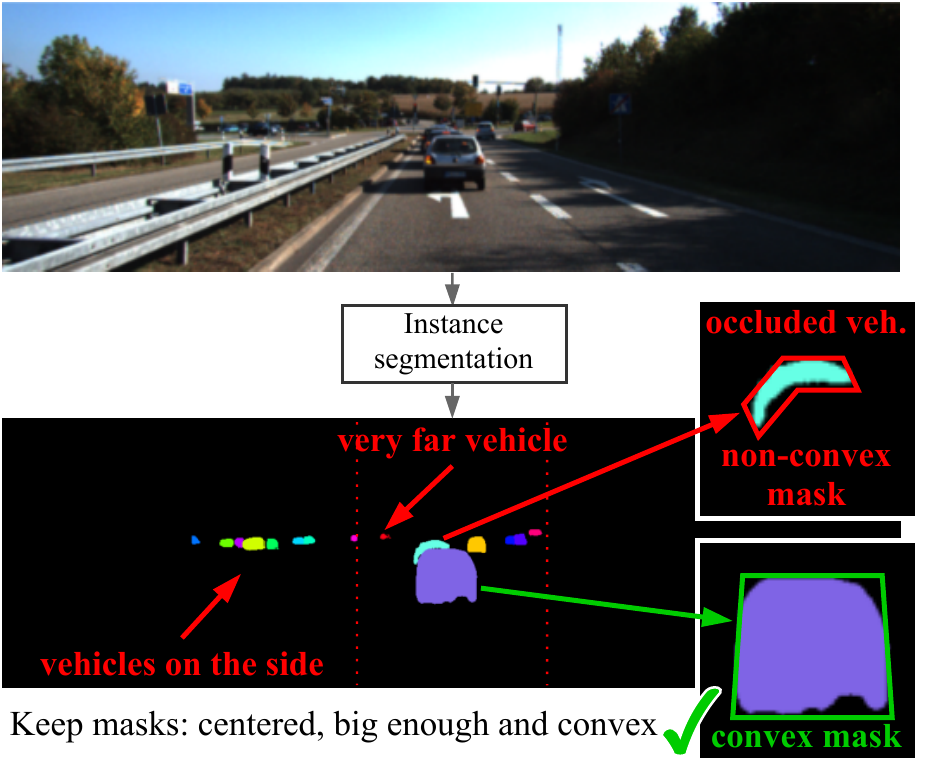}
    \caption{\textbf{Selecting vehicles to compute the CDR metric.} The aim is to extract the individual mask of the first vehicles in front of the ego-car. These are indeed vehicles affected by infinite-depth error due to a small relative motion, leading to potentially catastrophic consequences. The proposed CDR metric computes the rate of such failures over the test set.
    }
    \label{fig:catastrophic_metric}
\end{figure}

\parag{Instance mask filtering.}
For the first step of our CDR metric, we filter out irrelevant masks to only focus on vehicles typically concerned by the infinite-depth problem, \ie{} first vehicle in front, unoccluded and not too far. 
As we use a centered frontal camera, we begin by discarding vehicles that are not in the center of the scene. We also remove cars whose instance masks are too small, considered too far from the ego vehicle.
Then, to assess whether a car is occluded or not, we assume that a heavily occluded vehicle generally has a non-convex shape (\eg incised by the front vehicle) and that, on the contrary, the mask of a non-occluded car is approximately convex. 
The overall process is illustrated in \autoref{fig:catastrophic_metric} and further details are provided in the appendix.

\parag{CDR computation.}
CDR estimates the percentage of instances for which the relative depth error is above a manually-defined ``catastrophic'' threshold $\tau$. 

Within each segmentation mask $M_k$, indexed by $k \in \mathcal{K}$, we define the set $\mathcal{V}_k$ of pixels that possess a ground-truth LiDAR depth measurement: ${\mathcal{V}_k = \left\{ p \mid M_k(p) > 0 \wedge D_k(p) > 0 \right\}}$. Note that, as with $H_t$, $D(p)=0$ if and only if there is no LiDAR point projecting at $p$.
In the KITTI test set, the average size of $\mathcal{V}_k$ is 543.
The error $R_k$ made by the model on the instance mask $M_k$ is measured by the average signed depth error over $\mathcal{V}_k$: 
\begin{equation}
    R_k =  \frac{1}{|\mathcal{V}_k|} \sum_{p \in \mathcal{V}_k} \frac{\hat{D}_k(p) - D_k(p)}{D_k(p)},
\end{equation}
where $| \mathcal{V}_k |$ is the cardinality of $\mathcal{V}_k$. Please note that no absolute value is involved in the design of $R_k$ as we focus only on the infinite-depth problem, \ie $\hat{D}(p) > D(p)$, when a car is predicted catastrophically further away than its true position.

By thresholding the error $R_k$ and aggregating it over instances, we define the ``Catastrophic Distance Rate'' as: 
\begin{equation}
    \text{CDR}(\tau) = \frac{1}{|\mathcal{K}|} \sum_{k \in \mathcal{K}} \llbracket R_k > \tau \rrbracket,
\end{equation}
with $\llbracket \cdot \rrbracket$ the Iverson bracket, $|\mathcal{K}|$ the number of instance masks and $\tau$ a user-defined threshold.
For example, ${\text{CDR}(\tau = 0.5) = 20\%}$ indicates that the distance to front vehicles is over-estimated by more than 50\% of the true distance in 20\% of the cases.

%%%%%%%%%%%%%%%%%%%%%%%%%%%%%%%%%%%%%%%%%%%%
%%%%%%%%%%% QUANTITATIVE
%%%%%%%%%%%%%%%%%%%%%%%%%%%%%%%%%%%%%%%%%%%%
\subsection{Quantitative analysis}
\label{sec:infinite_depth:quantitative}

To verify our intuition that LiDAR self-supervision is a suitable means to mitigate the infinite-depth problem, we study three models:
\begin{itemize}
    \item A model that does not use the LiDAR signal at all, noted `Monodepth w/ IMU supervision', which heavily suffers from the infinite-depth issue;
    \item A model with LiDAR as input and for the P$n$P-estimated pose, but supervised solely with the photometric loss, noted `ACMNet$^\text{P}\PnP$'; 
    \item A model trained within the LiDARTouch framework, using LiDAR for the depth network, pose estimation and self-supervision, noted `\ssACMNet'.
\end{itemize}

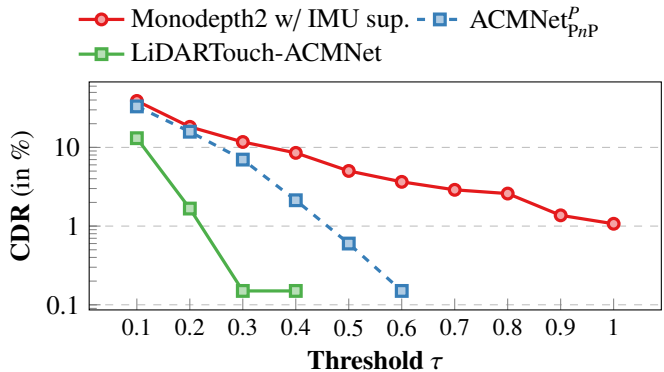
\begin{figure}[hb!]
\centering
\hspace{10pt}
\begin{tikzpicture}[trim axis left, trim axis right]
% manual link https://ftp.cc.uoc.gr/mirrors/CTAN/graphics/pgf/contrib/pgfplots/doc/pgfplots.pdf
    \begin{axis}[
        % cycle list https://tex.stackexchange.com/questions/170221/pgfplots-line-colors
        % link to color palette: http://www.traag.net/wp/wp-content/uploads/2014/06/display_colors1.png
        % load a color `cycle list' from the `colorbrewer' library
        cycle list/Set1-3,
        % define fill color for the marker
        mark list fill={.!40!white},
        % create new `cycle list` from existing `cycle list`
        cycle multiindex* list={
            Set1-3
                \nextlist
            my marks
                \nextlist
            [3 of]linestyles
                \nextlist
            very thick
                \nextlist
        },
        legend cell align={left}, % align legend on the left
        xlabel=\textbf{Threshold $\tau$},
        ylabel=\textbf{CDR} (in \%),
        %ylabel near ticks,
        ylabel style={yshift=-10pt},
        ymode=log,
        log ticks with fixed point,
        ytick pos=left,
        xlabel style={yshift=3pt},
        xtick pos=bottom,
        xtick=data,
        x tick label style = {font = \small, align = center},%, rotate = 70},
        grid=major,
        xmajorgrids=false,
        grid style=dashed,
        legend style={
            at={(0.44,1.38)},
            draw=none, % removes legend border
            anchor=north,legend columns=2,
        },
	    scale only axis,height=3cm,width=0.85\columnwidth,
        ]
        
    \addplot plot coordinates {
        (0.1, 38.96)
        (0.2, 18.26)
        (0.3, 11.72)
        (0.4, 8.52)
        (0.5, 5.02)
        (0.6, 3.65)
        (0.7, 2.89)
        (0.8, 2.59)
        (0.9, 1.37)
        (1, 1.07)
    };
    \addlegendentry{Monodepth2 w/ IMU sup.}
    
    \addplot plot coordinates {
        (0.1, 33.18)
        (0.2, 15.82)
        (0.3, 7.00)
        (0.4, 2.13)
        (0.5, 0.6)
        (0.6, 0.15)
        (0.7, 0)
        (0.8, 0)
        (0.9, 0)
        (1, 0)
    };
    \addlegendentry{ACMNet$^{\textit{P}}\PnP$}
    
    \pgfplotsset{cycle list shift=+3} 
    \addplot+[solid] plot coordinates {
        (0.1, 13.09)
        (0.2, 1.67)
        (0.3, 0.15)
        (0.4, 0.15)
        (0.5, 0.00)
        (0.6, 0.00)
        (0.7, 0.00)
        (0.8, 0.00)
        (0.9, 0.00)
        (1, 0.00)
    };
    \addlegendentry{\ssACMNet}

    \end{axis}
\end{tikzpicture}
\caption{\textbf{Plot of the CDR metric for various thresholds $\tau$.} $y$-axis is log-scaled.}
\label{fig:inf_depth_study}
\end{figure}

\newcommand{\infdepthimage}[1]{\includegraphics[width=0.255\linewidth]{#1}}
\newcommand{\zoominfdepthimage}[1]{\includegraphics[width=0.08\linewidth]{#1}}

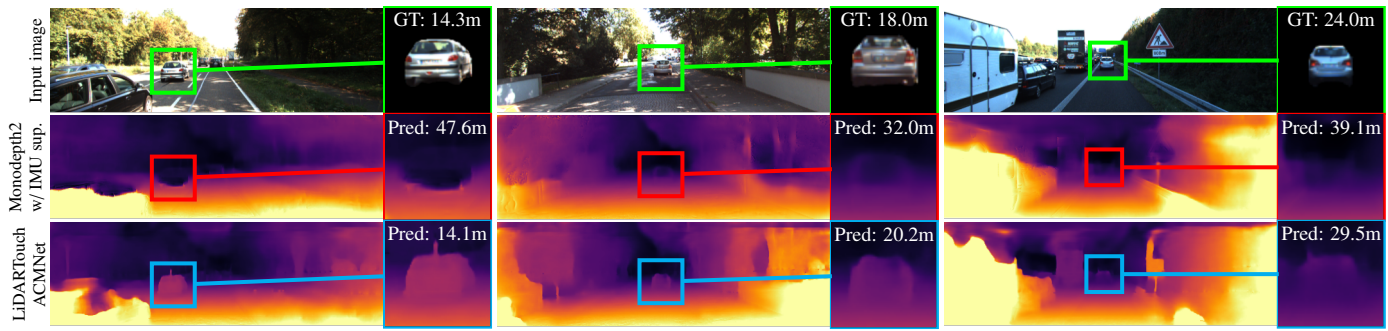
\begin{figure*}[h!]
%%%%%%%%%%%%%%%%%%%%%%%%%%%%%%%
%%%%% VAL 308
%%%%%%%%%%%%%%%%%%%%%%%%%%%%%%%
\resizebox{\textwidth}{!}{
\begin{tikzpicture}[
    every node/.style={inner sep=0,outer sep=0},
    label/.style = {
        inner sep=2pt,
        font=\scriptsize,
        align=center,
    },
    scores/.style = {
        inner sep=2pt, 
        outer sep=0pt,
        font=\footnotesize,
        text=white,
        align=center,
        anchor=north,
    },
    spy using outlines={magnification=7, every spy on node/.append style={thick}}
]
    %%%%%%%%%%%%%%%%%%%%% RGB %%%%%%%%%%%%%%%%
    \node (FigA) {\infdepthimage{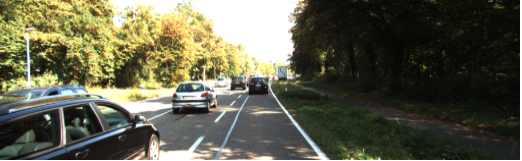}};
    %%% LABEL
    \node [anchor=east, label] at ($(FigA.west)$) {\rotatebox{90}{Input image}};
    %%% ZOOM
    \node [anchor=west, draw=green, ultra thick] (zoomFigA) at (FigA.east) {\zoominfdepthimage{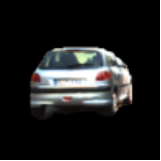}};
    %%% RECTANGLE
    \node[rectangle, draw=green, ultra thick, size=0.6cm] (rectFigA) at ($(FigA)+(-0.62,-0.15)$) {};
    %%% LINE
    \draw[green, ultra thick] (rectFigA) -- (zoomFigA);
    %%% SCORE
    \node [scores] at (zoomFigA.north) {GT: 14.3m};
    
    %%%%%%%%%%%%%%%%%%%%% DEPTH IMAGE ONLY  %%%%%%%%%%%%%%%%%%%%% 
    \node[below=0.05cm of FigA] (FigB) {\infdepthimage{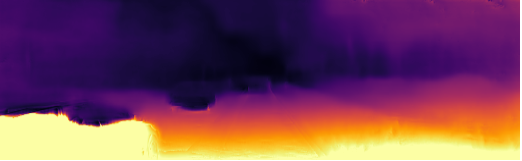}};
    %%% LABEL 
    \node [anchor=east, label] at ($(FigB.west)$){\rotatebox{90}{\parbox{2cm}{\centering Monodepth2\\w/ IMU sup.}}};
    %%% ZOOM
    \node [anchor=west, draw=red, ultra thick] (zoomFigB) at (FigB.east) {\zoominfdepthimage{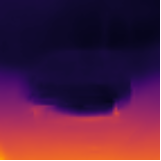}};
    %%% RECTANGLE
    \node[rectangle, draw=red, ultra thick, size=0.6cm] (rectFigB) at ($(FigB)+(-0.62,-0.15)$) {};
    %%% LINE
    \draw[red, ultra thick] (rectFigB) -- (zoomFigB);
    %%% SCORE
    \node [scores] at (zoomFigB.north) {Pred: 47.6m};
    
    %%%%%%%%%%%%%%%%%%%%% DEPTH LIDAR  %%%%%%%%%%%%%%%%%%%%%
    \node[below=0.05cm of FigB] (FigC) {\infdepthimage{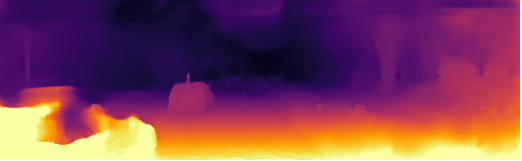}};
    %%% LABEL
    \node [anchor=east, label] at ($(FigC.west)$) {\rotatebox{90}{\parbox{2cm}{\centering LiDARTouch\\ACMNet}}};
    %\node [anchor=east, label] at ($(FigC.west)$) {\rotatebox{90}{Ours}};
    %%% ZOOM
    \node[anchor=west, draw=cyan, ultra thick] (zoomFigC) at (FigC.east) {\zoominfdepthimage{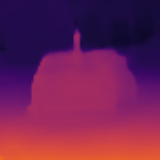}};
    %%% RECTANGLE
    \node[rectangle, draw=cyan, ultra thick, size=0.6cm] (rectFigC) at ($(FigC)+(-0.62,-0.15)$) {};
    %%% LINE
    \draw[cyan, ultra thick] (rectFigC) -- (zoomFigC);
    %%% SCORE
    \node [scores] at (zoomFigC.north) {Pred: 14.1m};
    
%%%%%%%%%%%%%%%%%%%%%%%%%%%%%%%
%%%%% VAL 395
%%%%%%%%%%%%%%%%%%%%%%%%%%%%%%%
    %%%%%%%%%%%%%%%%%%%%% RGB %%%%%%%%%%%%%%%%
    \node[right=0.1cm of zoomFigA] (FigA2) {\infdepthimage{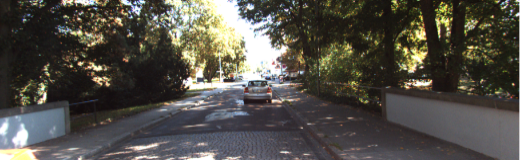}};
    %%% LABEL
    %\node [anchor=east] at ($(FigA.west)$) {\rotatebox{90}{Input image}};
    %%% ZOOM
    \node [anchor=west, draw=green, ultra thick] (zoomFigA2) at (FigA2.east) {\zoominfdepthimage{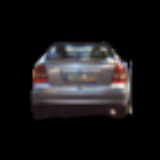}};
    %%% RECTANGLE
    \node[rectangle, draw=green, ultra thick, size=0.6cm] (rectFigA2) at ($(FigA2)+(-0.05,-0.1)$) {};
    %%% LINE
    \draw[green, ultra thick] (rectFigA2) -- (zoomFigA2);
    %%% SCORE
    \node [scores] at (zoomFigA2.north) {GT: 18.0m};
    
    %%%%%%%%%%%%%%%%%%%%% DEPTH IMAGE ONLY  %%%%%%%%%%%%%%%%%%%%% 
    \node[below=0.05cm of FigA2] (FigB2) {\infdepthimage{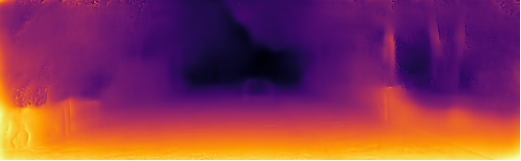}};
    %%% LABEL
    %\node [anchor=east] at ($(FigB.west)$) {\rotatebox{90}{\parbox{2cm}{\centering Image-only}}};
    %%% ZOOM
    \node [anchor=west, draw=red, ultra thick] (zoomFigB2) at (FigB2.east) {\zoominfdepthimage{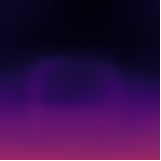}};
    %%% RECTANGLE
    \node[rectangle, draw=red, ultra thick, size=0.6cm] (rectFigB2) at ($(FigB2)+(-0.05,-0.1)$) {};
    %%% LINE
    \draw[red, ultra thick] (rectFigB2) -- (zoomFigB2);
    %%% SCORE
    \node [scores] at (zoomFigB2.north) {Pred: 32.0m};
    
    %%%%%%%%%%%%%%%%%%%%% DEPTH LIDAR  %%%%%%%%%%%%%%%%%%%%% 
    \node[below=0.05cm of FigB2] (FigC2) {\infdepthimage{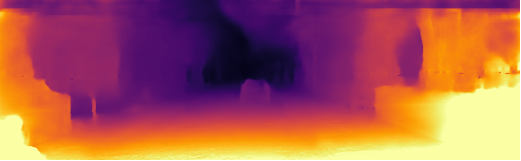}};
    %%% LABEL
    %\node [anchor=east] at ($(FigC.west)$) {\rotatebox{90}{\parbox{2cm}{\centering LiDARTouch}}};
    %%% ZOOM
    \node[anchor=west, draw=cyan, ultra thick] (zoomFigC2) at (FigC2.east) {\zoominfdepthimage{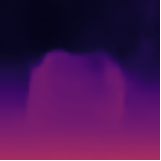}};
    %%% RECTANGLE
    \node[rectangle, draw=cyan, ultra thick, size=0.6cm] (rectFigC2) at ($(FigC2)+(-0.05,-0.1)$) {};
    %%% LINE
    \draw[cyan, ultra thick] (rectFigC2) -- (zoomFigC2);
    %%% SCORE
    \node [scores] at (zoomFigC2.north) {Pred: 20.2m};

%%%%%%%%%%%%%%%%%%%%%%%%%%%%%%%
%%%%% TEST 635
%%%%%%%%%%%%%%%%%%%%%%%%%%%%%%%
    %%%%%%%%%%%%%%%%%%%%% RGB %%%%%%%%%%%%%%%%
    \node[right=0.1cm of zoomFigA2] (FigA3) {\infdepthimage{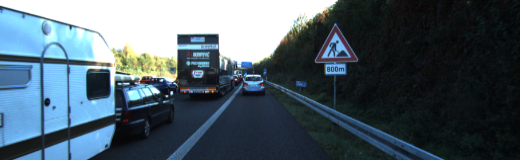}};
    %%% LABEL
    %\node [anchor=east] at ($(FigA.west)$) {\rotatebox{90}{Input image}};
    %%% ZOOM
    \node [anchor=west, draw=green, ultra thick] (zoomFigA3) at (FigA3.east) {\zoominfdepthimage{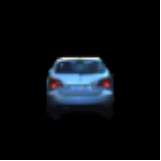}};
    %%% RECTANGLE
    \node[rectangle, draw=green, ultra thick, size=0.5cm] (rectFigA3) at ($(FigA3)+(-0.05,0)$) {};
    %%% LINE
    \draw[green, ultra thick] (rectFigA3) -- (zoomFigA3);
    %%% SCORE
    \node [scores] at (zoomFigA3.north) {GT: 24.0m};
    
    %%%%%%%%%%%%%%%%%%%%% DEPTH IMAGE ONLY  %%%%%%%%%%%%%%%%%%%%% 
    \node[below=0.05cm of FigA3] (FigB3) {\infdepthimage{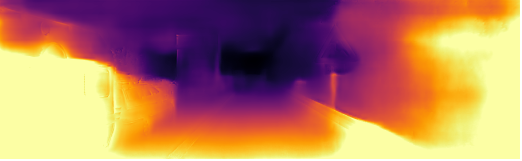}};
    %%% LABEL
    %\node [anchor=east] at ($(FigB.west)$) {\rotatebox{90}{\parbox{2cm}{\centering Image-only}}};
    %%% ZOOM
    \node [anchor=west, draw=red, ultra thick] (zoomFigB3) at (FigB3.east) {\zoominfdepthimage{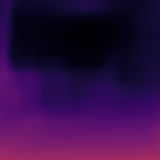}};
    %%% RECTANGLE
    \node[rectangle, draw=red, ultra thick, size=0.5cm] (rectFigB3) at ($(FigB3)+(-0.1,0)$) {};
    %%% LINE
    \draw[red, ultra thick] (rectFigB3) -- (zoomFigB3);
    %%% SCORE
    \node [scores] at (zoomFigB3.north) {Pred: 39.1m};
    
    %%%%%%%%%%%%%%%%%%%%% DEPTH LIDAR  %%%%%%%%%%%%%%%%%%%%% 
    \node[below=0.05cm of FigB3] (FigC3) {\infdepthimage{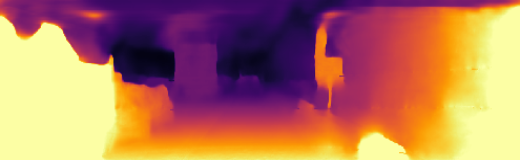}};
    %%% LABEL
    % \node [anchor=east] at ($(FigC.west)$) {\rotatebox{90}{\parbox{2cm}{\centering LiDARTouch}}};
    %%% ZOOM
    \node[anchor=west, draw=cyan, ultra thick] (zoomFigC3) at (FigC3.east) {\zoominfdepthimage{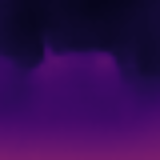}};
    %%% RECTANGLE
    \node[rectangle, draw=cyan, ultra thick, size=0.5cm] (rectFigC3) at ($(FigC3)+(-0.1,0)$) {};
    %%% LINE
    \draw[cyan, ultra thick] (rectFigC3) -- (zoomFigC3);
    %%% SCORE
    \node [scores] at (zoomFigC3.north) {Pred: 29.5m};
\end{tikzpicture}
}

\caption{\textbf{Mitigation of the infinite-depth problem.}
Self-supervised image-only approaches tend to predict objects with no relative-motion at an infinite depth, as indicated by the hole in the depth close-up (red).
In contrast, our LiDARTouch framework estimates the depth of these vehicles, as shown in the green close-up. Note that for the example in the middle, we verified that no LiDAR measurement falls on the car. This shows that our training framework can generalize well to cases where no LIDAR is available on critical moving vehicles.
\label{fig:quali_infinite_depth}
}
\end{figure*}

We plot the distribution of the CDR metric against the chosen threshold $\tau$ in \autoref{fig:inf_depth_study}. We observe that the more LiDAR information is integrated, the fewer catastrophic estimations occur. 

Indeed, ACMNet$^\text{P}\PnP$, which uses LiDAR both in input and pose, improves over Monodepth2 but is still affected by the infinite-depth issue. We also see a clear improvement of our \ssACMNet{} over the two other models.
For example, for $\tau=0.5$, \ie the distance of a car is overestimated by at least half, Monodepth2 has a metric score of 5.02\% while ACMNet$^\text{P}\PnP$ has 0.6\% and \ssACMNet{} 0.0\%. Such results show that Monodepth2 predictions cannot be trusted for downstream tasks such as car detection or free space estimation that are both required by functions like automatic emergency braking, keep-lane assist or adaptive cruise control. While ACMNet$^\text{P}\PnP$ reduces the likelihood of catastrophic estimation by 8 folds for $\tau=0.5$, 0.6\% is still too high to implement in a critical system intended for wide commercial use.

Overall, a network trained with our pipeline is significantly less impacted by the infinite-depth problem and we validate our hypothesis that, during training, the LiDAR self-supervision disambiguates cars estimated too far from their real distance. Hence, our models can accurately and safely handle moving objects with no relative motion, typical of cars in fluid traffic.

%%%%%%%%%%%%%%%%%%%%%%%%%%%%%%%%%%%%%%%%%%%%
%%%%%%%%%%% QUALITATIVE
%%%%%%%%%%%%%%%%%%%%%%%%%%%%%%%%%%%%%%%%%%%%

\subsection{Qualitative analysis}
\label{sec:infinite_depth:qualitative}
\begin{figure*}[]
\centering
\begin{tikzpicture}[
    image/.style = {
        text width=0.315\textwidth,
        inner sep=0pt, 
        outer sep=0pt
        },
    label/.style = {
        inner sep=2pt,
        font=\footnotesize,
        align=center,
    },
    node distance = 1mm and 1mm
]

%%%%%%%%%%%%%%%%%%%%%%%%% INPUTS %%%%%%%%%%%%%%%%%%%%%%%%
    %%%%%%%%%% col1
    \node [image] (col1in) {\includegraphics[width=\linewidth]{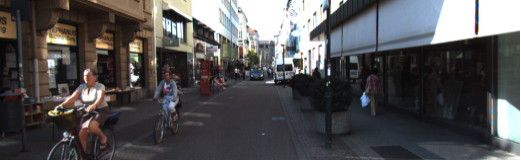}};
    
    %%%%%%%%%% col2
    \node [image, right=of col1in] (col2in) {\includegraphics[width=\linewidth]{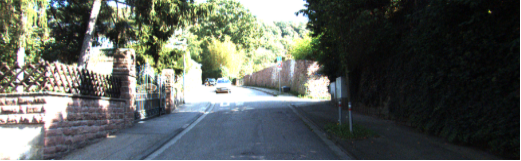}};
    
    %%%%%%%%%% col3
    \node [image, right=of col2in] (col3in) {\includegraphics[width=\linewidth]{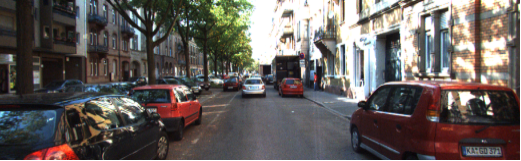}};
    
    % LEGEND
    \node [label, anchor=east] at ($(col1in.west)$) {\rotatebox{90}{Image Input}};
    
%%%%%%%%%%%%%%%%%%%%%%  MD2 imu  %%%%%%%%%%%%%%%%%%%%%%%%
    %%%%%%%%%% col1
    \node [image, below=of col1in] (col1MD2) {\includegraphics[width=\linewidth]{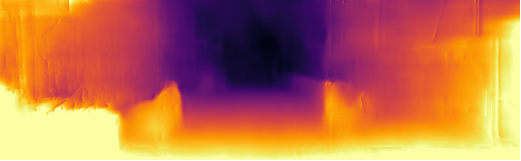}};
    
    %%%%%%%%%% col2
    \node [image, right=of col1MD2] (col2MD2) {\includegraphics[width=\linewidth]{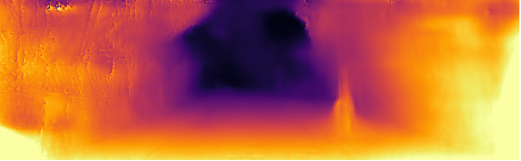}};
    
    %%%%%%%%%% col3
    \node [image, right=of col2MD2] (col3MD2) {\includegraphics[width=\linewidth]{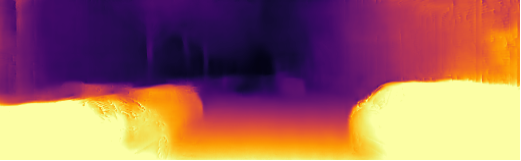}};
    
    % LEGEND
    \node [label, anchor=east] at ($(col1MD2.west)$) {\rotatebox{90}{\parbox{2cm}{\centering Monodepth2\\w/ IMU sup.}}};
    
%%%%%%%%%%%%%%%%%%%%%%%% LiDARTouch  %%%%%%%%%%%%%%%%%%%%%%%%
    
    %%%%%%%%%% col1
    \node [image, below=of col1MD2] (col1LT) {\includegraphics[width=\linewidth]{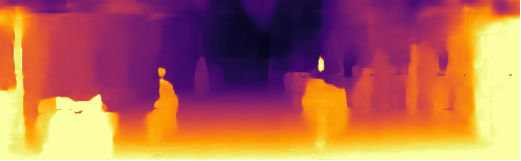}};
    
    %%%%%%%%%% col2
    \node [image, right=of col1LT] (col2LT) {\includegraphics[width=\linewidth]{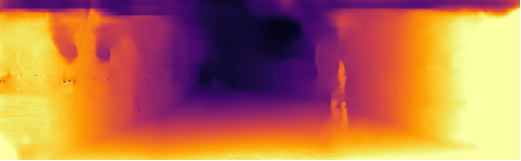}};
    
    %%%%%%%%%% col3
    \node [image, right=of col2LT] (col3LT) {\includegraphics[width=\linewidth]{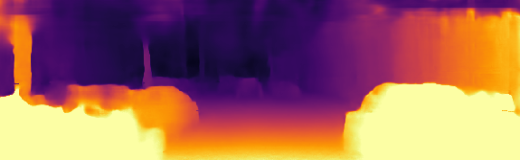}};
    
    % LEGEND
    \node [label, anchor=east] at ($(col1LT.west)$) {\rotatebox{90}{\parbox{2cm}{\centering LiDARTouch\\ACMNet}}};
    
%%%%%%%%%%%%%%%%%%%%%%%%% FULLY SUP  %%%%%%%%%%%%%%%%%%%%%%%%
    %%%%%%%%%% col1
    \node [image, below=of col1LT] (col1GT) {\includegraphics[width=\linewidth]{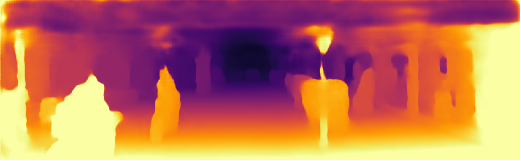}};
    
    %%%%%%%%%% col2
    \node [image, right=of col1GT] (col2GT) {\includegraphics[width=\linewidth]{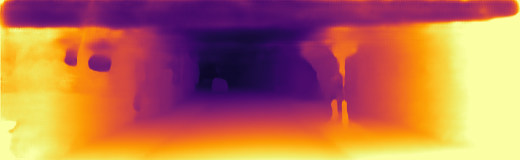}};
    
    %%%%%%%%%% col3
    \node [image, right=of col2GT] (col3GT) {\includegraphics[width=\linewidth]{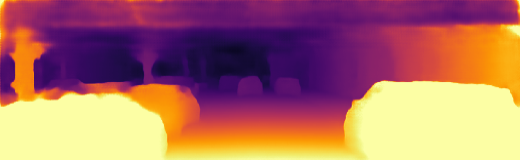}};
    
    % LEGEND
    \node [label, anchor=east] at ($(col1GT.west)$) {\rotatebox{90}{\parbox{2cm}{\centering GT sup.\\ACMNet}}};

% column
\node [label, anchor=north] at ($(col1GT.south)$) {(a)};

\node [label, anchor=north] at ($(col2GT.south)$) {(b)};

\node [label, anchor=north] at ($(col3GT.south)$) {(c)};
     
\end{tikzpicture}
\caption{\textbf{Qualitative comparison of LiDARTouch with other existing frameworks}. Monodepth2 is trained with IMU supervision. The model trained with GT supervision gives sharper depth estimates, but struggles in regions where GT signal is not available (e.g., top of the scene).}
\label{fig:qualitative_frameworks}
\end{figure*}
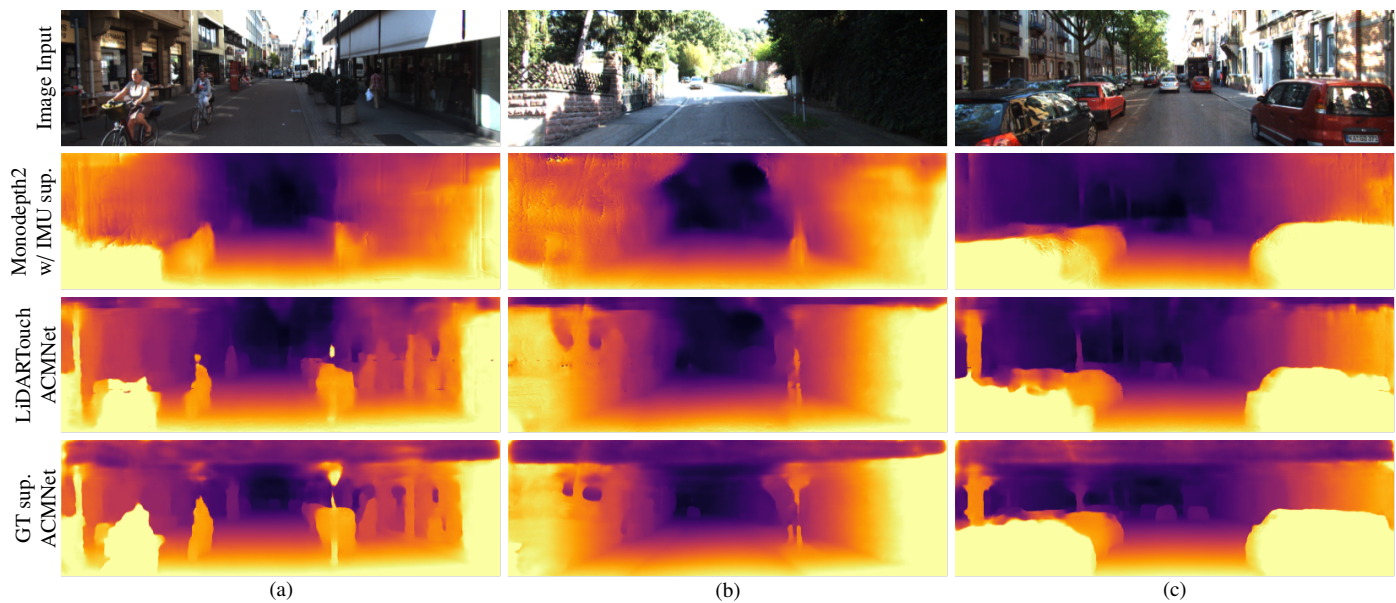

The three examples in \autoref{fig:quali_infinite_depth} illustrate
the improvement of our framework over the classic self-supervised camera-only pipeline.
On the leftmost column, we observe a typical `hole' in the depth map where \imagealone{} with IMU supervision estimates a vehicle three times more distant than in reality. in contrast to our model without such holes.

In addition to \autoref{fig:quali_infinite_depth}, we provide some qualitative analyses where we show the depth maps obtained for different frameworks in  \autoref{fig:qualitative_frameworks}. 
First, we observe better overall depth maps with \ssACMNet{} than with \imagealone{}.
For example, we better estimate the two moving cyclists in \autoref{fig:qualitative_frameworks}a  as well as the fine tree trunks in \autoref{fig:qualitative_frameworks}c.

As expected, the fully-supervised method ACMNet (GT-sup.) delivers the best-qualitative depth maps, as it leverages privileged ground-truth LiDAR depth during training.
However, we observe that self-supervised approaches (Monodepth2 and \ssACMNet{}) better estimate areas near the top of the scene.
This can be explained as LiDAR points are absent from regions above the road, which hinders ACMNet (GT-sup.) prediction in these regions due to the lack of supervisory signal it uses (last row in \autoref{fig:qualitative_frameworks}).

Despite the successful integration of LiDAR in LiDARTouch, we note that some local depth estimation artifacts still occur, similar to the maps obtained from self-supervised depth estimation methods.
Typically, this concerns distorted, reflective and color-saturated regions because the photometric reconstruction loss assumes Lambertian surfaces (cars in \autoref{fig:qualitative_frameworks}c).
Our model may also produce blurry depth predictions for small or thin objects, such as traffic signs (Figures \ref{fig:qualitative_frameworks}a and \ref{fig:qualitative_frameworks}b).

%%%%%%%%%%%%%%%%%%%%%%%%%%%%%%%%%%
%%%%%%%%%%% Conclusion %%%%%%%%%%% 
%%%%%%%%%%%%%%%%%%%%%%%%%%%%%%%%%%
\section{Conclusion}

In this paper, we introduce LiDARTouch, a novel self-supervised framework for depth estimation with few-beam LiDAR.
While being extremely sparse, we show that the LiDAR signal can be leveraged at three complementary levels of a self-supervised learning scheme. 
Across four different architectures, the LiDARTouch framework can reach competitive performances with respect to fully-supervised depth completion methods while being significantly cheaper and more annotation friendly.
Moreover, we show that the sparse LiDAR signal provides valuable cues to disambiguate monocular depth estimation at a global level as well as for moving objects.
Our method can be trained on any domain with no modification, and it can thus bring accurate and metric depth estimation at a fleet scale.

With our novel LiDARTouch framework, the new CDR metric to measure the infinite-depth problem, and the associate source code of our code, we hope to enable further research on the task of monocular depth prediction with minimal LiDAR input, typical of real-world assisted/automated driving systems.

\paragraph{\bf Acknowledgements} Karteek Alahari was supported in part by the ANR grant AVENUE (ANR-18-CE23-0011). This work was granted access to the HPC resources of IDRIS under the allocation 2021-101766 made by GENCI.

\section*{Authors' biography}
Florent Bartoccioni is a CIFRE PhD candidate in valeo.ai and Inria Grenoble Rhône-Alpes. He received the M.Sc. degree in computer science  from the École Normale Supérieure de Rennes, France, in 2019.
His research interests include dynamic scene forecasting and 3D perception for autonomous robots. 

\'Eloi Zablocki is a research scientist at valeo.ai. He obtained his Ph.D. at Sorbonne University in 2019, on multimodal machine learning with language and vision. His research interests include computer vision for scene understanding, motion forecasting, and explainability of machine learning models.

Patrick Pérez is Scientific Director of valeo.ai. Before joining Valeo, Patrick Pérez has been a researcher at Technicolor (2009-2018), Inria (1993-2000, 2004-2009) and Microsoft Research Cambridge (2000-2004). His research interests include multimodal scene understanding and computational imaging.

Matthieu Cord is a full professor at Sorbonne University. He is also a part-time principal scientist at valeo.ai. His research expertise includes computer vision, machine learning, and artificial intelligence. He is the author of more than 150 publications on image classification, segmentation, deep learning, and multimodal vision and language understanding. He is an honorary member of the Institut Universitaire de France and served from 2015 to 2018 as an AI expert at CNRS and at the French National Research Agency.

Karteek Alahari is a senior researcher at Inria in the Grenoble - Rhône-Alpes center. He was previously a postdoctoral fellow in the Inria WILLOW team at the Department of Computer Science in École Normale Supérieure, after completing his PhD in 2010 in the UK. His current research focuses on learning robust and effective visual representations, when only partially-supervised data is available.

\bibliographystyle{model2-names}
\bibliography{refs}

\begin{thebibliography}{55}
\expandafter\ifx\csname natexlab\endcsname\relax\def\natexlab#1{#1}\fi
\providecommand{\url}[1]{\texttt{#1}}
\providecommand{\href}[2]{#2}
\providecommand{\path}[1]{#1}
\providecommand{\DOIprefix}{doi:}
\providecommand{\ArXivprefix}{arXiv:}
\providecommand{\URLprefix}{URL: }
\providecommand{\Pubmedprefix}{pmid:}
\providecommand{\doi}[1]{\href{http://dx.doi.org/#1}{\path{#1}}}
\providecommand{\Pubmed}[1]{\href{pmid:#1}{\path{#1}}}
\providecommand{\bibinfo}[2]{#2}
\ifx\xfnm\relax \def\xfnm[#1]{\unskip,\space#1}\fi
%Type = Inproceedings
\bibitem[{Amiri et~al.(2019)Amiri, Loo and Zhang}]{semidepth}
\bibinfo{author}{Amiri, A.J.}, \bibinfo{author}{Loo, S.Y.},
  \bibinfo{author}{Zhang, H.}, \bibinfo{year}{2019}.
\newblock \bibinfo{title}{Semi-supervised monocular depth estimation with
  left-right consistency using deep neural network}, in:
  \bibinfo{booktitle}{IEEE ROBIO}.
%Type = Article
\bibitem[{Bradski(2000)}]{opencv_library}
\bibinfo{author}{Bradski, G.}, \bibinfo{year}{2000}.
\newblock \bibinfo{title}{The opencv library}.
\newblock \bibinfo{journal}{Dr. Dobb's Journal of Software Tools} .
%Type = Article
\bibitem[{Caesar et~al.(2019)Caesar, Bankiti, Lang, Vora, Liong, Xu, Krishnan,
  Pan, Baldan and Beijbom}]{nuscenes2019}
\bibinfo{author}{Caesar, H.}, \bibinfo{author}{Bankiti, V.},
  \bibinfo{author}{Lang, A.H.}, \bibinfo{author}{Vora, S.},
  \bibinfo{author}{Liong, V.E.}, \bibinfo{author}{Xu, Q.},
  \bibinfo{author}{Krishnan, A.}, \bibinfo{author}{Pan, Y.},
  \bibinfo{author}{Baldan, G.}, \bibinfo{author}{Beijbom, O.},
  \bibinfo{year}{2019}.
\newblock \bibinfo{title}{nuscenes: A multimodal dataset for autonomous
  driving}.
\newblock \bibinfo{journal}{arXiv preprint arXiv:1903.11027} .
%Type = Inproceedings
\bibitem[{Casser et~al.(2019a)Casser, Pirk, Mahjourian and
  Angelova}]{struct2depth}
\bibinfo{author}{Casser, V.}, \bibinfo{author}{Pirk, S.},
  \bibinfo{author}{Mahjourian, R.}, \bibinfo{author}{Angelova, A.},
  \bibinfo{year}{2019}a.
\newblock \bibinfo{title}{Depth prediction without the sensors: Leveraging
  structure for unsupervised learning from monocular videos}, in:
  \bibinfo{booktitle}{AAAI}.
%Type = Inproceedings
\bibitem[{Casser et~al.(2019b)Casser, Pirk, Mahjourian and
  Angelova}]{struct2depth_motion}
\bibinfo{author}{Casser, V.}, \bibinfo{author}{Pirk, S.},
  \bibinfo{author}{Mahjourian, R.}, \bibinfo{author}{Angelova, A.},
  \bibinfo{year}{2019}b.
\newblock \bibinfo{title}{Unsupervised monocular depth and ego-motion learning
  with structure and semantics}, in: \bibinfo{booktitle}{CVPR Workshop}.
%Type = Inproceedings
\bibitem[{Chang and Chen(2018)}]{psmnet}
\bibinfo{author}{Chang, J.R.}, \bibinfo{author}{Chen, Y.S.},
  \bibinfo{year}{2018}.
\newblock \bibinfo{title}{Pyramid stereo matching network}, in:
  \bibinfo{booktitle}{CVPR}.
%Type = Inproceedings
\bibitem[{Chang et~al.(2019)Chang, Lambert, Sangkloy, Singh, Bak, Hartnett,
  Wang, Carr, Lucey, Ramanan and Hays}]{Argoverse}
\bibinfo{author}{Chang, M.}, \bibinfo{author}{Lambert, J.},
  \bibinfo{author}{Sangkloy, P.}, \bibinfo{author}{Singh, J.},
  \bibinfo{author}{Bak, S.}, \bibinfo{author}{Hartnett, A.},
  \bibinfo{author}{Wang, D.}, \bibinfo{author}{Carr, P.},
  \bibinfo{author}{Lucey, S.}, \bibinfo{author}{Ramanan, D.},
  \bibinfo{author}{Hays, J.}, \bibinfo{year}{2019}.
\newblock \bibinfo{title}{Argoverse: {3D} tracking and forecasting with rich
  maps}, in: \bibinfo{booktitle}{CVPR}.
%Type = Inproceedings
\bibitem[{Cheng et~al.(2019)Cheng, Zhong, Dai, Ji and Li}]{lidarstereonet}
\bibinfo{author}{Cheng, X.}, \bibinfo{author}{Zhong, Y.}, \bibinfo{author}{Dai,
  Y.}, \bibinfo{author}{Ji, P.}, \bibinfo{author}{Li, H.},
  \bibinfo{year}{2019}.
\newblock \bibinfo{title}{Noise-aware unsupervised deep lidar-stereo fusion},
  in: \bibinfo{booktitle}{CVPR}.
%Type = Inproceedings
\bibitem[{Cordts et~al.(2016)Cordts, Omran, Ramos, Rehfeld, Enzweiler,
  Benenson, Franke, Roth and Schiele}]{cityscape}
\bibinfo{author}{Cordts, M.}, \bibinfo{author}{Omran, M.},
  \bibinfo{author}{Ramos, S.}, \bibinfo{author}{Rehfeld, T.},
  \bibinfo{author}{Enzweiler, M.}, \bibinfo{author}{Benenson, R.},
  \bibinfo{author}{Franke, U.}, \bibinfo{author}{Roth, S.},
  \bibinfo{author}{Schiele, B.}, \bibinfo{year}{2016}.
\newblock \bibinfo{title}{The {C}ityscapes dataset for semantic urban scene
  understanding}, in: \bibinfo{booktitle}{CVPR}.
%Type = Inproceedings
\bibitem[{Deng et~al.(2009)Deng, Dong, Socher, Li, Li and Li}]{ImageNet}
\bibinfo{author}{Deng, J.}, \bibinfo{author}{Dong, W.},
  \bibinfo{author}{Socher, R.}, \bibinfo{author}{Li, L.}, \bibinfo{author}{Li,
  K.}, \bibinfo{author}{Li, F.}, \bibinfo{year}{2009}.
\newblock \bibinfo{title}{{ImageNet}: {A} large-scale hierarchical image
  database}, in: \bibinfo{booktitle}{CVPR}.
%Type = Article
\bibitem[{Deng et~al.(2017)Deng, Todorovic and {Jan
  Latecki}}]{RGBD_unsup_detect}
\bibinfo{author}{Deng, Z.}, \bibinfo{author}{Todorovic, S.},
  \bibinfo{author}{{Jan Latecki}, L.}, \bibinfo{year}{2017}.
\newblock \bibinfo{title}{Unsupervised object region proposals for rgb-d indoor
  scenes}.
\newblock \bibinfo{journal}{CVIU} \bibinfo{volume}{154},
  \bibinfo{pages}{127--136}.
%Type = Article
\bibitem[{Douglas and Peucker(1973)}]{douglas_peucker}
\bibinfo{author}{Douglas, D.H.}, \bibinfo{author}{Peucker, T.K.},
  \bibinfo{year}{1973}.
\newblock \bibinfo{title}{Algorithms for the reduction of the number of points
  required to represent a digitized line or its caricature}.
\newblock \bibinfo{journal}{Cartographica: Intl.\ J.\ Geographic Information
  and Geovisualization} \bibinfo{volume}{10}, \bibinfo{pages}{112--122}.
%Type = Inproceedings
\bibitem[{Eigen et~al.(2014)Eigen, Puhrsch and Fergus}]{eigen}
\bibinfo{author}{Eigen, D.}, \bibinfo{author}{Puhrsch, C.},
  \bibinfo{author}{Fergus, R.}, \bibinfo{year}{2014}.
\newblock \bibinfo{title}{Depth map prediction from a single image using a
  multi-scale deep network}, in: \bibinfo{booktitle}{NeurIPS}.
%Type = Inproceedings
\bibitem[{Fu et~al.(2018)Fu, Gong, Wang, Batmanghelich and Tao}]{DORN}
\bibinfo{author}{Fu, H.}, \bibinfo{author}{Gong, M.}, \bibinfo{author}{Wang,
  C.}, \bibinfo{author}{Batmanghelich, K.}, \bibinfo{author}{Tao, D.},
  \bibinfo{year}{2018}.
\newblock \bibinfo{title}{Deep ordinal regression network for monocular depth
  estimation}, in: \bibinfo{booktitle}{CVPR}.
%Type = Article
\bibitem[{Gao et~al.(2003)Gao, Hou, Tang and Cheng}]{P3P}
\bibinfo{author}{Gao, X.}, \bibinfo{author}{Hou, X.}, \bibinfo{author}{Tang,
  J.}, \bibinfo{author}{Cheng, H.}, \bibinfo{year}{2003}.
\newblock \bibinfo{title}{Complete solution classification for the
  perspective-three-point problem}.
\newblock \bibinfo{journal}{IEEE TPAMI} \bibinfo{volume}{25},
  \bibinfo{pages}{930--943}.
%Type = Inproceedings
\bibitem[{Geiger et~al.(2012)Geiger, Lenz and Urtasun}]{kitti}
\bibinfo{author}{Geiger, A.}, \bibinfo{author}{Lenz, P.},
  \bibinfo{author}{Urtasun, R.}, \bibinfo{year}{2012}.
\newblock \bibinfo{title}{Are we ready for autonomous driving? {The} {KITTI}
  vision benchmark suite}, in: \bibinfo{booktitle}{CVPR}.
%Type = Inproceedings
\bibitem[{Godard et~al.(2019)Godard, Aodha, Firman and Brostow}]{monodepth2}
\bibinfo{author}{Godard, C.}, \bibinfo{author}{Aodha, O.M.},
  \bibinfo{author}{Firman, M.}, \bibinfo{author}{Brostow, G.J.},
  \bibinfo{year}{2019}.
\newblock \bibinfo{title}{Digging into self-supervised monocular depth
  estimation}, in: \bibinfo{booktitle}{ICCV}.
%Type = Inproceedings
\bibitem[{Godard et~al.(2017)Godard, {Mac Aodha} and Brostow}]{monodepth17}
\bibinfo{author}{Godard, C.}, \bibinfo{author}{{Mac Aodha}, O.},
  \bibinfo{author}{Brostow, G.J.}, \bibinfo{year}{2017}.
\newblock \bibinfo{title}{Unsupervised monocular depth estimation with
  left-right consistency}, in: \bibinfo{booktitle}{CVPR}.
%Type = Article
\bibitem[{Groenendijk et~al.(2020)Groenendijk, Karaoglu, Gevers and
  Mensink}]{depth_adversialreconstruct}
\bibinfo{author}{Groenendijk, R.}, \bibinfo{author}{Karaoglu, S.},
  \bibinfo{author}{Gevers, T.}, \bibinfo{author}{Mensink, T.},
  \bibinfo{year}{2020}.
\newblock \bibinfo{title}{On the benefit of adversarial training for monocular
  depth estimation}.
\newblock \bibinfo{journal}{CVIU} \bibinfo{volume}{190},
  \bibinfo{pages}{102848}.
%Type = Inproceedings
\bibitem[{Gruber et~al.(2019)Gruber, Bijelic, Heide, Ritter and
  Dietmayer}]{depth_evaluation_realistic}
\bibinfo{author}{Gruber, T.}, \bibinfo{author}{Bijelic, M.},
  \bibinfo{author}{Heide, F.}, \bibinfo{author}{Ritter, W.},
  \bibinfo{author}{Dietmayer, K.}, \bibinfo{year}{2019}.
\newblock \bibinfo{title}{Pixel-accurate depth evaluation in realistic driving
  scenarios}, in: \bibinfo{booktitle}{3DV}.
%Type = Inproceedings
\bibitem[{Guizilini et~al.(2021)Guizilini, Ambrus, Burgard and
  Gaidon}]{packnet-san}
\bibinfo{author}{Guizilini, V.}, \bibinfo{author}{Ambrus, R.},
  \bibinfo{author}{Burgard, W.}, \bibinfo{author}{Gaidon, A.},
  \bibinfo{year}{2021}.
\newblock \bibinfo{title}{Sparse auxiliary networks for unified monocular depth
  prediction and completion}, in: \bibinfo{booktitle}{CVPR}.
%Type = Inproceedings
\bibitem[{Guizilini et~al.(2020a)Guizilini, Ambrus, Pillai, Raventos and
  Gaidon}]{packnet}
\bibinfo{author}{Guizilini, V.}, \bibinfo{author}{Ambrus, R.},
  \bibinfo{author}{Pillai, S.}, \bibinfo{author}{Raventos, A.},
  \bibinfo{author}{Gaidon, A.}, \bibinfo{year}{2020}a.
\newblock \bibinfo{title}{{3D} packing for self-supervised monocular depth
  estimation}, in: \bibinfo{booktitle}{CVPR}.
%Type = Inproceedings
\bibitem[{Guizilini et~al.(2020b)Guizilini, Hou, Li, Ambrus and
  Gaidon}]{packnet-semguided}
\bibinfo{author}{Guizilini, V.}, \bibinfo{author}{Hou, R.},
  \bibinfo{author}{Li, J.}, \bibinfo{author}{Ambrus, R.},
  \bibinfo{author}{Gaidon, A.}, \bibinfo{year}{2020}b.
\newblock \bibinfo{title}{Semantically-guided representation learning for
  self-supervised monocular depth}, in: \bibinfo{booktitle}{ICLR}.
%Type = Inproceedings
\bibitem[{Guizilini et~al.(2019)Guizilini, Li, Ambrus, Pillai and
  Gaidon}]{packnet-semisup}
\bibinfo{author}{Guizilini, V.}, \bibinfo{author}{Li, J.},
  \bibinfo{author}{Ambrus, R.}, \bibinfo{author}{Pillai, S.},
  \bibinfo{author}{Gaidon, A.}, \bibinfo{year}{2019}.
\newblock \bibinfo{title}{Robust semi-supervised monocular depth estimation
  with reprojected distances}, in: \bibinfo{booktitle}{CoRL}.
%Type = Inproceedings
\bibitem[{He et~al.(2016)He, Zhang, Ren and Sun}]{resnet}
\bibinfo{author}{He, K.}, \bibinfo{author}{Zhang, X.}, \bibinfo{author}{Ren,
  S.}, \bibinfo{author}{Sun, J.}, \bibinfo{year}{2016}.
\newblock \bibinfo{title}{Deep residual learning for image recognition}, in:
  \bibinfo{booktitle}{CVPR}.
%Type = Inproceedings
\bibitem[{Jaritz et~al.(2018)Jaritz, de~Charette, Wirbel, Perrotton and
  Nashashibi}]{jaritz_sparse_and_dense}
\bibinfo{author}{Jaritz, M.}, \bibinfo{author}{de~Charette, R.},
  \bibinfo{author}{Wirbel, {\'{E}}.}, \bibinfo{author}{Perrotton, X.},
  \bibinfo{author}{Nashashibi, F.}, \bibinfo{year}{2018}.
\newblock \bibinfo{title}{Sparse and dense data with {CNNs}: {D}epth completion
  and semantic segmentation}, in: \bibinfo{booktitle}{3DV}.
%Type = Inproceedings
\bibitem[{Jaritz et~al.(2020)Jaritz, Vu, de~Charette, Wirbel and
  P{\'e}rez}]{xmuda}
\bibinfo{author}{Jaritz, M.}, \bibinfo{author}{Vu, T.H.},
  \bibinfo{author}{de~Charette, R.}, \bibinfo{author}{Wirbel, E.},
  \bibinfo{author}{P{\'e}rez, P.}, \bibinfo{year}{2020}.
\newblock \bibinfo{title}{{xMUDA}: Cross-modal unsupervised domain adaptation
  for {3D} semantic segmentation}, in: \bibinfo{booktitle}{CVPR}.
%Type = Inproceedings
\bibitem[{Kendall et~al.(2017)Kendall, Martirosyan, Dasgupta and
  Henry}]{kendall_stereo}
\bibinfo{author}{Kendall, A.}, \bibinfo{author}{Martirosyan, H.},
  \bibinfo{author}{Dasgupta, S.}, \bibinfo{author}{Henry, P.},
  \bibinfo{year}{2017}.
\newblock \bibinfo{title}{End-to-end learning of geometry and context for deep
  stereo regression}, in: \bibinfo{booktitle}{ICCV}.
%Type = Inproceedings
\bibitem[{Kingma and Ba(2015)}]{Adam}
\bibinfo{author}{Kingma, D.P.}, \bibinfo{author}{Ba, J.}, \bibinfo{year}{2015}.
\newblock \bibinfo{title}{Adam: {A} method for stochastic optimization}, in:
  \bibinfo{booktitle}{ICLR}.
%Type = Inproceedings
\bibitem[{Koestler et~al.(2020)Koestler, Yang, Wang and
  Cremers}]{3DdetectWithoutLabels}
\bibinfo{author}{Koestler, L.}, \bibinfo{author}{Yang, N.},
  \bibinfo{author}{Wang, R.}, \bibinfo{author}{Cremers, D.},
  \bibinfo{year}{2020}.
\newblock \bibinfo{title}{Learning monocular {3D} vehicle detection without
  {3D} bounding box labels}, in: \bibinfo{booktitle}{GCPR}.
%Type = Inproceedings
\bibitem[{Kumar et~al.(2018)Kumar, Milz, Witt, Simon, Amende, Petzold, Yogamani
  and Pech}]{monocular_fisheye_camera}
\bibinfo{author}{Kumar, V.R.}, \bibinfo{author}{Milz, S.},
  \bibinfo{author}{Witt, C.}, \bibinfo{author}{Simon, M.},
  \bibinfo{author}{Amende, K.}, \bibinfo{author}{Petzold, J.},
  \bibinfo{author}{Yogamani, S.K.}, \bibinfo{author}{Pech, T.},
  \bibinfo{year}{2018}.
\newblock \bibinfo{title}{Monocular fisheye camera depth estimation using
  sparse lidar supervision}, in: \bibinfo{booktitle}{IEEE ITSC}.
%Type = Inproceedings
\bibitem[{Kuznietsov et~al.(2017)Kuznietsov, St{\"{u}}ckler and
  Leibe}]{Kuznietsov}
\bibinfo{author}{Kuznietsov, Y.}, \bibinfo{author}{St{\"{u}}ckler, J.},
  \bibinfo{author}{Leibe, B.}, \bibinfo{year}{2017}.
\newblock \bibinfo{title}{Semi-supervised deep learning for monocular depth map
  prediction}, in: \bibinfo{booktitle}{CVPR}.
%Type = Article
\bibitem[{Lee and Medioni(2016)}]{RGBD_nav_impaired}
\bibinfo{author}{Lee, Y.H.}, \bibinfo{author}{Medioni, G.},
  \bibinfo{year}{2016}.
\newblock \bibinfo{title}{Rgb-d camera based wearable navigation system for the
  visually impaired}.
\newblock \bibinfo{journal}{CVIU} \bibinfo{volume}{149},
  \bibinfo{pages}{3--20}.
%Type = Article
\bibitem[{Lepetit et~al.(2009)Lepetit, Moreno-Noguer and Fua}]{EPnP}
\bibinfo{author}{Lepetit, V.}, \bibinfo{author}{Moreno-Noguer, F.},
  \bibinfo{author}{Fua, P.}, \bibinfo{year}{2009}.
\newblock \bibinfo{title}{Epnp: An accurate o(n) solution to the pnp problem}.
\newblock \bibinfo{journal}{IJCV} \bibinfo{volume}{81},
  \bibinfo{pages}{155--166}.
%Type = Article
\bibitem[{Lowe(2004)}]{sift_Lowe}
\bibinfo{author}{Lowe, D.G.}, \bibinfo{year}{2004}.
\newblock \bibinfo{title}{Distinctive image features from scale-invariant
  keypoints}.
\newblock \bibinfo{journal}{IJCV} \bibinfo{volume}{60},
  \bibinfo{pages}{91--110}.
%Type = Inproceedings
\bibitem[{Loza et~al.(2006)Loza, Mihaylova, Canagarajah and Bull}]{ssim}
\bibinfo{author}{Loza, A.}, \bibinfo{author}{Mihaylova, L.},
  \bibinfo{author}{Canagarajah, N.}, \bibinfo{author}{Bull, D.R.},
  \bibinfo{year}{2006}.
\newblock \bibinfo{title}{Structural similarity-based object tracking in video
  sequences}, in: \bibinfo{booktitle}{IEEE FUSION}.
%Type = Inproceedings
\bibitem[{Ma et~al.(2019)Ma, Cavalheiro and Karaman}]{selfsup_sparse_to_dense}
\bibinfo{author}{Ma, F.}, \bibinfo{author}{Cavalheiro, G.V.},
  \bibinfo{author}{Karaman, S.}, \bibinfo{year}{2019}.
\newblock \bibinfo{title}{Self-supervised sparse-to-dense: {S}elf-supervised
  depth completion from {LiDAR} and monocular camera}, in:
  \bibinfo{booktitle}{ICRA}.
%Type = Inproceedings
\bibitem[{Ma and Karaman(2018)}]{sparse_to_dense}
\bibinfo{author}{Ma, F.}, \bibinfo{author}{Karaman, S.}, \bibinfo{year}{2018}.
\newblock \bibinfo{title}{Sparse-to-dense: Depth prediction from sparse depth
  samples and a single image}, in: \bibinfo{booktitle}{ICRA}.
%Type = Inproceedings
\bibitem[{Mahjourian et~al.(2018)Mahjourian, Wicke and Angelova}]{Vid2Depth}
\bibinfo{author}{Mahjourian, R.}, \bibinfo{author}{Wicke, M.},
  \bibinfo{author}{Angelova, A.}, \bibinfo{year}{2018}.
\newblock \bibinfo{title}{Unsupervised learning of depth and ego-motion from
  monocular video using {3D} geometric constraints}, in:
  \bibinfo{booktitle}{CVPR}.
%Type = Article
\bibitem[{Mohan and Valada(2021)}]{mohan2020efficientps}
\bibinfo{author}{Mohan, R.}, \bibinfo{author}{Valada, A.},
  \bibinfo{year}{2021}.
\newblock \bibinfo{title}{{EfficientPS}: {E}fficient panoptic segmentation}.
\newblock \bibinfo{journal}{IJCV} \bibinfo{volume}{129}, \bibinfo{pages}{1551
  -- 1579}.
%Type = Inproceedings
\bibitem[{Ng et~al.(2020)Ng, Radia, Chen, Wang, Gog and Gonzalez}]{bevseg}
\bibinfo{author}{Ng, M.}, \bibinfo{author}{Radia, K.}, \bibinfo{author}{Chen,
  J.}, \bibinfo{author}{Wang, D.}, \bibinfo{author}{Gog, I.},
  \bibinfo{author}{Gonzalez, J.}, \bibinfo{year}{2020}.
\newblock \bibinfo{title}{{BEV-Seg}: {B}ird's eye view semantic segmentation
  using geometry and semantic point cloud}, in: \bibinfo{booktitle}{CVPR}.
%Type = Inproceedings
\bibitem[{Park et~al.(2020)Park, Joo, Hu, Liu and
  Kweon}]{non_local_spatial_propagation}
\bibinfo{author}{Park, J.}, \bibinfo{author}{Joo, K.}, \bibinfo{author}{Hu,
  Z.}, \bibinfo{author}{Liu, C.K.}, \bibinfo{author}{Kweon, I.S.},
  \bibinfo{year}{2020}.
\newblock \bibinfo{title}{Non-local spatial propagation network for depth
  completion}, in: \bibinfo{booktitle}{ECCV}.
%Type = Inproceedings
\bibitem[{Philion and Fidler(2020)}]{lift_splat_shoot}
\bibinfo{author}{Philion, J.}, \bibinfo{author}{Fidler, S.},
  \bibinfo{year}{2020}.
\newblock \bibinfo{title}{Lift, splat, shoot: Encoding images from arbitrary
  camera rigs by implicitly unprojecting to {3D}}, in:
  \bibinfo{booktitle}{ECCV}.
%Type = Inproceedings
\bibitem[{Srikanth et~al.(2019)Srikanth, Ansari, R., Sharma, Murthy and
  Krishna}]{infer}
\bibinfo{author}{Srikanth, S.}, \bibinfo{author}{Ansari, J.A.},
  \bibinfo{author}{R., K.R.}, \bibinfo{author}{Sharma, S.},
  \bibinfo{author}{Murthy, J.K.}, \bibinfo{author}{Krishna, K.M.},
  \bibinfo{year}{2019}.
\newblock \bibinfo{title}{{INFER:} {IN}termediate representations for {FuturE}
  {pRediction}}, in: \bibinfo{booktitle}{IROS}.
%Type = Inproceedings
\bibitem[{Sun et~al.(2020)Sun, Kretzschmar, Dotiwalla, Chouard, Patnaik, Tsui,
  Guo, Zhou, Chai, Caine et~al.}]{waymo_opendataset}
\bibinfo{author}{Sun, P.}, \bibinfo{author}{Kretzschmar, H.},
  \bibinfo{author}{Dotiwalla, X.}, \bibinfo{author}{Chouard, A.},
  \bibinfo{author}{Patnaik, V.}, \bibinfo{author}{Tsui, P.},
  \bibinfo{author}{Guo, J.}, \bibinfo{author}{Zhou, Y.}, \bibinfo{author}{Chai,
  Y.}, \bibinfo{author}{Caine, B.}, et~al., \bibinfo{year}{2020}.
\newblock \bibinfo{title}{Scalability in perception for autonomous driving:
  Waymo open dataset}, in: \bibinfo{booktitle}{CVPR}.
%Type = Article
\bibitem[{Tang et~al.(2020)Tang, Tian, Feng, Li and
  Tan}]{guided_depth_completion}
\bibinfo{author}{Tang, J.}, \bibinfo{author}{Tian, F.P.},
  \bibinfo{author}{Feng, W.}, \bibinfo{author}{Li, J.}, \bibinfo{author}{Tan,
  P.}, \bibinfo{year}{2020}.
\newblock \bibinfo{title}{Learning guided convolutional network for depth
  completion}.
\newblock \bibinfo{journal}{IEEE TIP} .
%Type = Inproceedings
\bibitem[{Uhrig et~al.(2017)Uhrig, Schneider, Schneider, Franke, Brox and
  Geiger}]{sparsity_invariant_cnns}
\bibinfo{author}{Uhrig, J.}, \bibinfo{author}{Schneider, N.},
  \bibinfo{author}{Schneider, L.}, \bibinfo{author}{Franke, U.},
  \bibinfo{author}{Brox, T.}, \bibinfo{author}{Geiger, A.},
  \bibinfo{year}{2017}.
\newblock \bibinfo{title}{Sparsity invariant {CNNs}}, in:
  \bibinfo{booktitle}{3DV}.
%Type = Inproceedings
\bibitem[{Wang et~al.(2018)Wang, Buenaposada, Zhu and Lucey}]{DDVO}
\bibinfo{author}{Wang, C.}, \bibinfo{author}{Buenaposada, J.M.},
  \bibinfo{author}{Zhu, R.}, \bibinfo{author}{Lucey, S.}, \bibinfo{year}{2018}.
\newblock \bibinfo{title}{Learning depth from monocular videos using direct
  methods}, in: \bibinfo{booktitle}{CVPR}.
%Type = Inproceedings
\bibitem[{Watson et~al.(2019)Watson, Firman, Brostow and
  Turmukhambetov}]{depth_hints}
\bibinfo{author}{Watson, J.}, \bibinfo{author}{Firman, M.},
  \bibinfo{author}{Brostow, G.J.}, \bibinfo{author}{Turmukhambetov, D.},
  \bibinfo{year}{2019}.
\newblock \bibinfo{title}{Self-supervised monocular depth hints}, in:
  \bibinfo{booktitle}{ICCV}.
%Type = Inproceedings
\bibitem[{Xu et~al.(2019)Xu, Zhu, Shi, Zhang, Bao and
  Li}]{Depth_Normal_Constraints}
\bibinfo{author}{Xu, Y.}, \bibinfo{author}{Zhu, X.}, \bibinfo{author}{Shi, J.},
  \bibinfo{author}{Zhang, G.}, \bibinfo{author}{Bao, H.}, \bibinfo{author}{Li,
  H.}, \bibinfo{year}{2019}.
\newblock \bibinfo{title}{Depth completion from sparse {LiDAR} data with
  depth-normal constraints}, in: \bibinfo{booktitle}{ICCV}.
%Type = Article
\bibitem[{Yamanaka et~al.(2020)Yamanaka, Matsumoto, Takahashi and
  Fujii}]{adversarial_patch}
\bibinfo{author}{Yamanaka, K.}, \bibinfo{author}{Matsumoto, R.},
  \bibinfo{author}{Takahashi, K.}, \bibinfo{author}{Fujii, T.},
  \bibinfo{year}{2020}.
\newblock \bibinfo{title}{Adversarial patch attacks on monocular depth
  estimation networks}.
\newblock \bibinfo{journal}{{IEEE} Access} \bibinfo{volume}{8}.
%Type = Inproceedings
\bibitem[{Yin and Shi(2018)}]{GeoNet}
\bibinfo{author}{Yin, Z.}, \bibinfo{author}{Shi, J.}, \bibinfo{year}{2018}.
\newblock \bibinfo{title}{{GeoNet}: {U}nsupervised learning of dense depth,
  optical flow and camera pose}, in: \bibinfo{booktitle}{CVPR}.
%Type = Inproceedings
\bibitem[{Zeng et~al.(2019)Zeng, Luo, Suo, Sadat, Yang, Casas and
  Urtasun}]{neural_motion_paper}
\bibinfo{author}{Zeng, W.}, \bibinfo{author}{Luo, W.}, \bibinfo{author}{Suo,
  S.}, \bibinfo{author}{Sadat, A.}, \bibinfo{author}{Yang, B.},
  \bibinfo{author}{Casas, S.}, \bibinfo{author}{Urtasun, R.},
  \bibinfo{year}{2019}.
\newblock \bibinfo{title}{End-to-end interpretable neural motion planner}, in:
  \bibinfo{booktitle}{CVPR}.
%Type = Article
\bibitem[{Zhao et~al.(2021)Zhao, Gong, Fu and Tao}]{ACMNet}
\bibinfo{author}{Zhao, S.}, \bibinfo{author}{Gong, M.}, \bibinfo{author}{Fu,
  H.}, \bibinfo{author}{Tao, D.}, \bibinfo{year}{2021}.
\newblock \bibinfo{title}{Adaptive context-aware multi-modal network for depth
  completion}.
\newblock \bibinfo{journal}{IEEE TIP} .
%Type = Inproceedings
\bibitem[{Zhou et~al.(2017)Zhou, Brown, Snavely and Lowe}]{sfm_learner}
\bibinfo{author}{Zhou, T.}, \bibinfo{author}{Brown, M.},
  \bibinfo{author}{Snavely, N.}, \bibinfo{author}{Lowe, D.G.},
  \bibinfo{year}{2017}.
\newblock \bibinfo{title}{Unsupervised learning of depth and ego-motion from
  video}, in: \bibinfo{booktitle}{CVPR}.

\end{thebibliography}

%%%%%%%%%%%%%%%%%%%%%%%%%%%%%%%%%%%%%%%%%%%%%%%%%%%%%%%%%%%%%%%%%%
%%%%%%%%%%%%%%%% APPENDIX %%%%%%%%%%%%%%%%%%%%%%%%%%%%%%%%%%%%%%
%%%%%%%%%%%%%%%%%%%%%%%%%%%%%%%%%%%%%%%%%%%%%%%%%%%%%%%%%%%%%%%%%

\appendix
%\section{Appendix}

\section{Implementation details}
\label{app:implem_details}

%%%%%%%%%%%%%%%
%%%%%% TRAINING
%%%%%%%%%%%%%%%
\parag{Training.} 
We train all our models for 30 epochs using the Adam optimizer~\citep{Adam} with $\beta_1 = 0.9$ and $\beta_2 = 0.999$. The initial learning rate is set to $10^{-4}$ and divided by two halfway through training.

% Regularization
In all training pipelines, following common practice~\citep{monodepth17,packnet,monodepth2}, we add an edge-aware smoothing regularization loss to encourage the predicted depth map $\hat{D}_{\tgt}$ to be locally smooth while taking into account sharp boundaries:
\begin{equation}
    \label{eq:smooth}
    L_\text{smooth} = \lvert \partial_x \hat{D}_{\tgt} \rvert e^{- \lvert \partial_x I_{\tgt} \rvert}  + \lvert \partial_y \hat{D}_{\tgt} \rvert e^{- \lvert \partial_y I_{\tgt} \rvert}, 
\end{equation}
with the index $p$ over pixels omitted for clarity.

%%%%%%%%%%%%%%%
%%%%%% Monodepth2-L
%%%%%%%%%%%%%%%
\parag{Monodepth2 extension.} 
Our \imagesparselidar{} architecture is similar to \imagealone{} at the difference that we use a second ResNet-18 encoder specifically for the LiDAR modality. We only remove the first batch-normalization layer of the LiDAR ResNet, as using it would imply the computation of ineffective statistics given that the LiDAR input mostly contains zeros (encoding measurement absence).

%%%%%%%%%%%%%%%
%%%%%% POSE
%%%%%%%%%%%%%%%
\parag{Pose estimation.}
To solve the \pnp{} problem, we use an open-source implementation of \pnp{} methods with RANSAC from the OpenCV library~\citep{opencv_library}.
We use 100 iterations and a reprojection error threshold of 2.
Even after RANSAC, the remaining outliers are numerous enough to hinder training. Therefore, we remove the relative pose estimates for which the translation magnitude $\lVert \hat{r} \rVert$ is too large. In effect, we first compute the median value of translation magnitude for each relative pose of the train set. Then, we remove all examples that are too far-off the median.
When using a pose network, we follow~\citep{monodepth2} and use a ResNet-18 taking two images in input and outputting the parameters of $\hat{P}_{\tgt \shortrightarrow \src}$, the rigid transformation between the two views.

%%%%%%%%%%%%%%%
%%%%%% Evaluation
%%%%%%%%%%%%%%%
\parag{Evaluation after rescaling.} 
% RESCALING
Baselines and models from prior works that only provide relative-depth maps have their predictions rescaled so that they have the same mean compared to the ground truth against which they are evaluated. This is mentioned as `\emph{gt rescaled}' in \autoref{tab:sota}. For methods that directly produce metric depth maps, like ours, we do not apply this post-processing procedure and depth maps are kept at the originally-predicted scale.

%%%%%%%%%%%%%%%
%%%%%% METRIC
%%%%%%%%%%%%%%%
\parag{CDR Metric.} 
To compute results with our CDR metric, we first extract instance masks with EfficientPS~\citep{mohan2020efficientps}. 
Among these masks, we want to focus only on those of close-by, non-occluded vehicles, \ie{} first vehicles in front of the ego-car. These vehicles are particularly prone to infinite-depth mistakes, with safety-critical consequences when it happens.
To do this selection, vehicles that are not in front of the ego-car are discarded, as measured by not belonging to the central band of the scene (size is 20\% of the image width) captured by the front camera.
Vehicle having instance masks calculated with fewer than 20 pixels are considered too far from the ego vehicle.
Then, to assess whether a car is occluded or not, we assume that a heavily occluded vehicle generally has a non-convex shape (\eg incised by the front vehicle) and that, on the contrary, the  mask of a non-occluded car is approximately convex. 
Accordingly, we first smooth segmentation masks and fill noisy areas where the intensity changes rapidly (\eg edges, small holes from the wheels) by applying a  morphological \emph{dilation} operator. We use a square kernel of size 10 and 4 iterations for this operation.
The masks now being smoothed, we then approach their shape by a polygon from which we can tell if they are convex or not. 
To approximate each pixel blob  by a polygon, we use the Douglas–Peucker algorithm~\citep{douglas_peucker}. The algorithm ensures the fit of the approximated polygon with an accuracy parameter dependent of the pixel blob size.
After this first filtering step, 657 valid masks remain out of the 4460 vehicle masks of the KITTI test split. 

%%%%%%%%%%%%%%%
%%%%%% Artificial sparsification
%%%%%%%%%%%%%%%
\parag{Extracting 4 beams from 64-beam point clouds.}
In the KITTI dataset, the LiDAR data in a frame is provided as a unique point cloud, that is, a set of $(x,y,z)$ coordinates, without the beam indexes, \ie which of the 64 lasers has been used for each measurement.
We needed to recover this information for our experiments. 
Fortunately, in KITTI the points are recorded in an orderly manner.
The points of one beam follow the points of another in the direction of laser rotation (counter-clockwise).
This means that, inside the data stream of a same frame, each rotation completion indicates a change of beam. 
More precisely, the coordinate basis of the LiDAR is oriented with $x$: positive forward and $y$: positive to the left of the car. Then we can compute the horizontal angle in radian of each point with:
\begin{equation}
    \phi = \text{arctan2}(y,x).
\end{equation}
We use the 2-argument arctangent instead of classic arctangent, $\text{arctan}(y/x)$, as the latter cannot distinguish between diametrically opposite directions.
Then, by computing the horizontal angle (azimuth) of each point, we can separate data for each beam by detecting when $\phi$ changes from $360^{\circ}$ to $0^{\circ}$ in the stream of points.
This way, we have access to the ring index for each LiDAR point and can, thus, freely sparsify the LiDAR data.

% CODE RELEASED 
\parag{Code release.}
To enable comparison with our work in the future, all the processing steps described above will be included in the source code we plan to release.
We will also release pre-trained models with our code for training and evaluating them.

%%%%%%%%%%%%%%%
%%%%%% Overfitted convergence
%%%%%%%%%%%%%%%
\section{Overfitting to input LiDAR}
\label{app:overfit}

\begin{figure*}[t]
 \centering
 \begin{subfigure}[h]{0.49\linewidth}
     \centering
     \includegraphics[trim = 0 0 1.5cm 0, clip,width=\linewidth]{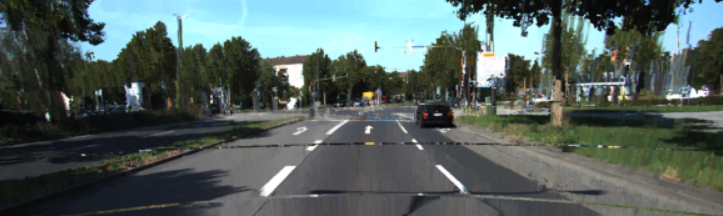}
     \caption{Reconstructed view from source image, depth and pose predictions}
     \label{fig:overfit:reconstruction}
 \end{subfigure}
 \hfill
 \begin{subfigure}[h]{0.49\linewidth}
     \centering
     \includegraphics[trim = 0 0 1.5cm 0, clip,width=\linewidth]{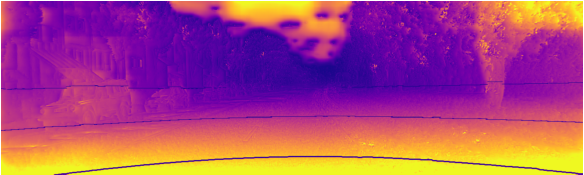}
     \caption{Deth estimation}
     \label{fig:overfit:pred}
 \end{subfigure}
 \\[2ex] % more vertical space between the figures
  \begin{subfigure}[h]{0.49\linewidth}
     \includegraphics[trim = 0 0 1.5cm 0, clip,width=\linewidth]{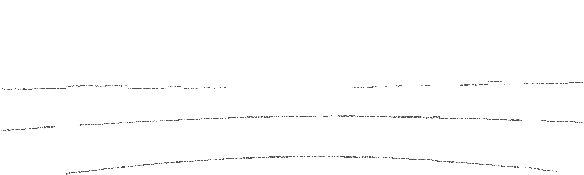}
     \caption{Mask on pixels without LiDAR data}
     \label{fig:overfit:mask_wo_lidar}
 \end{subfigure}
 \hfill
 \begin{subfigure}[h]{0.49\linewidth}
     \includegraphics[trim = 0 0 1.5cm 0, clip,width=\linewidth]{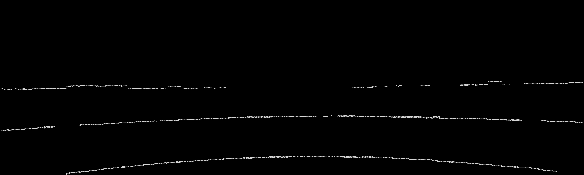}
     \caption{Mask on pixels with LiDAR data}
     \label{fig:overfit:mask_w_lidar}
 \end{subfigure}

\caption{\textbf{Predictions of a model overfitting to LiDAR input.}
We show (a) a reconstructed view from source image, depth and pose prediction (b) a depth estimation considered as \emph{overfitted} to the LiDAR input (c) a binary mask where the value is 1 for pixels without LiDAR data and 0 otherwise (d) a binary mask where the value is 1 for pixels with LiDAR data and 0 otherwise.
}
\label{fig:overfit}
\end{figure*}

\begin{figure}[] 
\centering
\begin{tikzpicture}[
    image/.style = {
        text width=\linewidth,
        inner sep=0pt, 
        outer sep=0pt,
        },
    label/.style = {
        inner sep=2pt,
        font=\small,
        align=center,
        color=white
    },
    node distance = 1mm and 1mm
]

\node [image] (med_disp_pred)
{\includegraphics[width=\linewidth]{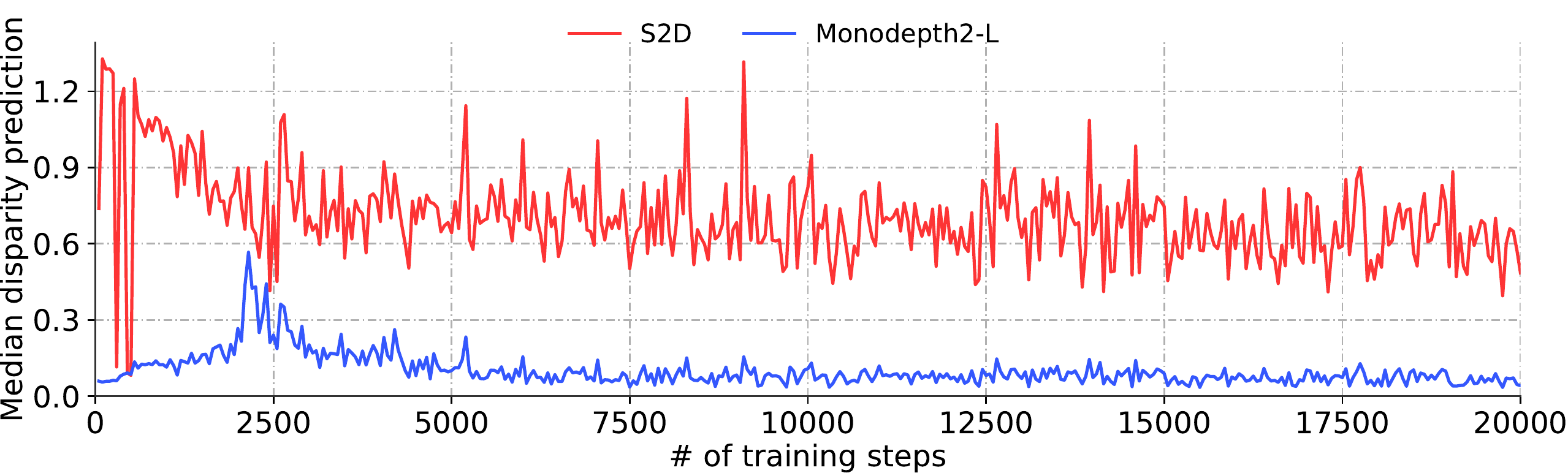}};

\node [image, below=of med_disp_pred] (masked_med_disp_pred) {\includegraphics[width=\linewidth]{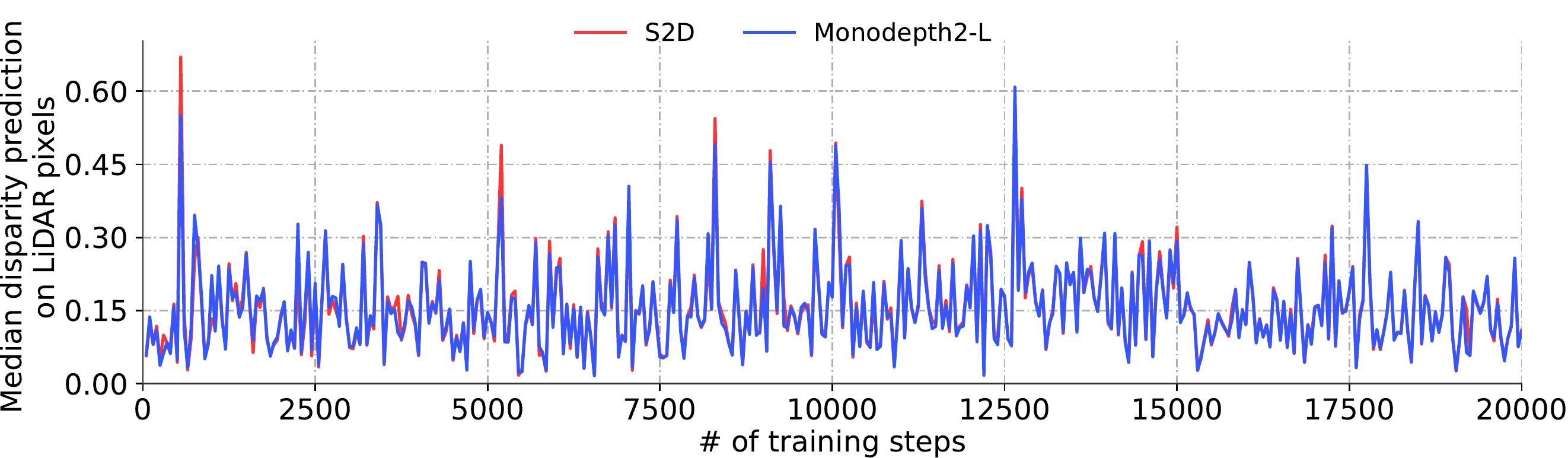}};

\node [image, below=of masked_med_disp_pred] (pose) {\includegraphics[width=\linewidth]{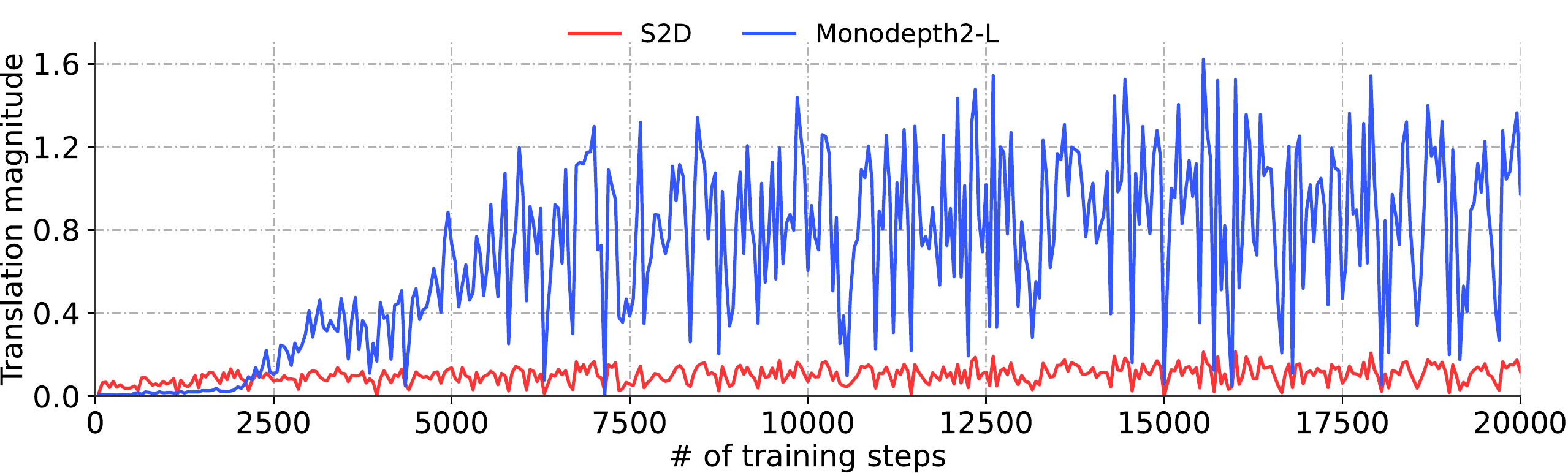}};

\end{tikzpicture}
\caption{\textbf{Statistics for the depth and pose outputs over a training run for S2D (\emph{overfitted}) and Monodepth2-L (\emph{metric}) with a pose network and (`P+L$_4$') supervision.} Respectively from top to bottom, we provide the median value of the disparity predicted by the depth network on pixels without LiDAR data (see \autoref{fig:overfit:mask_wo_lidar}), then on pixels with LiDAR data (see \autoref{fig:overfit:mask_w_lidar}), and the magnitude of the relative pose's translation component (i.e., by how much the pose network estimates the car moved between two views).}
\label{fig:overfit_pred_stats}
\end{figure}

In this section, we provide qualitative examples as well as elements of analysis for the convergence behavior observed on ACMNet, NLSPN and S2D that we call `\emph{overfitted} to LiDAR input'. 
To this end, we compare S2D (\emph{overfitted}) to Monodepth2-L (\emph{metric}) trained with a pose network and (`P+L$_4$') supervision for 30 epochs. 
In essence, we refer to models as \emph{overfitted} when most of the depth prediction is consistent but only \emph{relative}, while depth prediction is only metric on pixels with LiDAR data.
On \autoref{fig:overfit:pred}, we can clearly observe the difference in scale between areas with and without LiDAR data.
Likewise, we can quantitatively observe the existence of two distinct scales within predictions of S2D. In the middle plot of \autoref{fig:overfit_pred_stats}, the median value of the inverse depth prediction (disparity) on pixels with LiDAR are roughly the same for S2D and \imagesparselidar{}, they are both scaled metrically. On the other hand, in the top plot of \autoref{fig:overfit_pred_stats} showing the median value of disparity on pixels without LiDAR, there is a clear difference between \imagesparselidar{}, that is properly scaled, and S2D that converged to a random scale.

\begin{figure}[] 
\centering
\begin{tikzpicture}[
    image/.style = {
        text width=\linewidth,
        inner sep=0pt, 
        outer sep=0pt,
        },
    label/.style = {
        inner sep=2pt,
        font=\small,
        align=center,
        color=white
    },
    node distance = 1mm and 1mm
]

\node [image] (lidar)
{\includegraphics[width=\linewidth]{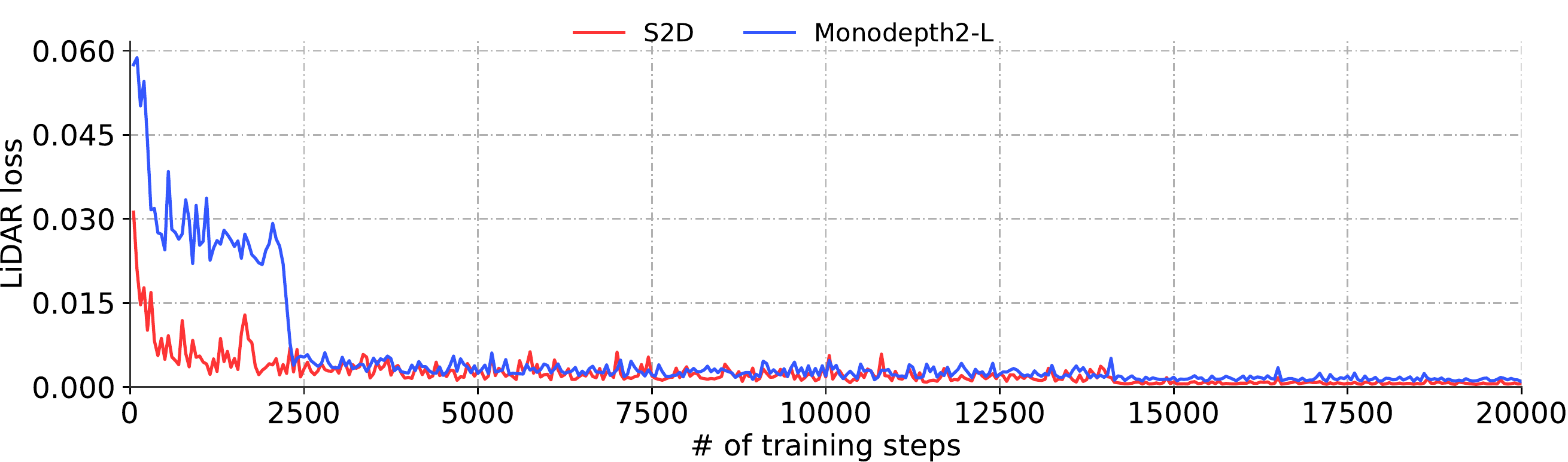}};

\node [image, below=of lidar] (photo) {\includegraphics[width=\linewidth]{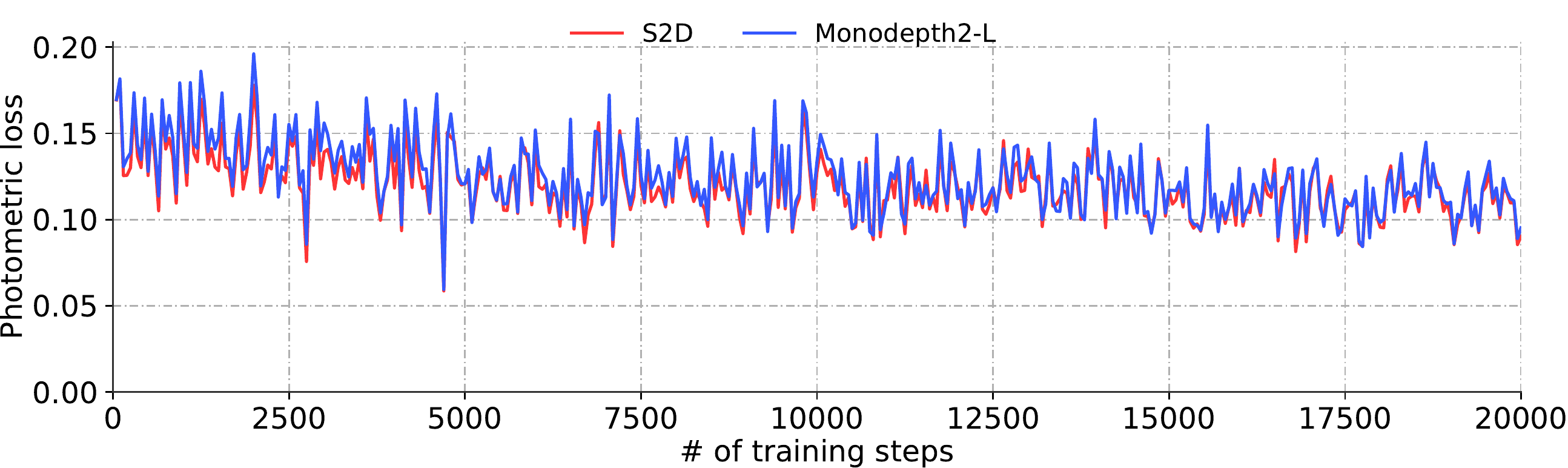}};

\node [image, below=of photo] (smooth) {\includegraphics[width=\linewidth]{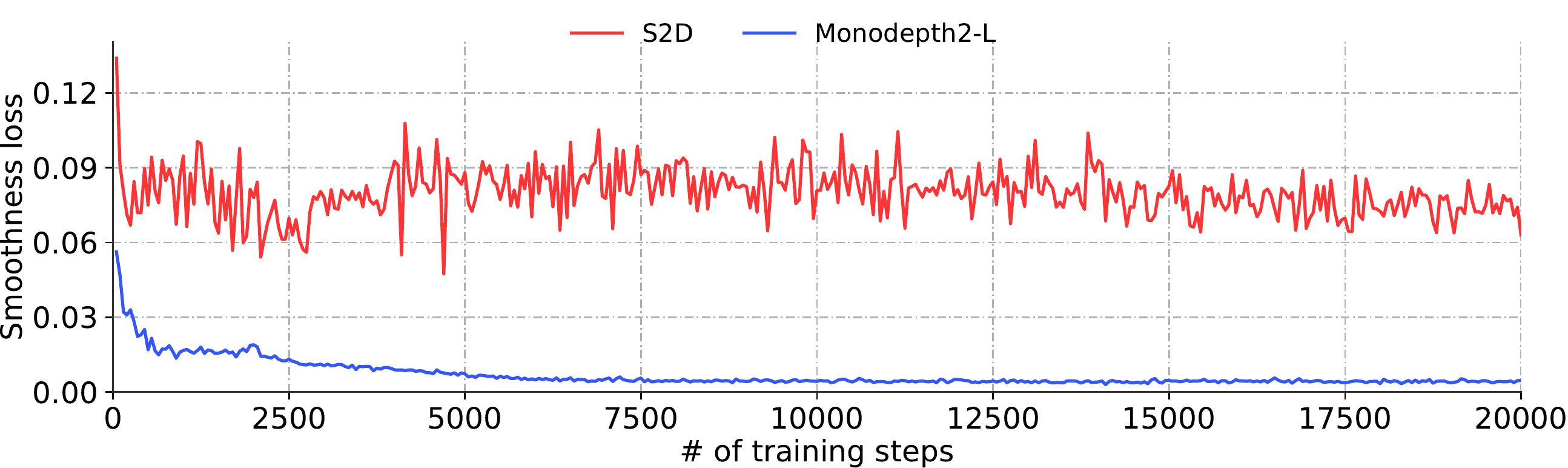}};

\end{tikzpicture}
\caption{\textbf{Loss values over a training run for S2D (\emph{overfitted}) and Monodepth2-L (\emph{metric}) with a pose network and (`P+L$_4$') supervision.} Respectively from top to bottom, we provide the values over a training run for the self-supervised LiDAR loss, the photometric loss as well as the smoothness loss. }
\label{fig:overfit_quantitative_losses}
\end{figure}

From a supervisory perspective, the depth network is stuck within a local minima where the photometric loss is mostly minimized apart on pixels with LiDAR data where it is clear the pixels are projected at different scale (see \autoref{fig:overfit:reconstruction}). The amount of pixels with LiDAR data being very small, the erroneous photometric loss is on these areas is strongly dampened by the average over the whole image. So strongly dampened that that photometric loss between S2D and \imagesparselidar{}, respectively an \emph{overfitted} and a \emph{metric} model, almost perfectly match (see photometric loss~\autoref{fig:overfit_quantitative_losses}). At the same time the LiDAR loss has already reached a minimum and the smoothness loss is not powerful enough to regularize this convergence behavior.

This convergence profile is expected because there are an infinite number of depth prediction scales for which the photometric loss is minimized over areas with no LiDAR data. Hence, there is an infinite number of local minima leading to this \emph{overfitted} behavior. On the contrary, when using LiDAR self-supervision, only one depth prediction scale exists, the \emph{metric} one, to obtain a globally coherent reconstruction. We propose a solution to this problem for S2D as well as ACMNet and NLSPN in \autoref{app:dilated}.

%%%%%%%%%%%%%%%
%%%%%% Ablation of LiDAR - further analysis
%%%%%%%%%%%%%%%
\section{Ablation of LiDAR: further analysis}
\label{app:ablation}

In this section, we analyse results for learning setups not described in \autoref{sec:lidar_influence:ablative_study}. In particular, we continue to study the use of a pose network instead of \pnp{}, with `P', `P+IMU' or `P+L$_4$+IMU' supervision.

Overall, we observe very poor performances with the use of the pose network.
First, we note that the use of photometric reconstruction only (`P' in \autoref{tab:scaling}) leads to to \textbf{relative} depth for all networks (dark gray cells in the \autoref{tab:scaling}). Indeed, this setup is well known as being an ill-posed problem \citep{sfm_learner, monodepth2, DDVO, packnet}; the pose provided by the monocular pose network can only be relative without additional information, and the depth estimation is thus unscaled as well.

To enforce a metric scale, we train the pose network with additional supervision in the form of an IMU prior (`P+IMU'), as explained in \autoref{sec:protocol:pose_baselines}.
While this helps \imagealone{} and \imagesparselidar{} to correctly train, ACMNet, NLSPN and S2D architectures cannot reach good performances when a joint alignment between a pose and depth network is required (see~\autoref{app:imu_prior} for more details). 

With further supervision from the input LiDAR (`P+L$_4$+IMU'), we can slightly increase results for \imagealone{} and \imagesparselidar{} as well as significantly boosting results for ACMNet compared to the (`P+IMU') setup (253\% increase). However, similar to ACMNet, NLSPN and S2D in the (`P+L$_4$') setup (see \autoref{sec:lidar_influence:ablative_study}), NLSPN and S2D tends to overfit the input LiDAR. Hence, we use the same \emph{dilation} procedure, as detailed in \autoref{app:dilated}, for these models to avoid overfitting the LiDAR input.

%%%%%%%%%%%%%%%
%%%%%% Dilated LiDAR
%%%%%%%%%%%%%%%
\section{Dilated LiDAR}
\label{app:dilated}

Contrarily to \imagealone{} and \imagesparselidar{}, when trained with a pose network and LiDAR self-supervision, the networks ACMNet, NLSPN and S2D tend to overfit the LiDAR. Most of the depth prediction is consistent but only relative, while depth prediction on pixels with LiDAR data is metric (see \autoref{app:overfit}
for an example). 
The main difference between these architectures is that \imagealone{} and \imagesparselidar{} are supervised at multiple scales (1:1, 1:2, 1:4 and 1:8) while ACMNet, NSLPN and S2D are only supervised at the final resolution (1:1). Supervision at the lowest scale (1:8) artificially increases the number of pixels getting supervision from LiDAR as a LiDAR point spans multiple pixels when projected at low resolutions.  

We hypothesize that the mono-scale training is the cause of overfitting to LiDAR input when training with LiDAR self-supervision.
This is confirmed by the fact that, when \imagesparselidar{} is only supervised at the scale 1:1, the model collapses into the \emph{overfitted} regime which highlights the importance of multi-scale training.

\begin{figure}[] 
\centering
\begin{tikzpicture}[
    image/.style = {
        text width=\linewidth,
        inner sep=0pt, 
        outer sep=0pt,
        draw,
        line width=0.3mm
        },
    label/.style = {
        inner sep=2pt,
        font=\small,
        align=center,
        color=white
    },
    node distance = 1mm and 1mm
]

%%%%%%%%%%%%%%%%%%%%%%%%%%  Normal  %%%%%%%%%%%%%%%%%%%%%%%%
    \node [image, line width=0.2mm] (sparse) %{\includegraphics[width=\linewidth]{figures/LIDAR/val_287_sparselidar.png}};
    {\includegraphics[width=\linewidth]{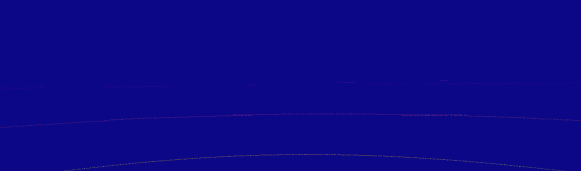}};
    % LEGEND
    \node [label, anchor=north] at ($(sparse.north)$) {\emph{Minimal} / \emph{Sparse} 4-beam};
    
%%%%%%%%%%%%%%%%%%%%%%%% Dilated  %%%%%%%%%%%%%%%%%%%%%%%%
    \node [image, below=of sparse] (dilated) {\includegraphics[width=\linewidth]{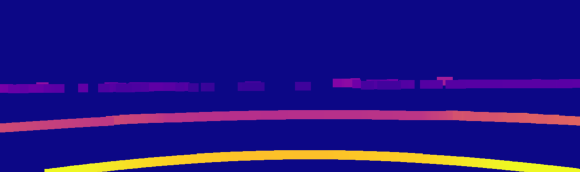}};

    % LEGEND
    \node [label, anchor=north] at ($(dilated.north)$) {\emph{Dilated} 4-beam};

\end{tikzpicture}
\caption{\textbf{Visual difference between vanilla and dilated LiDAR.}}
\label{fig:dilated_lidar}
\end{figure}

As modifying the mono-scale networks is non-trivial, we propose to self-supervise with a dilated LiDAR to compensate for the lack of multi-scale supervision and to avoid overfitting the LiDAR input. More precisely, we apply two iterations of a \emph{dilation} morphological operator with a kernel of $10\times 10$ on the 4-beam LiDAR at the supervision level only (i.o.w., we do not apply \emph{dilation} on the LiDAR input). The aim is to increase the number of pixels receiving LiDAR supervision, albeit in a noisy manner, (\autoref{fig:dilated_lidar}). 
This simple procedure, while remaining a trick, enables mono-scale architectures to avoid overfitting the input LiDAR and to converge to metric depth estimation. On the other hand, none of the architectures need such special care when trained under our LiDARTouch framework. We report results of models trained with this procedure with the superscript $\dagger$ in\autoref{tab:scaling}.

In addition to this strategy, we explored various experimental setups and combination of hyper-parameters when training with (P+L4) and (P+L4+IMU) for mono-scale networks:
\begin{itemize}
    \item Dividing the sparse LiDAR depth values (used as input and/or ground-truth) by a factor $\alpha$ at train time and multiply depth prediction consequently at validation.
    The network still overfits to LiDAR data with $\alpha \in \{10,100,1000\}$.
    \item Decreasing the contribution of the depth loss in the global objective to mitigate the overfitting behavior to LiDAR points. 
    With $\lambda \in \{1, 1e-1, 1e-2, 1e-3\}$, the model still overfits the LiDAR. With $\lambda \in \{1e-4, 1e-5\}$ the network stops overfitting the LiDAR data but the depth estimation becomes only relative instead of being metric.
    \item Increasing the contribution of the smoothness loss in the global objective. By doing so, we hoped to uniformize the scale of the depth prediction on pixels without LiDAR that are neighbors to pixels with LiDAR. The network still overfits to LiDAR data with $\lambda \in \{1e-1, 1e-2, 1e-3\}$.
    \item Varying learning rate from $1e-3$ to $1e-5$. The network still overfits to LiDAR data.
\end{itemize}

%%%%%%%%%%%%%%%
%%%%%% PnP scaling
%%%%%%%%%%%%%%%
\section{Pose scaling is critical when using a \pnp{} pose estimation with photometric loss only}

Most of the depth network's learning signal comes from the reconstruction of the target image from the source image. For a given scale, a correct photometric reconstruction corresponds to a unique pair of depth and pose. Hence, for one to be metrically scaled, both the depth and the pose have to be metric. However, the networks are initialized randomly and thus need to jointly align and converge to a metric scale. 

On the other hand, when using \pnp{}, the estimated pose is metric thanks to LiDAR data (see \autoref{sec:learning_system:pose_setups}), thus, only the depth network has to converge to the correct scale.
However, this may produce a large difference in scale at initialization between the pose and depth, provoking unstable training for the depth network. Thus, one strategy we adopt to stabilize training is to divide the translation component of the \pnp{} pose by 10 and multiply the depth prediction by 10 at inference time. Models trained with this strategy are indicated with the superscript $\ast$ in \autoref{tab:scaling}.

To circumvent these difficult training behaviors, we can use the \pnp{} method to produce metric poses, and further enforce the collapse of the depth solutions to a metric scale with additional LiDAR self-supervision. This is consistently verified with the use of photometric and LiDAR supervisions (P+$\text{L}_4$) for each of the five architectures considered and leads to the best results compared to any other configuration (see \autoref{tab:scaling}). These results demonstrate that the use of LiDAR both as self-supervision and in pose computation yields performance on-par or better than camera-only setups.

\section{Poor performances for ACMNet, NLSPN and S2D when trained with P+IMU}
\label{app:imu_prior}

Unfortunately, we cannot make these models converge to metric depth estimations. We describe below the combination of hyper-parameters we experimented with:
\begin{itemize}
    \item Dividing the pose GT (translation magnitude) by 10, 100, 1000 and multiplying depth predictions consequently.
    \item Varying the contribution of the smoothness loss with $\lambda \in \{1e-1, 1e-2, 1e-3\}$.
    \item Varying learning rate from $1e-3$ to $1e-5$. 
\end{itemize}

\noindent In all these cases, \textbf{the networks still converge to bad quality depth estimations}.

We also investigate \imagesparselidar{} only supervised at the biggest scale to evaluate the influence of multi-scale training in the `P+IMU' setup. We found that performances slightly decreased, but the network still converges to metric depth estimations. Hence, in this setup, multi-scale training does not seem to be crucial.

\end{document}